Data and Text Mining

# wiseR: An end-to-end structure learning and deployment framework for causal graphical models

Shubham Maheshwari[1], Khushbu Pahwa[2], Tavpritesh Sethi[1*]

[1] Indraprastha Institute of Information Technology, Okhla Industrial Estate, Phase III, Near Govind Puri Metro Station, New Delhi, Delhi 110020, India. [2] Delhi Technological University, Bawana Rd, Shahbad Daulatpur Village, Rohini, Delhi, 110042

*To whom correspondence should be addressed.

## Abstract

**Summary:** Structure learning offers an expressive, versatile and explainable approach to causal and mechanistic modeling of complex biological data. We present *wiseR*, an open source application for learning, evaluating and deploying robust causal graphical models using graph neural networks and Bayesian networks. We demonstrate the utility of this application through application on for biomarker discovery in a COVID-19 clinical dataset.

**Availability and implementation:** Web application and source code of the R package are available at wiseR. Result dashboard at COVID19_Biomarkers

**Contact:** tavpriteshsethi@iiitd.ac.in | shubham14101@iiitd.ac.in

**Supplementary information:** Supplementary material is available for download online

## 1 Introduction

Discovering actionable knowledge from data is a major challenge in biology. Modern machine learning (ML) and deep learning have achieved major breakthroughs in fitting datasets. However, correlation does not imply causation, thus limiting the real world potential of these approaches to guide actions. Structure learning is an expressive approach for causal discovery from data (Drton and Maathuis 2017) and the visual nature of graphical structures makes them excellent for interactive decision making (Prosperi et al. 2020; Kleinberg and Hripcsak 2011; Richens, Lee, and Johri 2020). In small datasets with few variables, causal graphs can be built by using expert knowledge. However, this approach is not applicable in most real-world biological datasets which are high-dimensional. Fitting a unified graphical model without relying upon pairwise relationships is known to be an NP-hard problem. Recent advances in score-based and constraint-based approaches to learn Bayesian Networks (Scutari 2017) and more recently Graph ML and Graph Neural Networks (GNNs) (Yu et al. 2019) offer an opportunity to learn these directly from data. These approaches lead to a Directed Acyclic Graph (DAGs) with edge directions encoding causality or the flow of probabilistic influence, hence allowing causal discovery. The learned DAGs also provide a probabilistic model-based representation of knowledge which is appropriate for building actionable models from biomedical datasets. However, there are currently no open-source tools that allow end-to-end building, validating, learning actions and deploying these models.

In this paper, we present w*iseR*, an interactive R package integrating state-of-the-art structure learning methods including Graph Neural Networks, and Bayesian Networks for causal structure learning, visualization, probabilistic inference and model deployment as dashboards.

## 2 Implementation

*wiseR* was created with core R, libraries and in-house pipelines for statistical evaluation and visualization. The app is primarily built using shiny and other associated shiny libraries for the UI needs, visNetwork for the graphical needs and bnlearn, HydeNet for the working logic. Core python machine learning libraries such as pandas, numpy, scipy, pytorch are used for GNN structure learning needs processed through R. It is a cross-platform web app available to use on any system with the R (v>3.5.0) installed. The app has been tested on multiple operating systems using different browsers, ensuring wider availability. To accommodate complex data, file limit has been extended to 8000 MB and provision for parallel computing has been provided, results for which may vary from hardware to hardware. The app can easily be installed and launched as:

```
>library('devools')
>install_github('tavlab-iiitd/wiseR')
>wiser()
```

## 3 Functionalities

The pipeline has been designed for maximum exploitation of complex data, to assist in mining useful information and identify decision-making points which users can then learn optimal decision paths for. The app is divided into 6 tabs; 'App Settings', 'Data', 'Association Network', 'Bayesian Network' ,'Decision Networks 'and 'Publish your dashboard', each of which covers a functionality as suggested in the name. Through 'App Settings' users can enable parallel processing and choose no. of parallel nodes (hardware dependent). Dataset is uploaded and pre-processed to fit learning requirements in the 'Data' tab which also enables variable exploration among other features. A broad relation between the variables can be understood by building and exploring association networks through the 'Association Network' tab or the user can directly skip to 'Bayesian Network' for structure learning. The app enables structure learning using algorithms and parameters of choice, with provision for bootstrapped learning, editing and initializing learned structures and uploading externally learned graphs. 'Hill Climbing', 'Grow-Shrink' and 'DAG-GNN' are some of the graph structure learning methods available through the app. The learned structure can then be analyzed for insights using graph exploration tools at disposal. Variables of interest are used as evidence/event to build inference plots and actionable results can then be reported, all this can be done in the 'Bayesian Network' tab. To learn optimal decision paths for the actionable points the user can move to the 'Decision Network' tab which utilizes the learned structure as the decision network. The user can set the utility node along with the value for each outcome which they wish to maximize/minimize and set the decision nodes for which they want to learn the optimal policy. The app makes it possible to learn the best decision path on the utility node as well as compare it with other courses of actions. The app also has provision to download a customized dashboard as an R package, which can be used to explore and infer from the results independently and is groundbreaking in terms of ease of publishing. This feature is enabled in the 'Publish your dashboard tab'. A detailed tutorial with a walkthrough of the app layout and functionalities is also available in the 'App Tutorial' section as well as a supplementary folder(wiseR_Tutorial.pdf) along with a use-case, replicating validated results to show the utility of Bayesian networks.

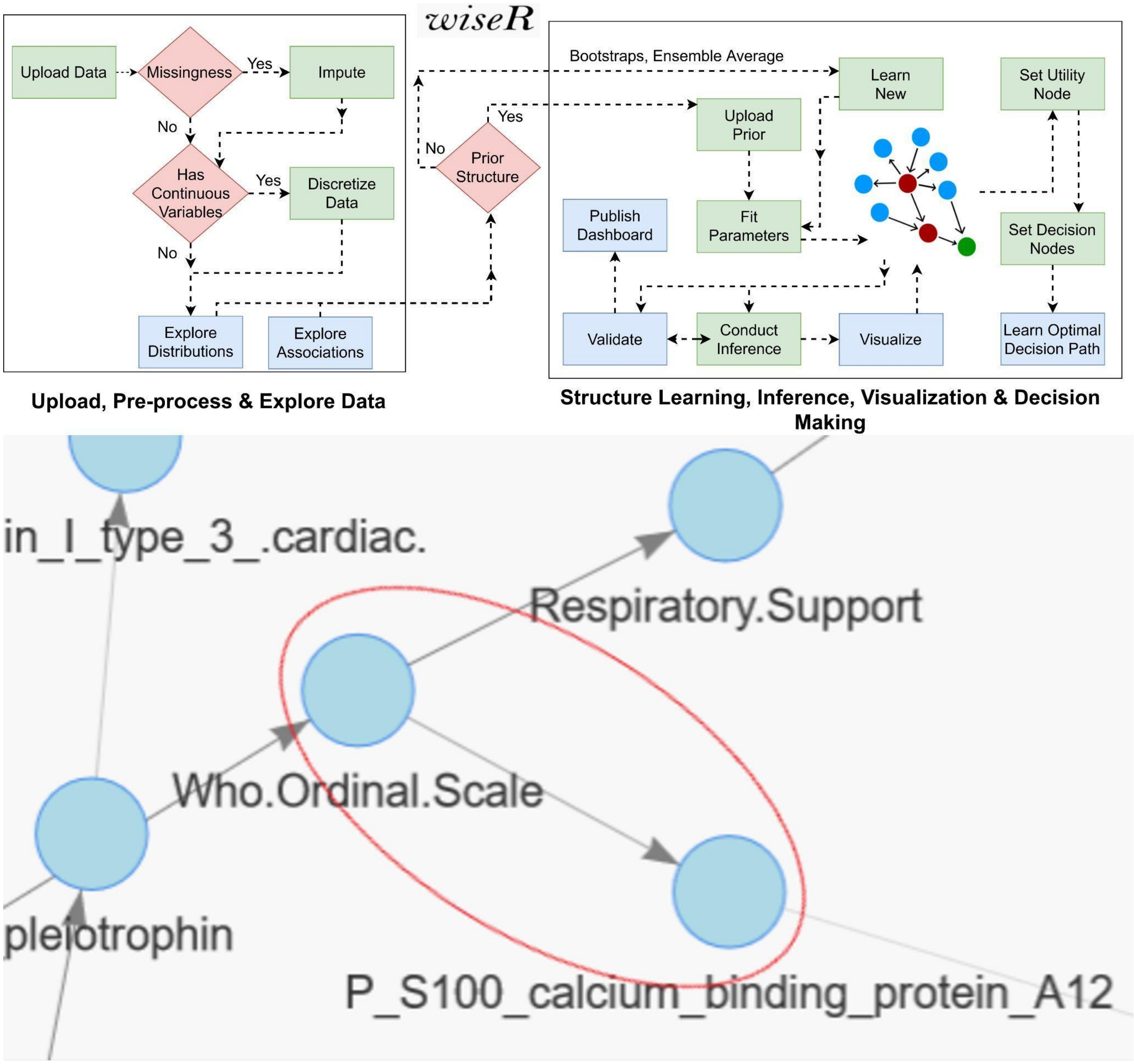

**Fig. 1.** Workflow of the App on the top; learned structure at the bottom verifying causality between S100A12 protein and covid-19 severity (Measures using WHO ordinal scale)

## 4 Case Study: COVID-19 Biomarker Discovery

Covid-19 has many countries strapped for medical resources, highlighting the need for allocation of these resources to those more in need. Several markers have been associated with severity in covid patients, none with enough prognostic power, thus need for better biomarkers is essential. We utilized the wiseR app on a comprehensive clinical and multi-omics dataset of covid-19 patients (Su et al. 2020), and discovered S100A12 as the strongest actionable node for predicting the severity of infection in covid-19 patients as described on WHO ordinal scale. The causality between the 2 is shown in the bottom half of Fig-1. As processed using wiseR. Our results were verified in a recent paper finding the potential of S100A12 as a single transcript to guide clinical decisions (Lei 2021). Interestingly, S100A12 is also a known biomarker for sepsis and wiseR was able to prioritize this from a set of potential markers, while allowing for causal discovery and probabilistic inference. A detailed walkthrough for reproducibility of results, processed dataset and exact inferences drawn from this causal link can be found in the appendix section below. Identifying such prominent prognostic biomarkers amidst a pandemic such as COVID-19 can help translational discovery, early identification, and design of better interventions, potentially saving many lives. A dashboard to explore the network or evaluate the findings is also made available, generated through wiseR itself, highlighting its efficacy in improving reach and understanding of medical research.

## 5 Conclusion

We present wiseR, an end-to-end causal discovery web-application using explainable graphical models. This application has been validated and tested for numerous use cases in public health (Sethi et al. 2019; Awasthi et al. 2020; Jha et al. 2021) and we show for the first time the use of such an approach for biomarker discovery for predicting severity in COVID-19 patients. wiseR provides state-of-the-art algorithms in graph neural networks and Bayesian Networks for causal discovery. The end-to-end framework allows for deployment of these models in translational scenarios by experimental biologists and clinicians without the need for expertise in causal networks.

## Acknowledgements

This work was supported by the DBT/Wellcome Trust India Alliance Fellowship IA/CPHE/14/1/501504 awarded to Tavpritesh Sethi. We are also thankful to Anant Mittal for testing the application and providing suggestions.

## References

Awasthi, Raghav, Prachi Patel, Vineet Joshi, Shama Karkal, and Tavpritesh Sethi. 2020. “Learning Explainable Interventions to Mitigate HIV Transmission in Sex Workers Across Five States in India.” *NeurIPS Workshop (2020).*

Drton, Mathias, and Marloes H. Maathuis. 2017. “Structure Learning in Graphical Modeling.” *Annual Review of Statistics and Its Application* 4 (1): 365–93.

Jha, Indra Prakash, Raghav Awasthi, Ajit Kumar, Vibhor Kumar, and Tavpritesh Sethi. 2021. “Learning the Mental Health Impact of COVID-19 in the United States With Explainable Artificial Intelligence: Observational Study.” *JMIR Mental Health* 8 (4): e25097.

Kleinberg, Samantha, and George Hripcsak. 2011. “A Review of Causal Inference for Biomedical Informatics.” *Journal of Biomedical Informatics* 44 (6): 1102–12.

Lei, Hongxing. 2021. “A Single Transcript for the Prognosis of Disease Severity in COVID-19 Patients.” *Scientific Reports* 11 (1): 12174.

Prosperi, Mattia, Yi Guo, Matt Sperrin, James S. Koopman, Jae S. Min, Xing He, Shannan Rich, Mo Wang, Iain E. Buchan, and Jiang Bian. 2020. “Causal Inference and Counterfactual Prediction in Machine Learning for Actionable Healthcare.” *Nature Machine Intelligence* 2 (7): 369–75.

Richens, Jonathan G., Ciarán M. Lee, and Saurabh Johri. 2020. “Improving the Accuracy of Medical Diagnosis with Causal Machine Learning.” *Nature Communications* 11 (1): 3923.

Scutari, Marco. 2017. “Bayesian Network Constraint-Based Structure Learning Algorithms: Parallel and Optimized Implementations in the Bnlearn R Package.” *Journal of Statistical Software*. https://doi.org/10.18637/jss.v077.i02.

Sethi, Tavpritesh, Anant Mittal, Shubham Maheshwari, and Samarth Chugh. 2019. “Learning to Address Health Inequality in the United States with a Bayesian Decision Network.” *Proceedings of the AAAI Conference on Artificial Intelligence* 33 (01): 710–17.

Su, Yapeng, Daniel Chen, Dan Yuan, Christopher Lausted, Jongchan Choi, Chengzhen L. Dai, Valentin Voillet, et al. 2020. “Multi-Omics Resolves a Sharp Disease-State Shift between Mild and Moderate COVID-19.” *Cell* 183 (6): 1479–95.e20.

Yu, Yue, Jie Chen, Tian Gao, and Mo Yu. 2019. “DAG-GNN: DAG Structure Learning with Graph Neural Networks.” *ICML (2019)*.

## Appendix

# Structure Learning Based Discovery of S100A12 as a marker for COVID-19 Severity

## 1 Result dashboard installation

> library('devools')
> install_github('tavlab-iiitd/ COVID19_Biomarkers')
> COVID19_Biomarkers::biomarkers()

## 2 Supplementary Material

- **data.csv**: Contains the discretized and filtered data used for structure learning and inferences
- **blacklist.csv**: Contains the list of edges blacklisted from structure learning as shown above.
- **wiseR_Tutorial**.pdf: Detailed tutorial of the wiseR application with use case examples in inference and decision from real world datasets.

## 3 Data Processing

- We utilize an open source covid-patient dataset. (Su et al., 2020)
- We merge the section of the data with information on Protein analysis, Metabolomes and Clinical variables. PCA is performed to trim down Protein and Metabolome variables, to keep only the most relevant and contributing ones.
- Data is then uploaded to wiseR as ".CSV" and Discretized

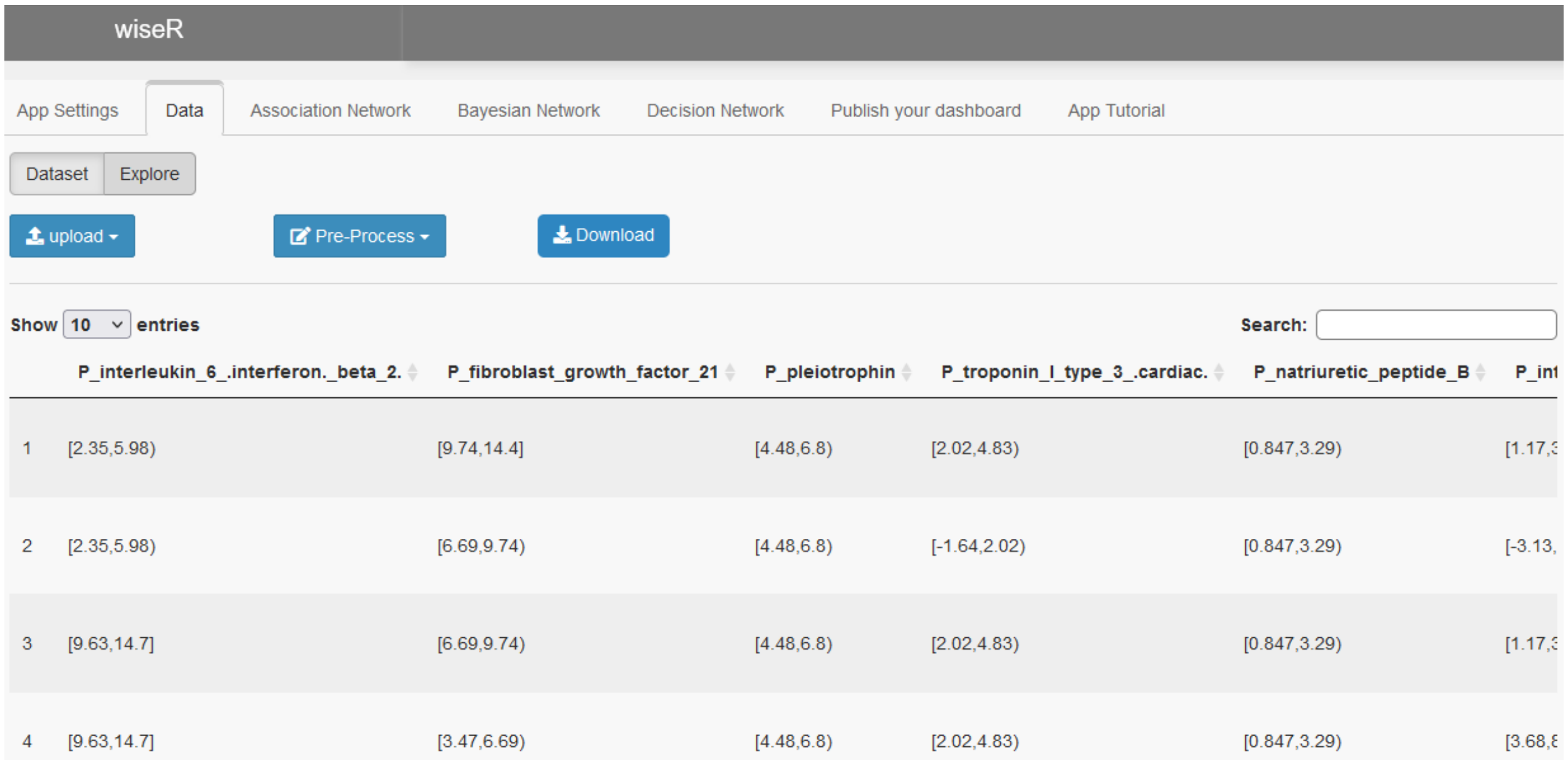

Fig. 1. Shows data uploaded and discretized through the wiseR app.

- A blacklist of edges is created to prevent flow of causality from covid variables to pre-covid variables. These blacklist edges are saved in a "from-to" table in a ".csv" file.

# 4 Structure Learning

- We perform Bayesian Structure Learning on the data using wiseR and the following parameters:-
  - Algorithm: Hill Climbing
  - Network Score: Akaike Information Criterion
  - Parameter Fitting: Bayesian Parameter Estimation
  - Blacklisting: Yes (csv file of blacklist provided)
  - Bootstrap Model: Yes
  - No. of Bootstrap:51
  - Edge Strength: >0.51
  - Direction Strength: >0.51
  - Sample Portion: 1

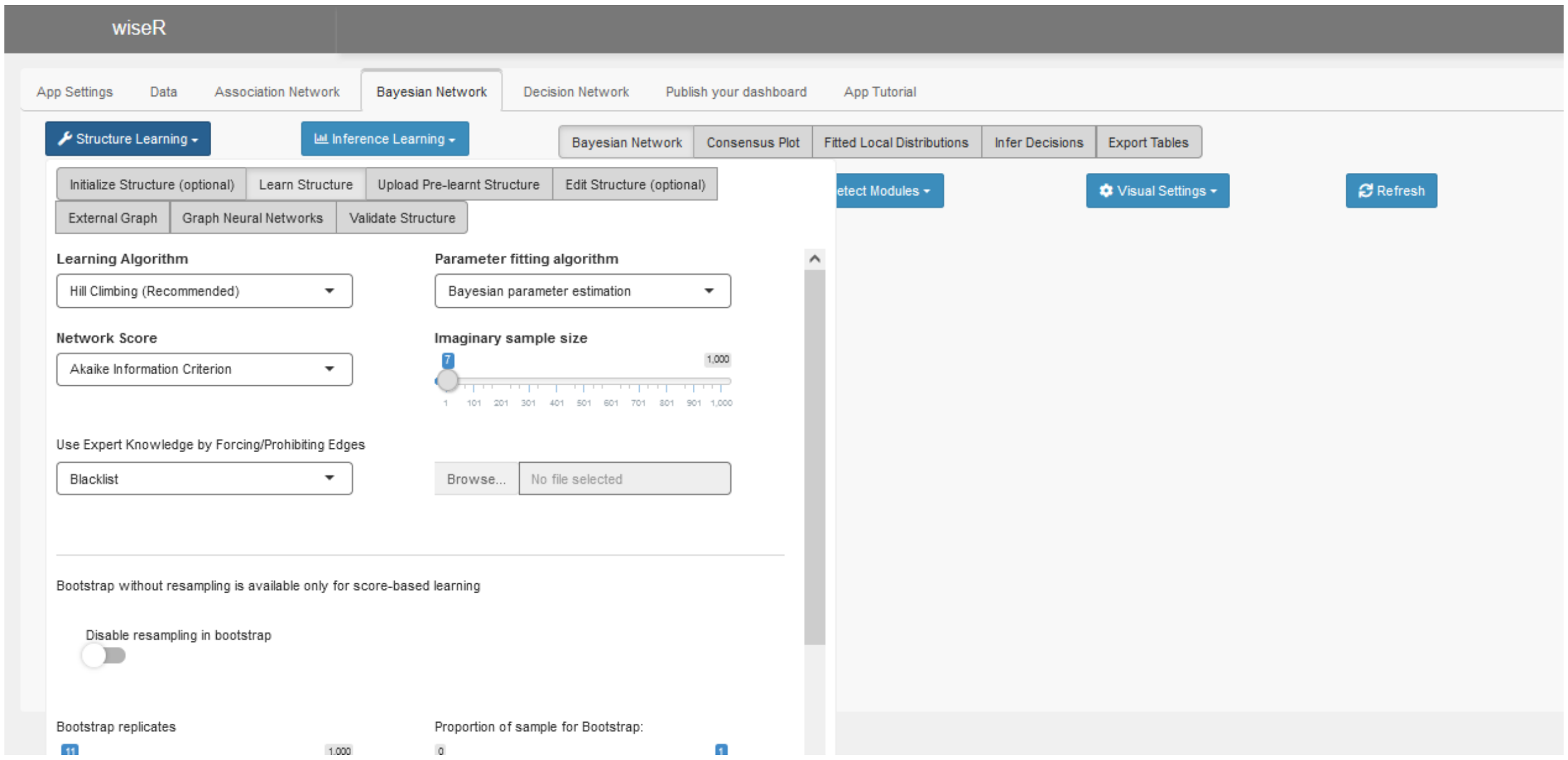

Fig. 2. Parameters of Bayesian Structure Learning

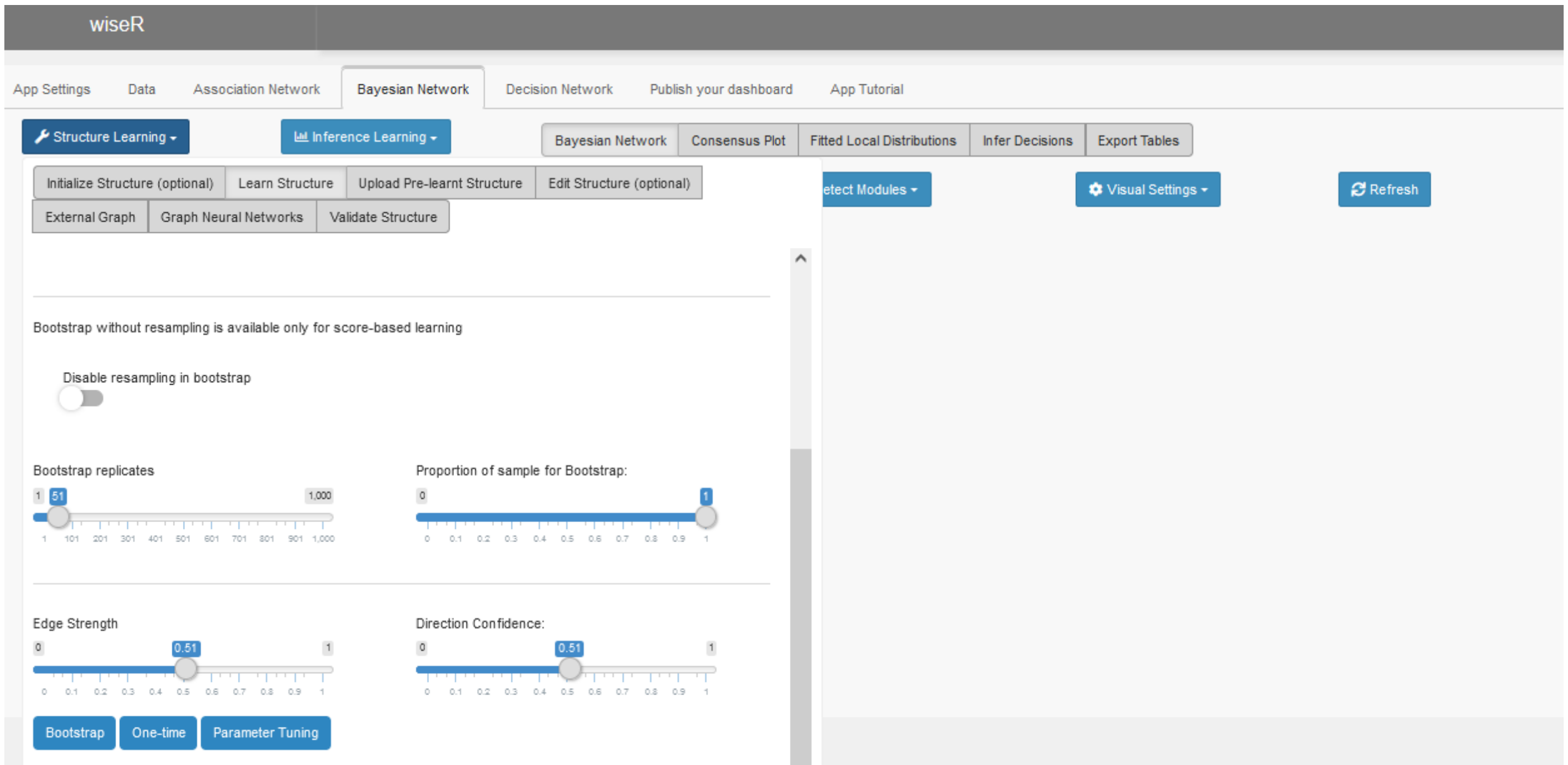

Fig. 3. Parameters of Bootstrap Learning

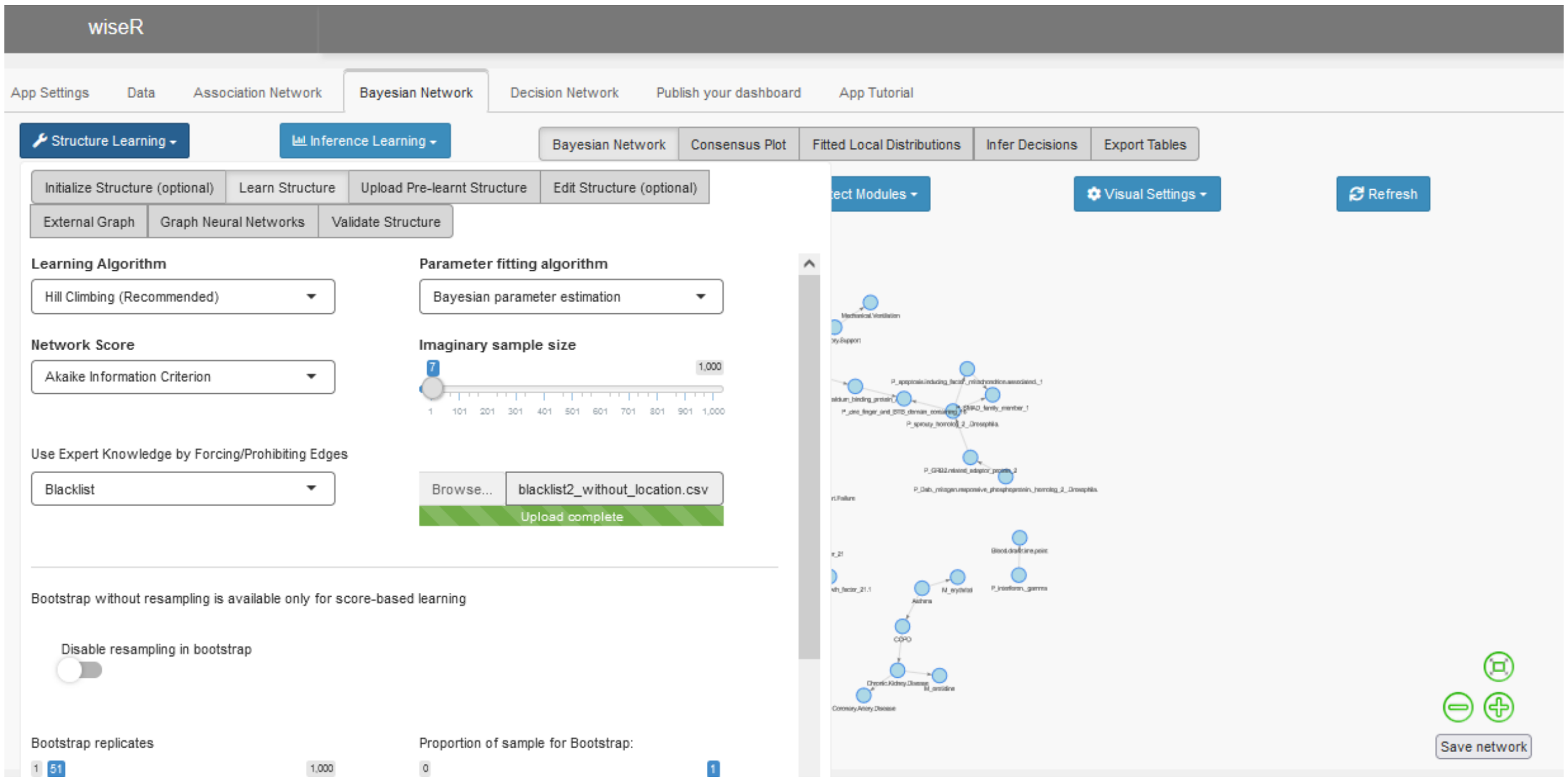

Fig. 4. Utilizing Blacklist Edges

## 3 Understanding Network and Inference

- We can observe in the learned structure the link between Who Ordinal Scale for Covid Severity and S100A12

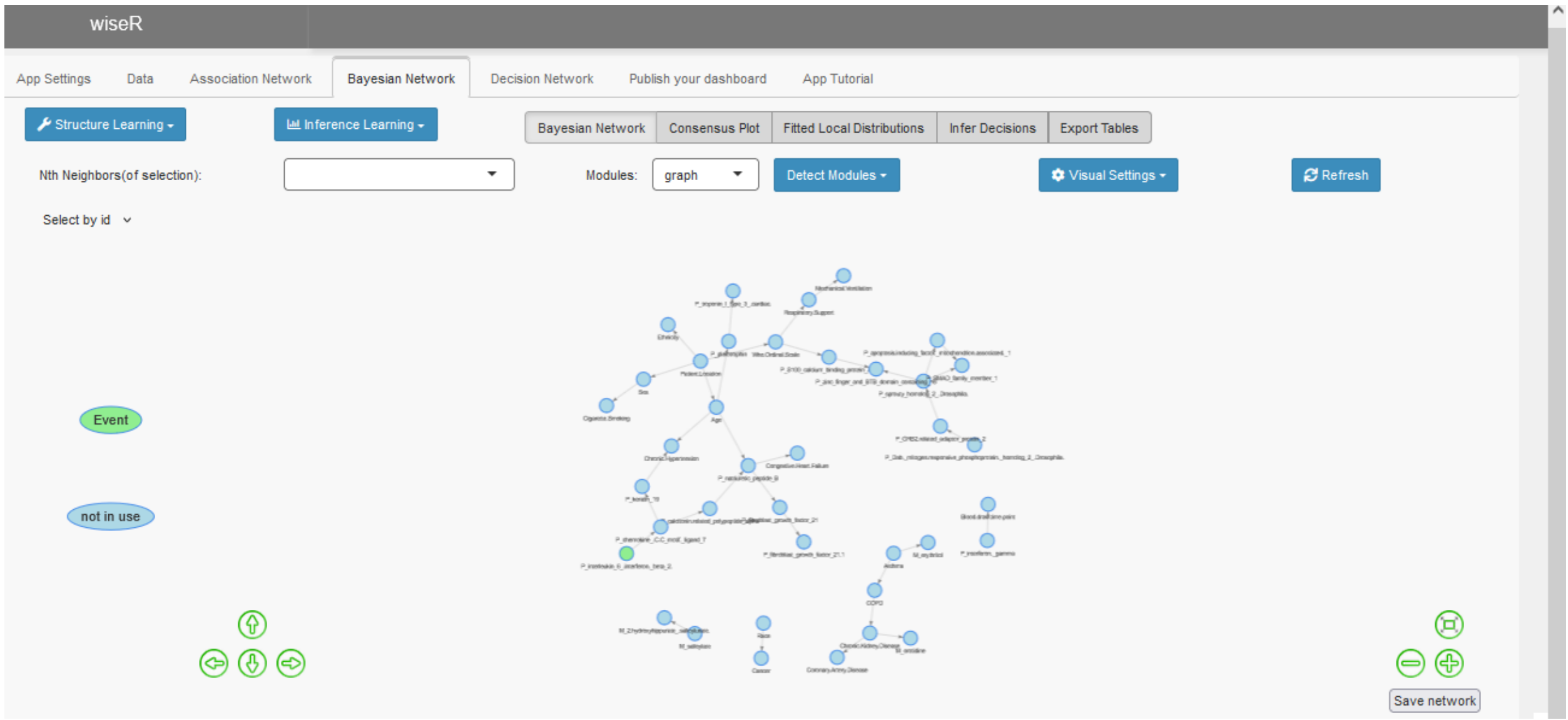

Fig. 5. Learned Network Structure

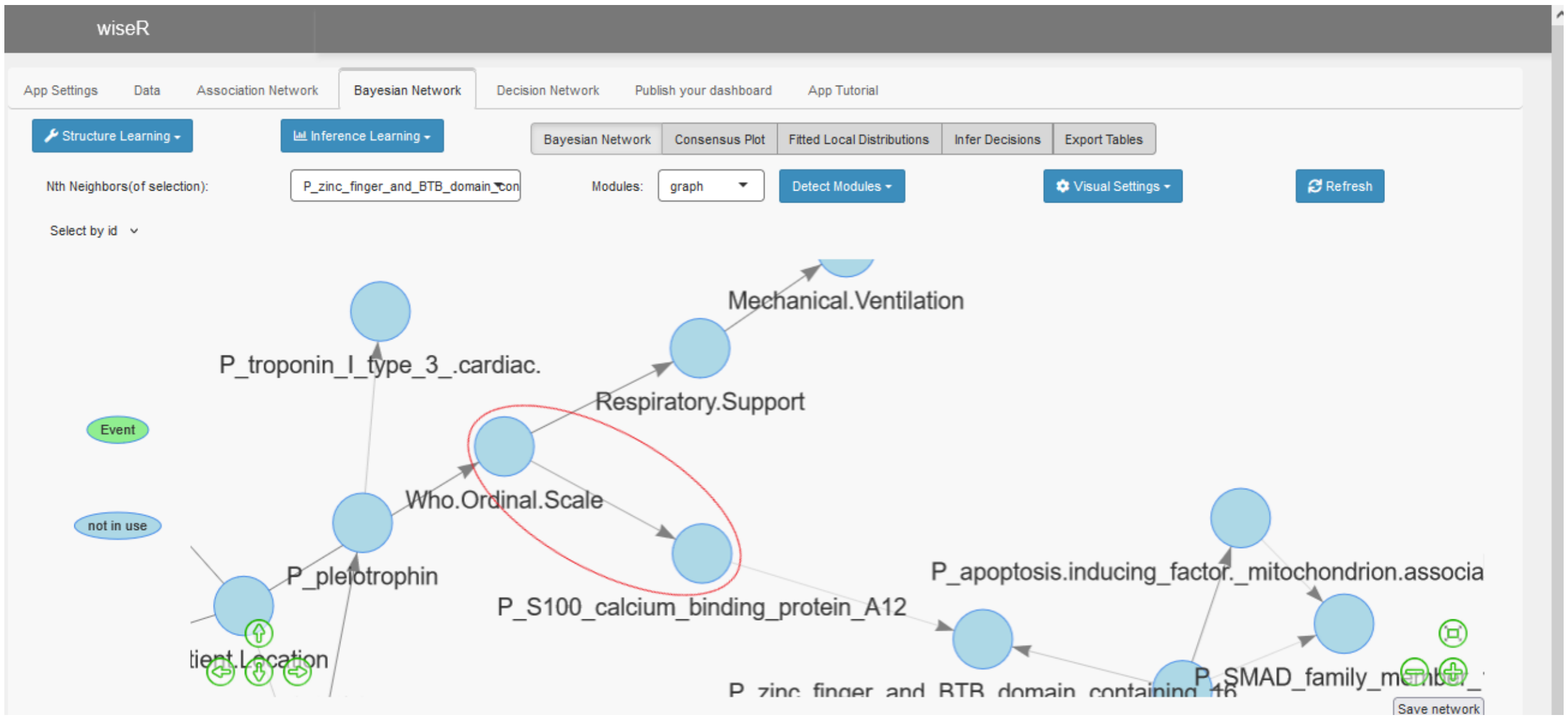

Fig. 6. Directed edge between WOS and S100A12

- Our network just validates the causality already verified (Lei, 2021)
- We use an inference plot to show that S100A12 protein markers are positively causal to covid severity. WOS rate covid severity on a scale of 1-7, 1 being the mildest to 7 being most severe. Below plots show how increased value of S100A12 protein markers indicate the likelihood of covid severity in patients.

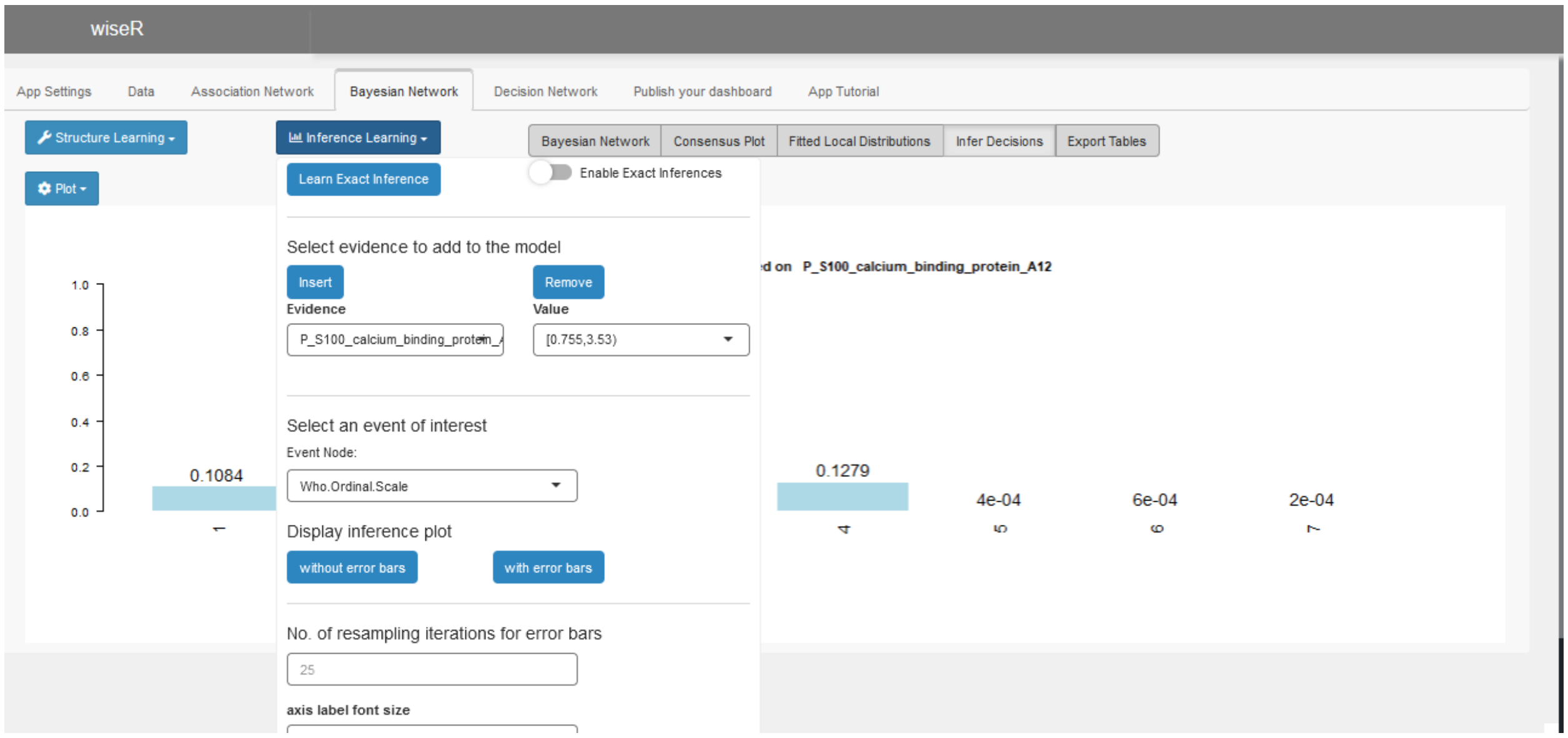

Fig. 7. Set S100A12 in the lowest range

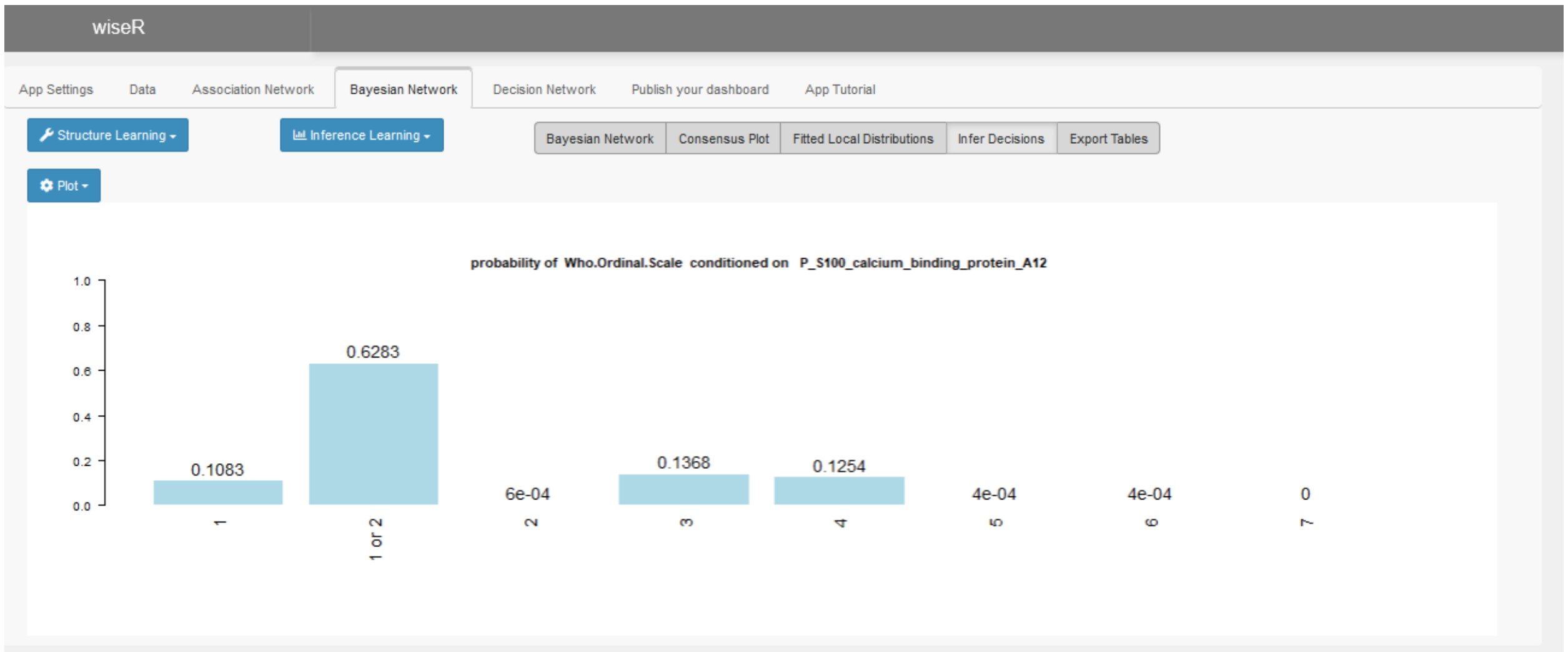

Fig. 8. Lowest Bracket for S100A12 markers indicate high chances of mild covid

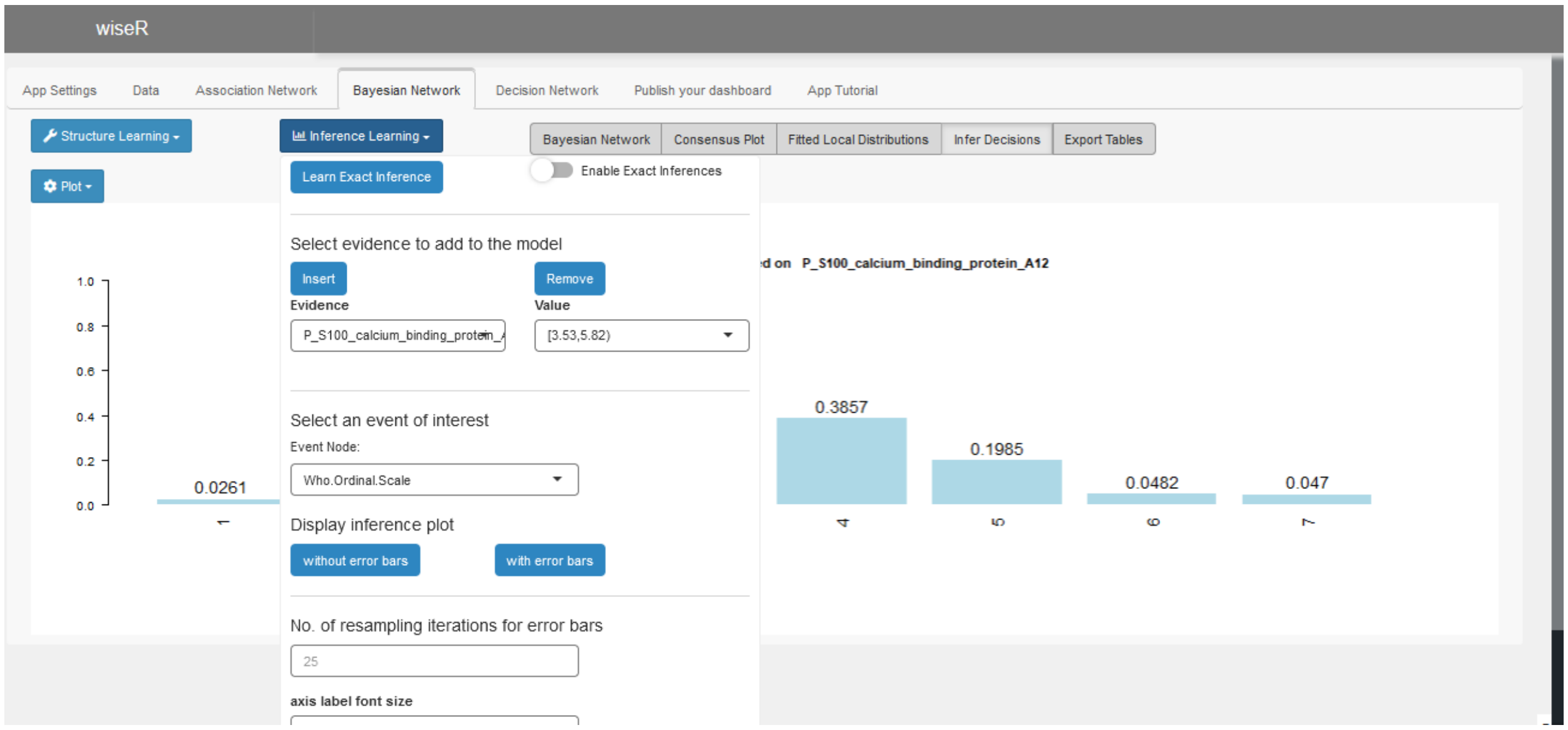

Fig. 9. Set S100A12 in the middle range

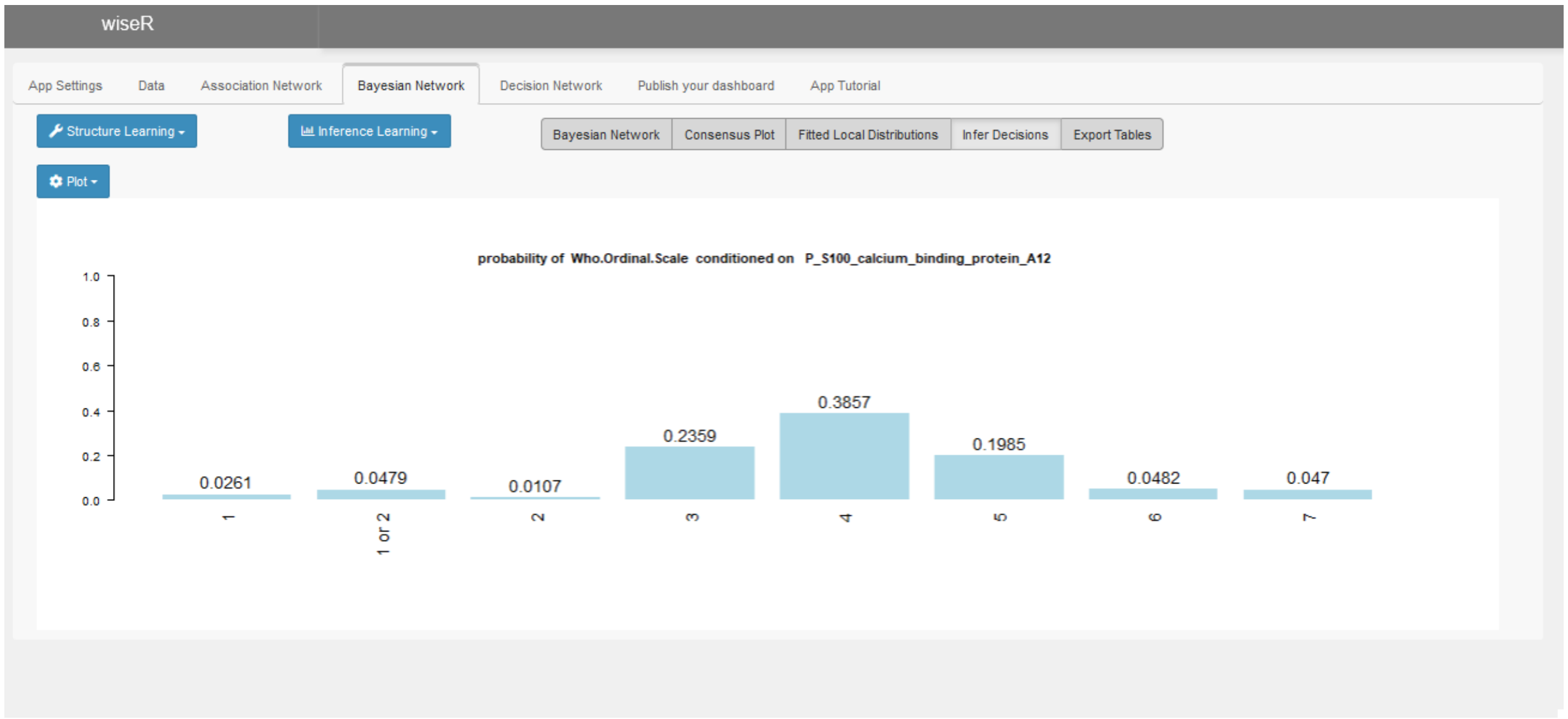

Fig. 10. Middle Bracket for S100A12 markers indicate high chances of moderate to severe covid

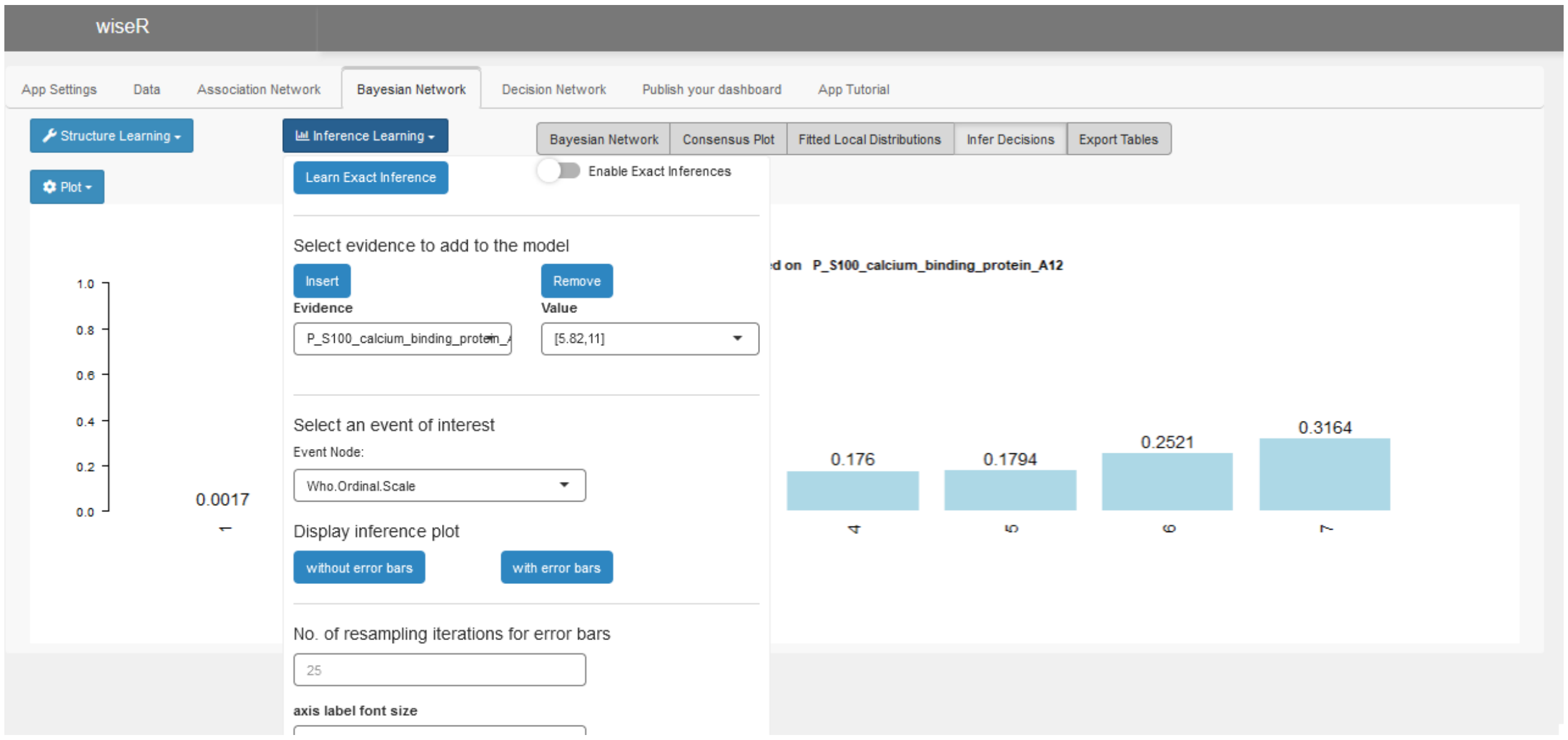

Fig. 11. Set S100A12 in the Highest range

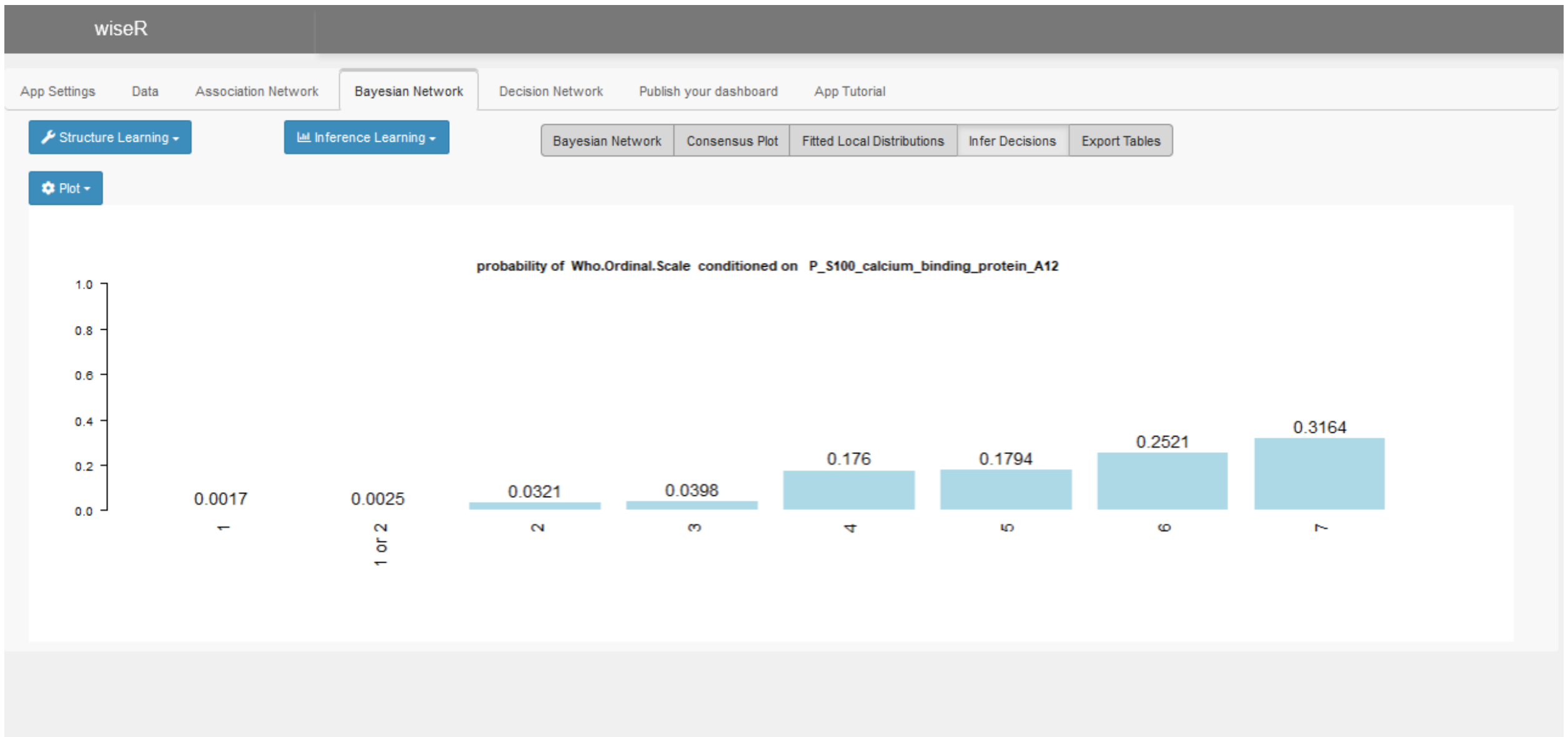

Fig. 12. Highest Bracket for S100A12 markers indicate high chances of severe covid

# wiseR- Comprehensive Tutorial for Structure Learning and Decision Making

## 1 Getting Started

- To install latest developmental version of wiseR in R
  > **devtools::install_github('tavlab-iiitd/wiseR',build_vignettes=TRUE)**
- To launch the app:
  > **wiseR::wiser()**

## 2 Bayesian Networks

### 2.1 Why Bayesian Networks?

Networks are one of the most intuitive representations of complex data. However, most networks rely on pairwise associations which limits their use in making decisions. Bayesian Networks (BNs) are a class of probabilistic graphical models which can provide quantitative insights from data. These can be used both for probabilistic reasoning and causal inference depending upon the study design.

A BN is a directed acyclic graph and provides a single joint-multivariate fit on the data with a list of conditional independencies defining the model structure. Unlike multiple pairwise measures of association, fitting a model decreases the chance of false edges because the structure has to agree with global and local distributions. Importantly, unlike most other forms of Artificial Intelligence and Machine Learning, BNs are not a black-box model. These are transparent, interpretable and help the user in reasoning about the data generative process by looking at the motifs. This can be immensely useful in learning systems such as healthcare where feedback between the clinicians and the learning system is paramount for adoption. The structure (independencies) of a BN can be learnt directly from data using machine learning or be specified by the user, thus allowing expert knowledge to be injected. After the dependence structure within the data is learnt (or specified) the parameters are learnt on the network using one of many possible approaches. wiseR provides most of the existing approaches such as constraint based algorithms and score-based algorithms for parametrization of the Bayesian Network. Recommendations as per the state-of-the-art in the literature are specified at each step of the BN learning process (i.e. the recommendation of score based algorithms over constraint based algorithms for learning structure and Bayesian Information Criteria over Akaike Information Criteria for evaluating the fit). Parametrization of the network enables predictions and inferences. These inferences can be purely observational ("seeing" the state of a variable and its effect on neighbours) or causal ("doing" something to a variable, i.e. interventions and observing the effect on downstream nodes). wiseR provides scoring methods both for observational (e.g. Bayesian Information Criterion) and interventional (e.g. modified Bayesian Dirichlet Equivalent score) datasets.

The flexibility provided by BNs in probabilistic reasoning and causal inference has made these one of the most widely used artificial intelligence methods in Computer Science. However, the sophistication required to code all the features has limited their use by domain experts outside of computer science, e.g. clinicians. wiseR plugs this gap by creating a GUI for the domain experts and is an end-to-end solution for learning, decision making and deploying decision tools as dashboards, all from the wiseR platform.

### 2.2 What do the motifs (junctions) reveal about the data-generating process

Carefully learnt BNs are representations of the generative process that produced the data. These generative processes are captured in the the form of three basic motifs present in the network: chain, fork and collider junctions.

### 2.3 Chain (Mediator effect)

This motif is the simplest structure in a BN with a sequence of nodes pointing in the same direction. The intermediate nodes are called mediators and conditioning on any of the mediators makes the flanking nodes independent of each other. A mediator can be thought of as a mechanism, hence capturing the information about the mechanism removes the need for capturing the triggering process, hence making the triggering process independent of the outcome.

### 2.4 Fork (Confounders effect)

The second type of structure observed in a BN is a common parent pointing towards two or more child nodes. The parent represents the common cause and not accounting for the parent is a common source of error (confounding) in many statistical models built upon real world data. A folk example of such an effect is Age as a confounder (common cause) of IQ and shoe-size in children. Failure to condition upon age will lead to a spurious correlation between the IQ and shoe-size and change our reasoning (hence predictions) about the system being modeled.

### 2.5 Collider or a V-structure (Inter-causal reasoning)

These are one of the most interesting motifs in a Bayesian Network and are indicated by two nodes pointing towards a single child node. Although the two parent nodes are not causally connected, conditioning on the state of the common child opens up a ghost-path for probabilistic inference flow (inter-causal reasoning) which can explain many intriguing paradoxes (e.g. the Monty Hall problem).

## 3 Walkthrough of wiseR functionalities

### 3.1 Pipeline

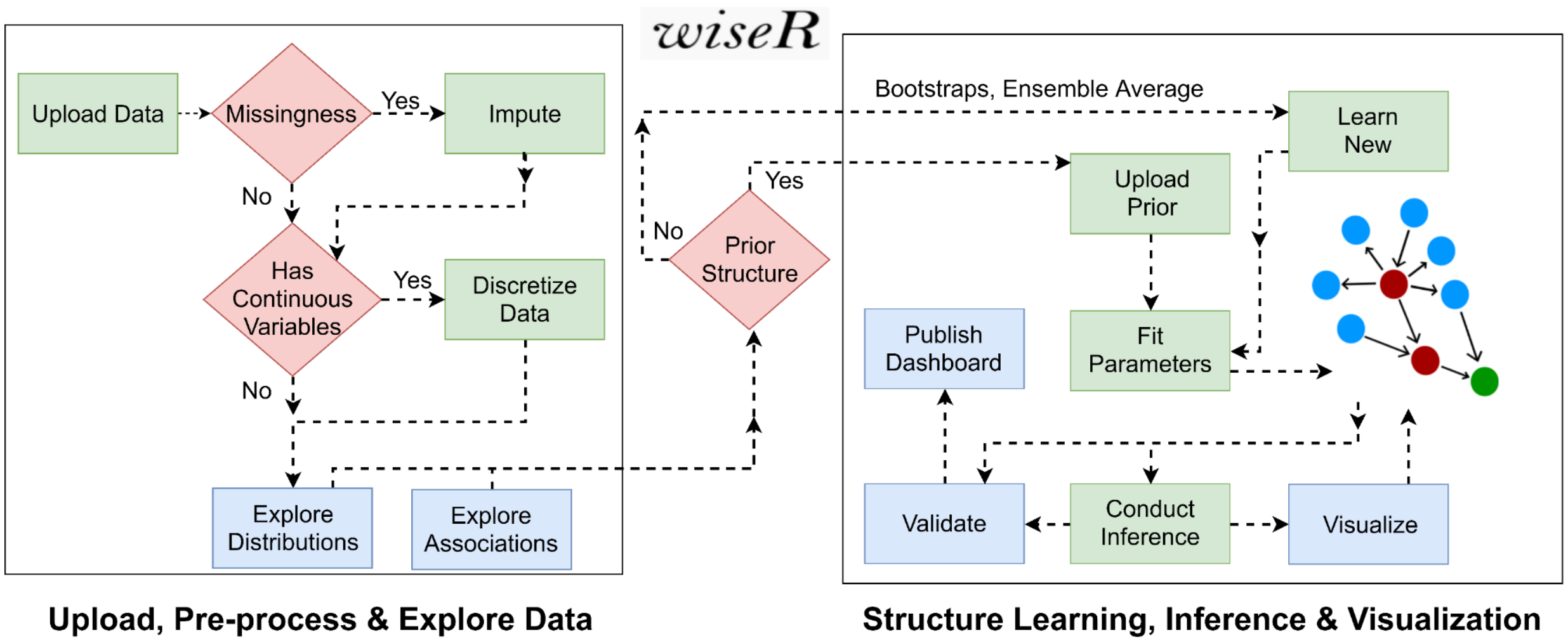

Figure S1. Flowchart showing logic of wiseR

### 3.2 Start

Calling the wiseR() function from the wiseR package launches the application in the default browser. The recommended browser for wiseR is Chrome. On launching the application for the first time, it conducts background checks, which takes 5-10 seconds depending upon the machine. This delay only happens on the first launch and all subsequent launches take less than a couple of seconds to initialize the user interface.

The toolbar on the left side of the page has icons for navigating to home, analysis engine, github development version and team information at any point of time.

Click “Start Analyzing” for navigating to the analysis engine with tabs specifying the functions. Hovering over most of the tabs brings up a tool-tip that briefly describes the function performed on that action.

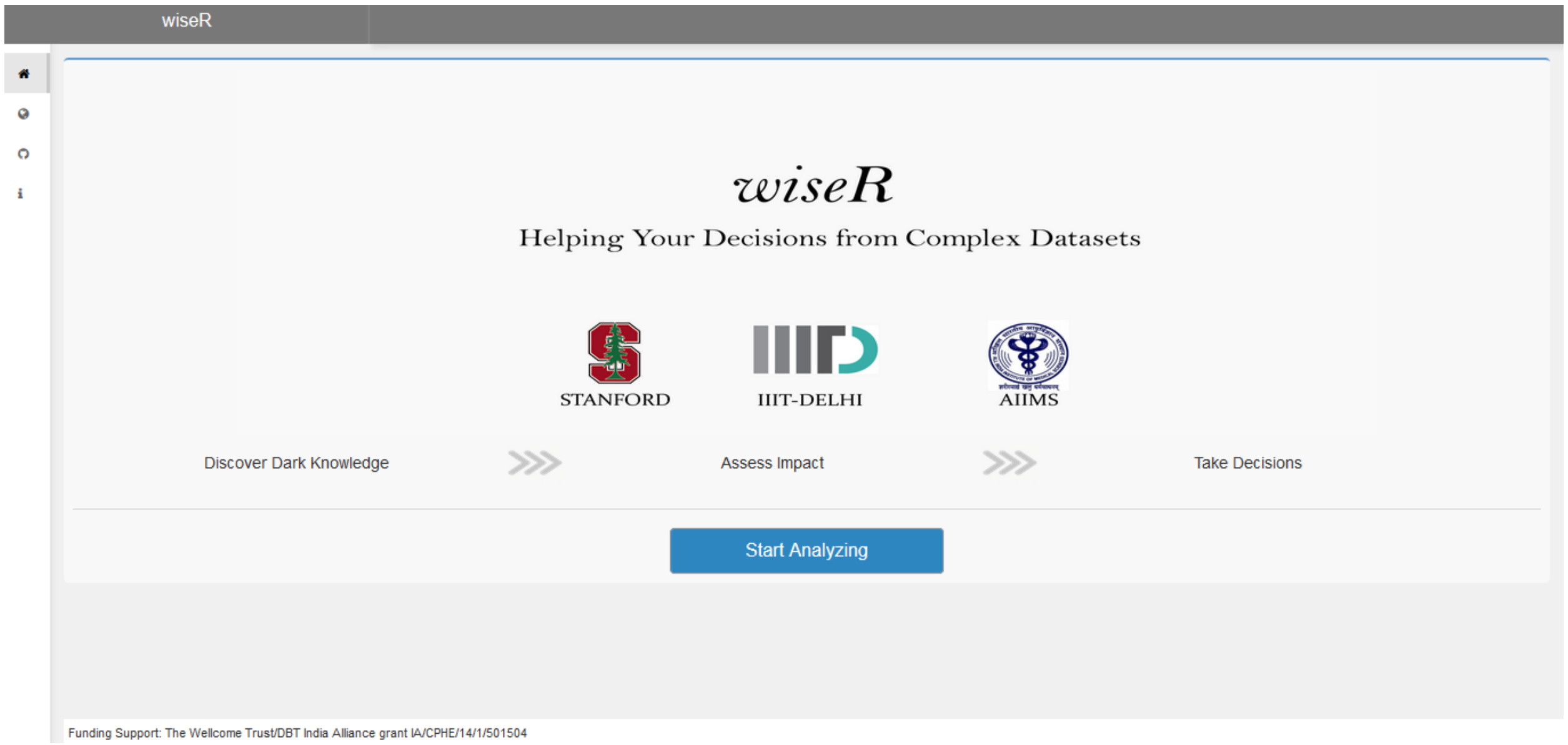

Figure S2. Homepage of the wiseR application

### 3.3 Main page of the wiseR engine

“Start Analyzing” takes us to the main page of the wiseR engine. An intuitive left-to-right ordering of tabs guides the user into the analysis. This page has 5 tabs named ‘App Settings’,‘Data’,‘Association Network’,‘Bayesian Network’ and ‘Publish your dashboard’, each tab covering a specific functionality (Figure S3). Each of these is described next.

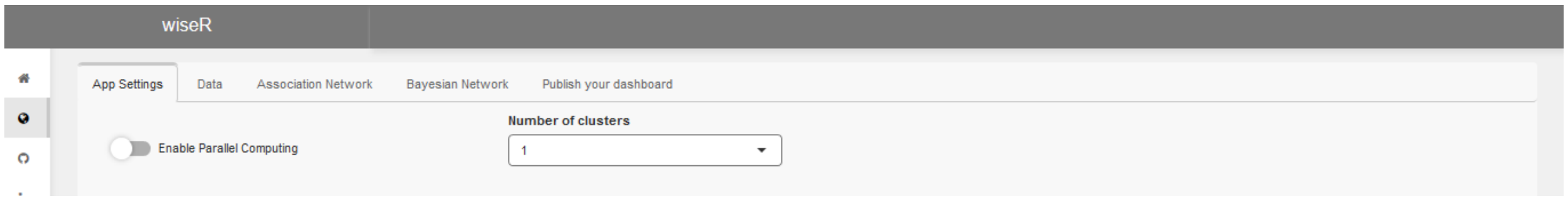

Figure S3. Functional tabs on the main page of the wiser engine and the first tab (App Settings) for parallel computation

### 3.4 App Settings

This tab (Figure S3) is used to set the parallel computing option (number of cores) for learning BN structure. Learning structure is known to be an NP-hard problem and may be time consuming on large datasets. Setting the number of clusters allows each core to learn a structure when bootstrap learning is performed. Since learning one structure cannot be parallelized, this is an example of task parallelization.

### 3.5 Data

This tab is used to load, pre-process and explore the dataset (Figure S4.)

The dataset panel contains the upload and preprocess menu while the explore panel is used to visualize the distribution of the data.

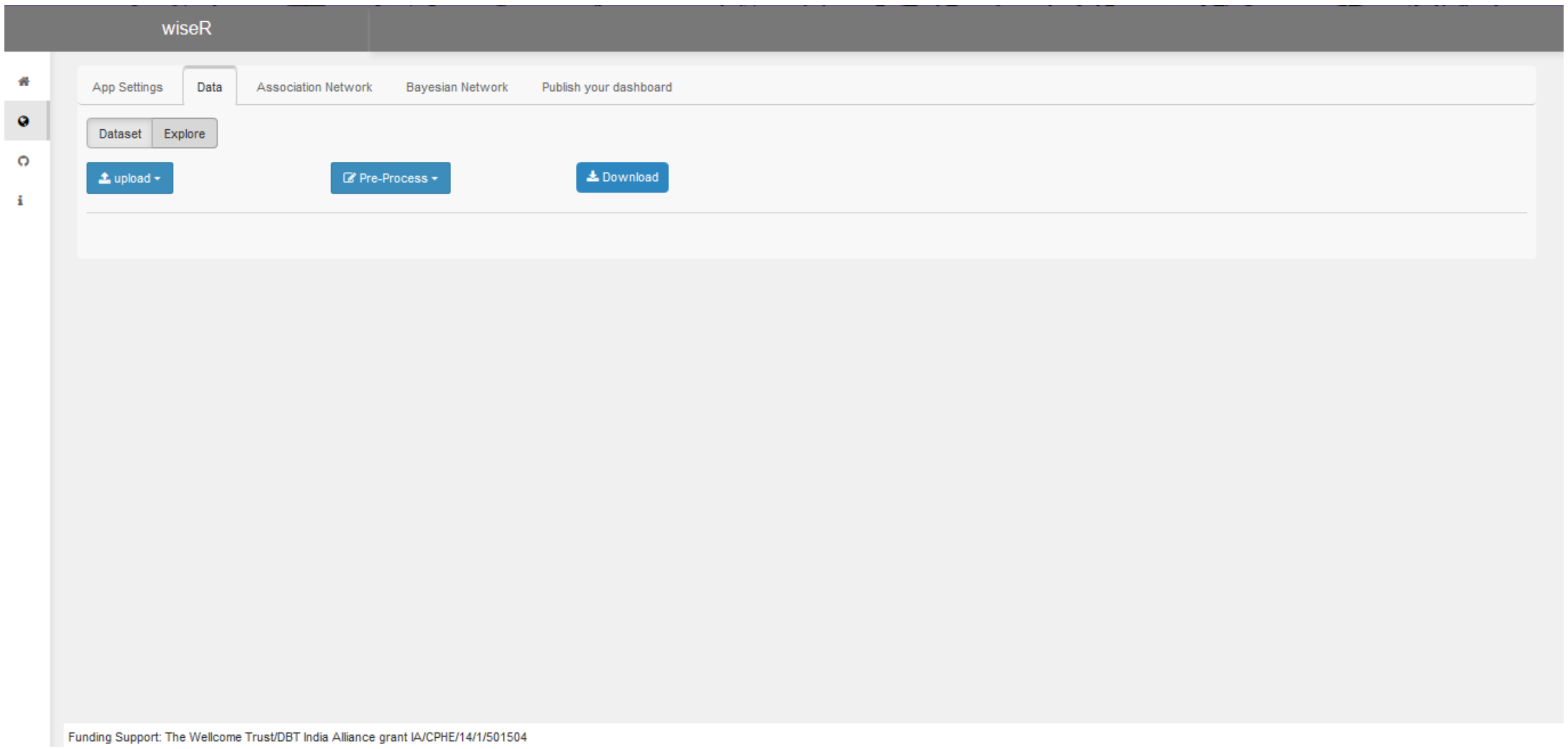

Figure S4. Data upload, exploration and pre-processing functions.

### 3.5.1 Upload

Users can upload their own data in one of the '.CSV','.RData' & '.txt' formats. For large data, saving the data as .RData is recommended for its compression of data, hence efficient storage. For '.txt' files different options of file formatting are provided such as:

- Comma Separated
- Semi-colon Seperated
- Tab separated
- Space Separated

New users trying to learn about BNs or testing the functionality of the app can work with the preloaded datasets (Alarm, Hailfinder, Asia, Coronary) to replicate BN analysis commonly demonstrated with these datasets.

Users must note that the default setting of the app is to treat all variables with less than 53 levels as factor variables. Variables with more than 53 levels are treated as numeric variables and the user is prompted for discretization of such numeric variables to factors. While this is a convenient general option, it may be true that a numeric variable may have less than 53 levels in a particular dataset. In such cases, the user can un-check this option and specify the numeric variables explicitly to be converted to factors in the next step.

Current limit of file upload is 8000 mb, however it is recommended to use .RData for large datasets.

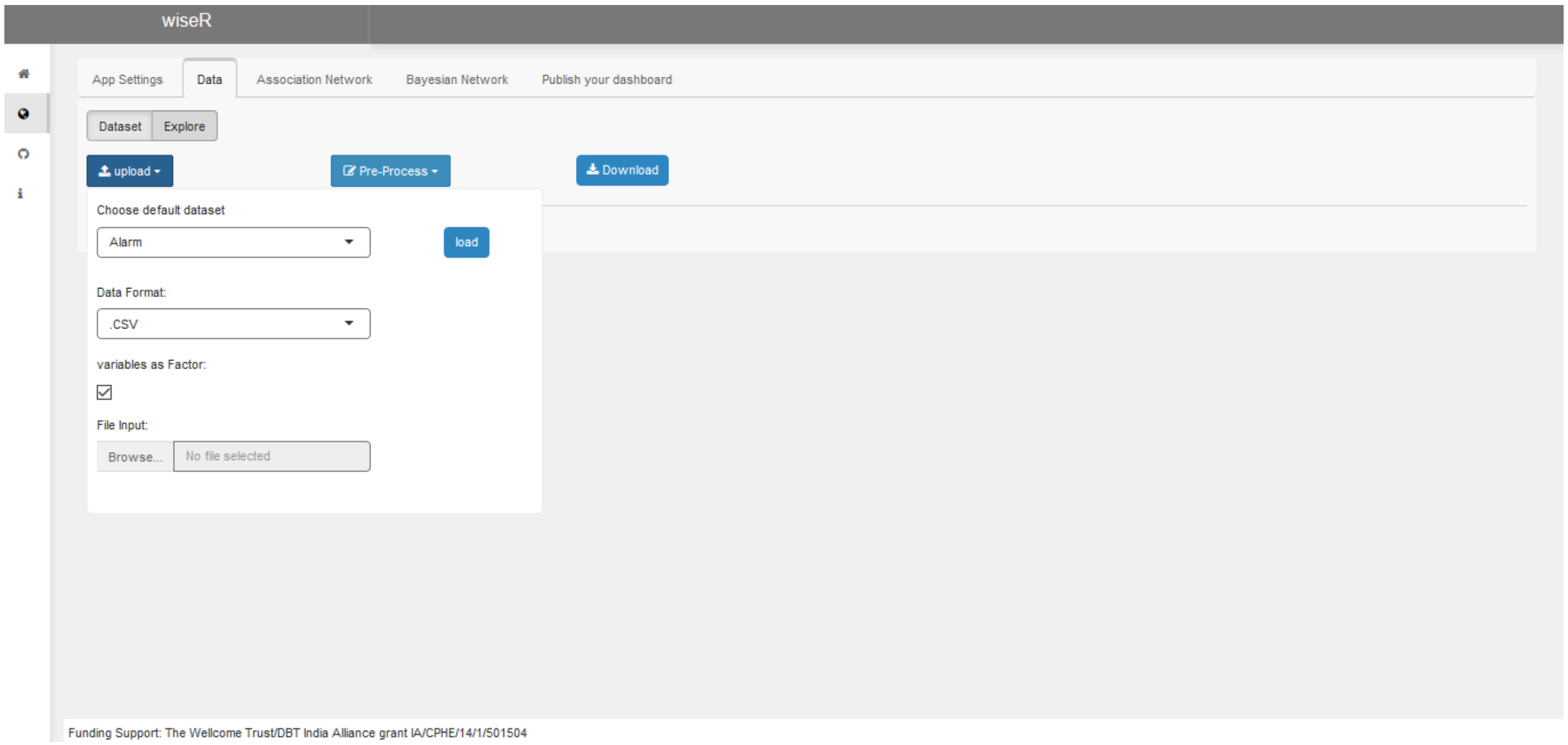

Figure S5. Features of the upload subsection of the Data tab.

### 3.5.2 Pre-process

The preprocess menu is used to edit the data before using it through the app. It is essential that data is discrete and complete for learning association networks and Bayesian networks. Multiple options to pre-process the data have been provided.

- User can convert any variable which has been misclassified as numeric to factor
- User can convert any variable which has been misclassified as factor to numeric
- User can choose from a number of algorithms to discretize their data - Hybrid(Recommended),Hartemink(Recommended),K-means,Quantile,Frequency,Interval
- Hybrid discretization is the default and it automatically chooses the best option for discretizing the variable. If discretization fails with the current method, it uses the next preferable method in the following sequence- k-means, quantile and interval discretization in that order of preference.
- Hartemink requires 2 parameters 'break' and 'ibreak' which can be passed using the text input boxes.
- Transpose the data frame if necessary.
- Sort variables alphabetically.
- Drop/Reset variables from the analysis. Using the delete button after entering variable names drops these from the analysis and reset button restores the original data.them from the data using the delete button.
- Select Interventions. The learning algorithm will recognize the selected variable as intervention, and will use modified Bayesian Dirichlet Equivalent score for structure calculation after accounting for the intervention.

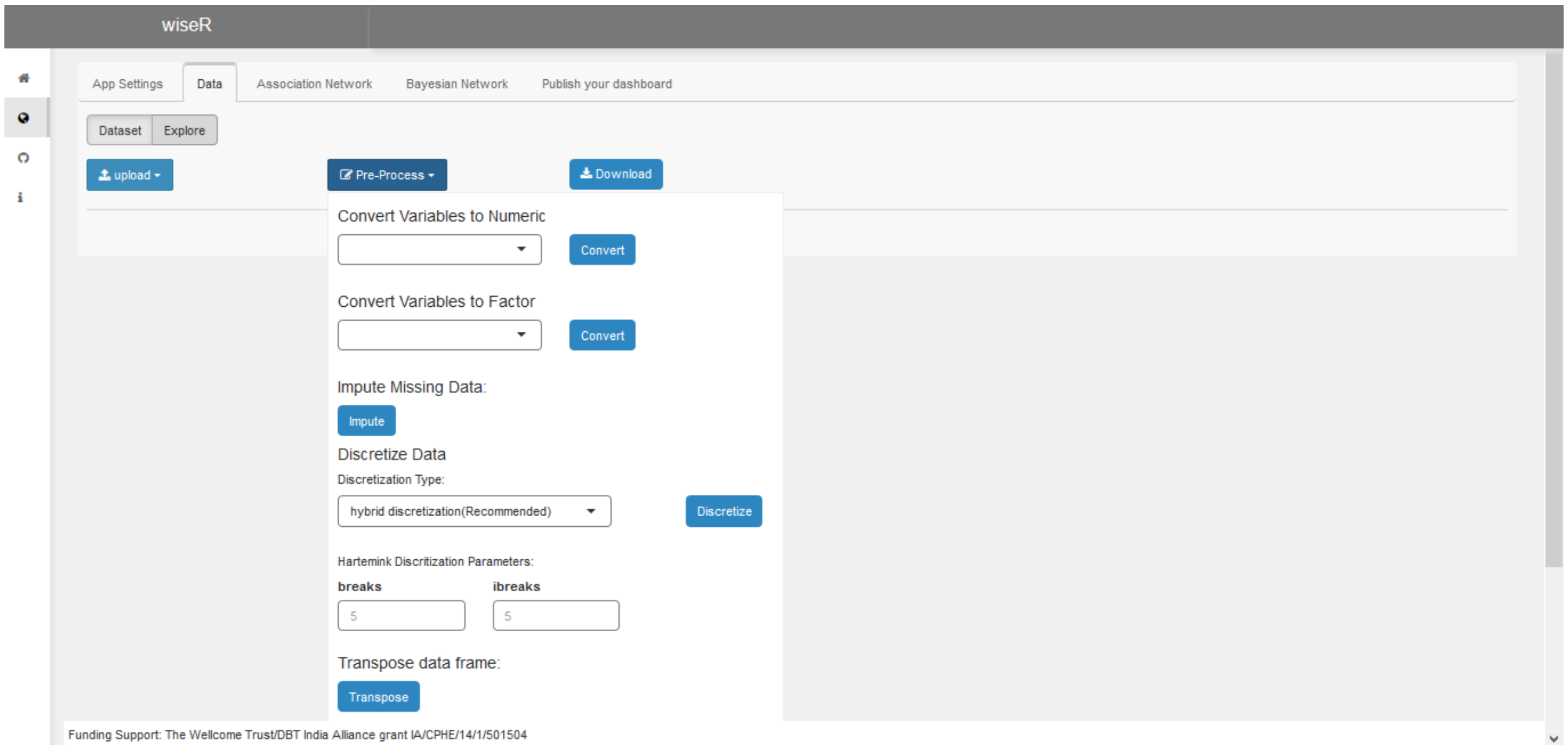

Figure S6. The data pre-processing sub-section of Data tab allows imputation, discretization, type-conversion, variable deletion and transposing and setting of interventional variables.

### 3.5.3 Download

Users can download the pre-processed dataset in current state as a .CSV file

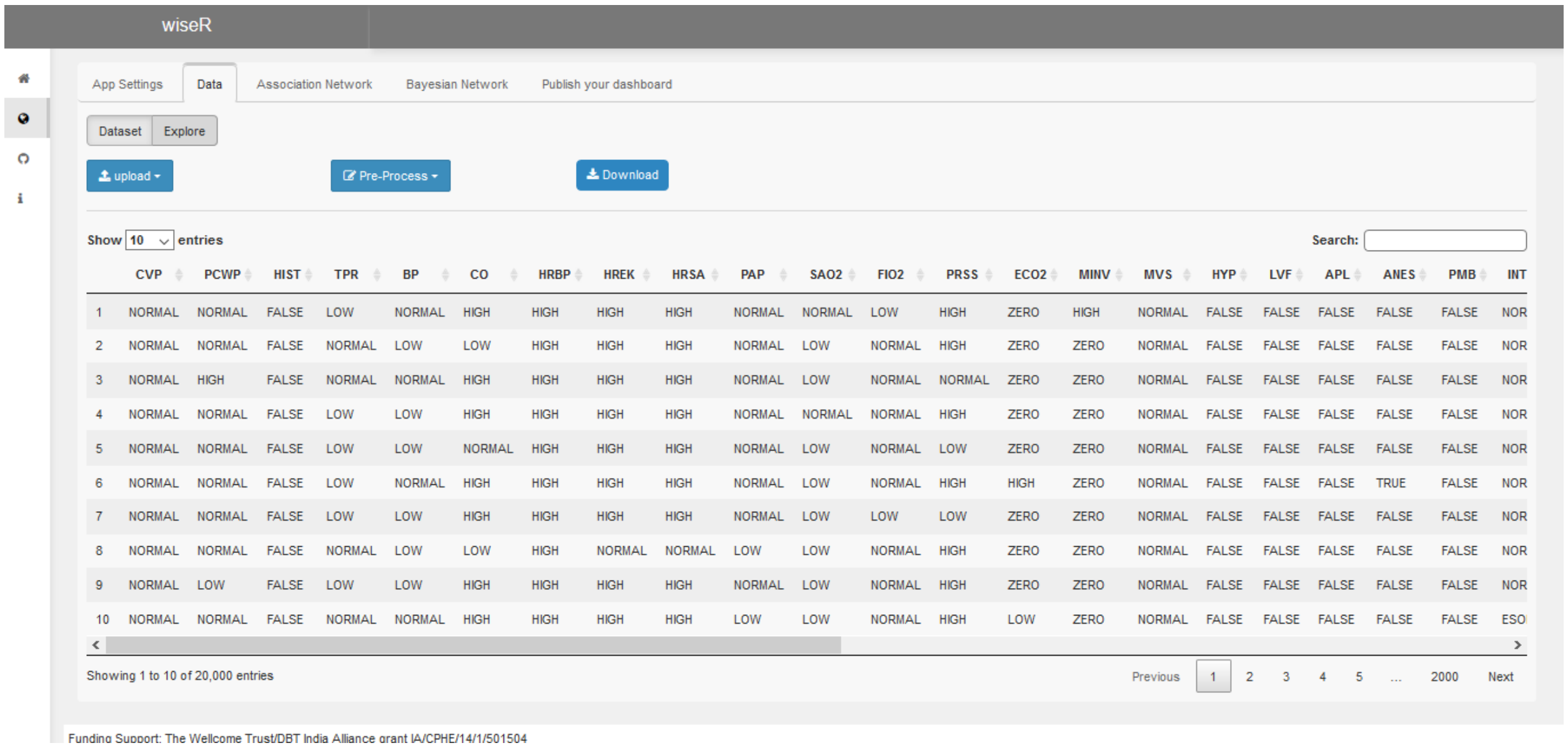

Figure S7. Users can download the pre-processed data as CSV files.

### 3.5.4 Exploratory Data Analysis

Pre-processed data can be explored by looking at their frequency distributions visualized as barplots. This is an important step in the analysis as it allows the users to catch any anomalies that may bias the downstream analysis.

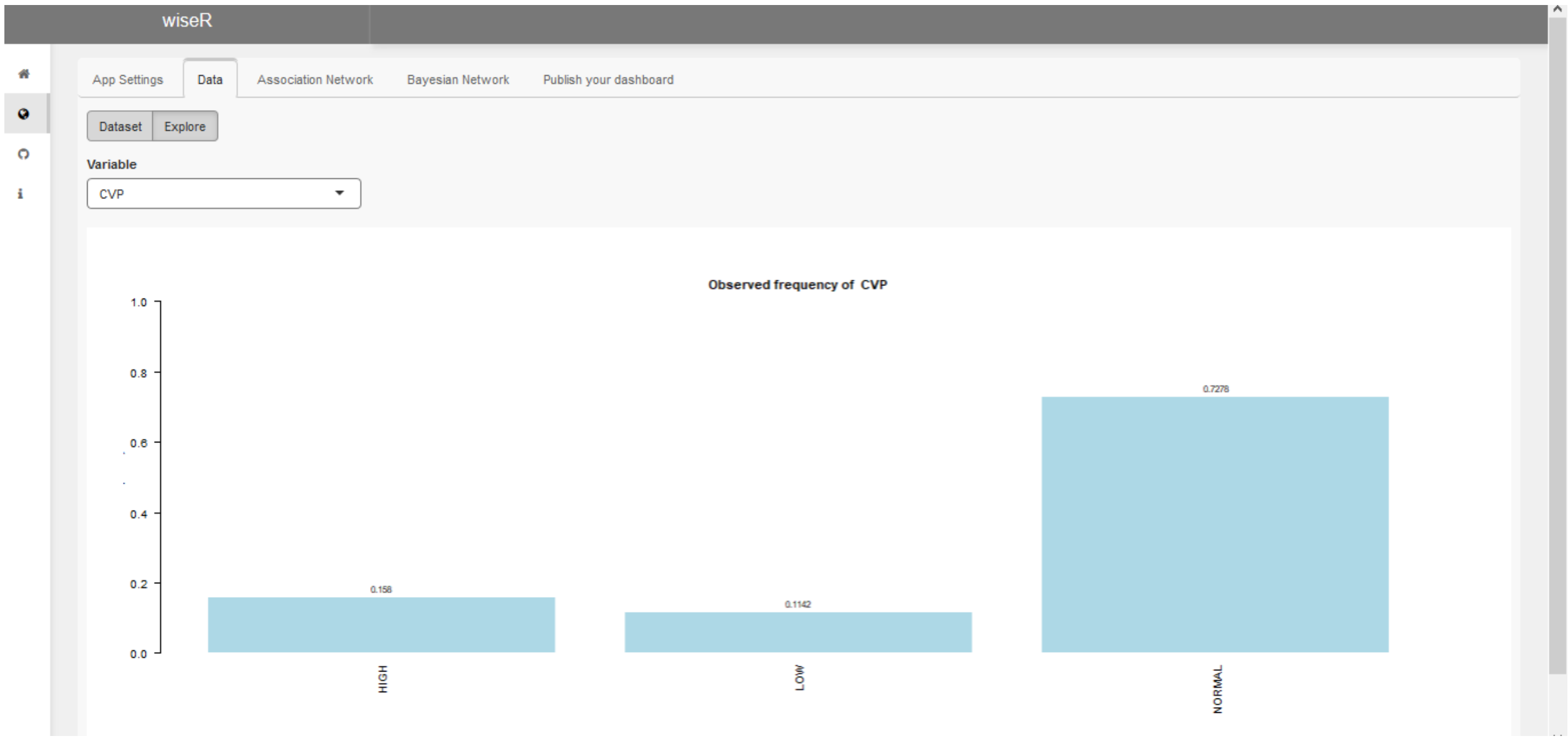

Figure S8. Exploratory data analysis such as barplots of distributions.

## 3.6 Association Network

Before going to Bayesian Network analysis, it is instructive to visualize the association graph constructed upon the data. Learning association networks on the data before Bayesian learning can give useful insights. Additionally the association network can serve as a starting structure for Bayesian Network structure learning. This can be achieved by using the edgelist downloaded from association network analysis in the structure learning part of the app. Various association scores are provided in the app as explained in the next section.

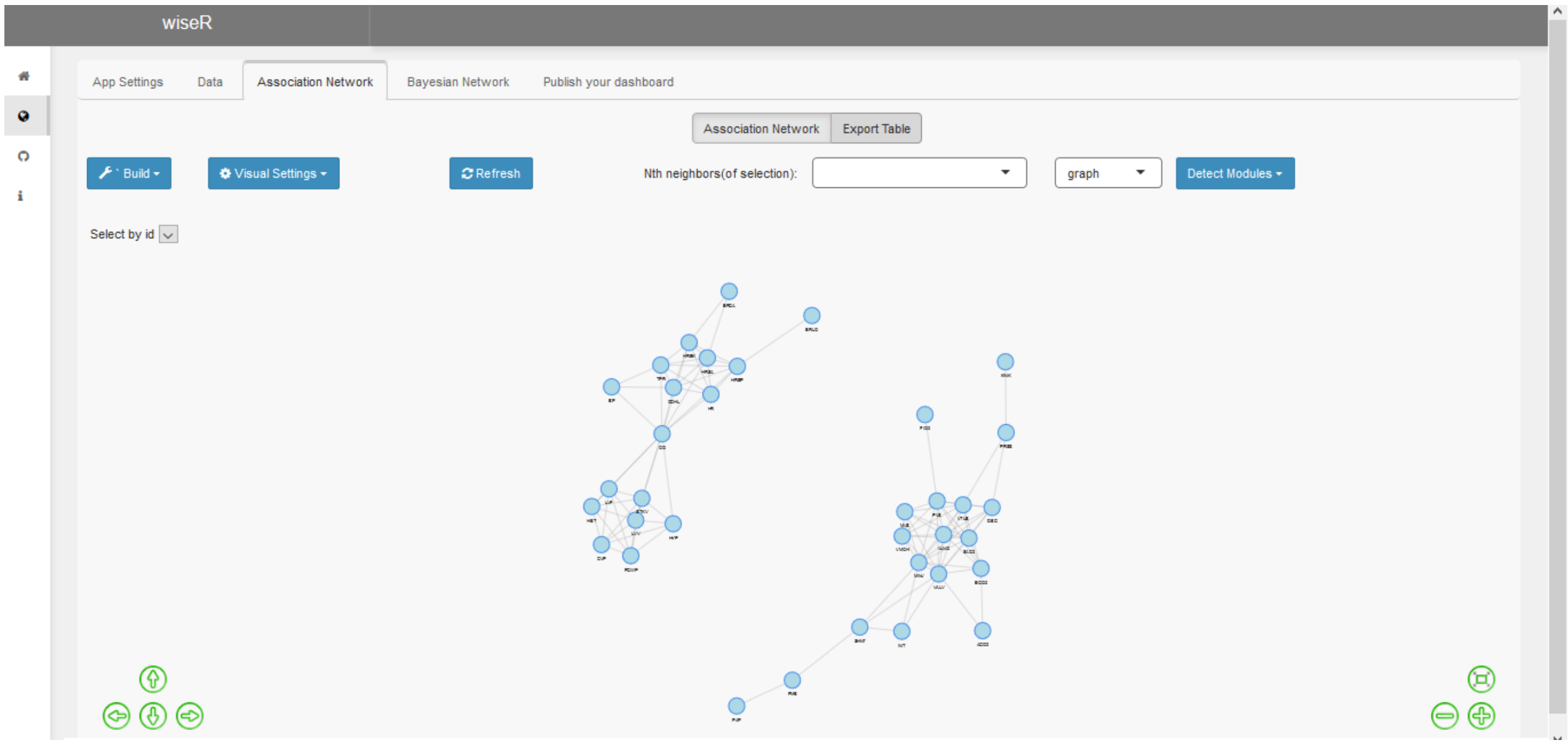

Figure S9. Build and Explore Association networks

### 3.6.1 Build

The build menu is used to build association networks. It consists of different metrics for association strength which are used to build association networks including:

- cramer's V(Recommended)
- cohen's D
- Goodman kruskal lambda
- Tschuprow's T

Each method returns a value between 0 and 1. 0 being two variables are not associated and 1 being highly associated. Users can select a threshold value between 0 and 1 for eliminating edges below the threshold.

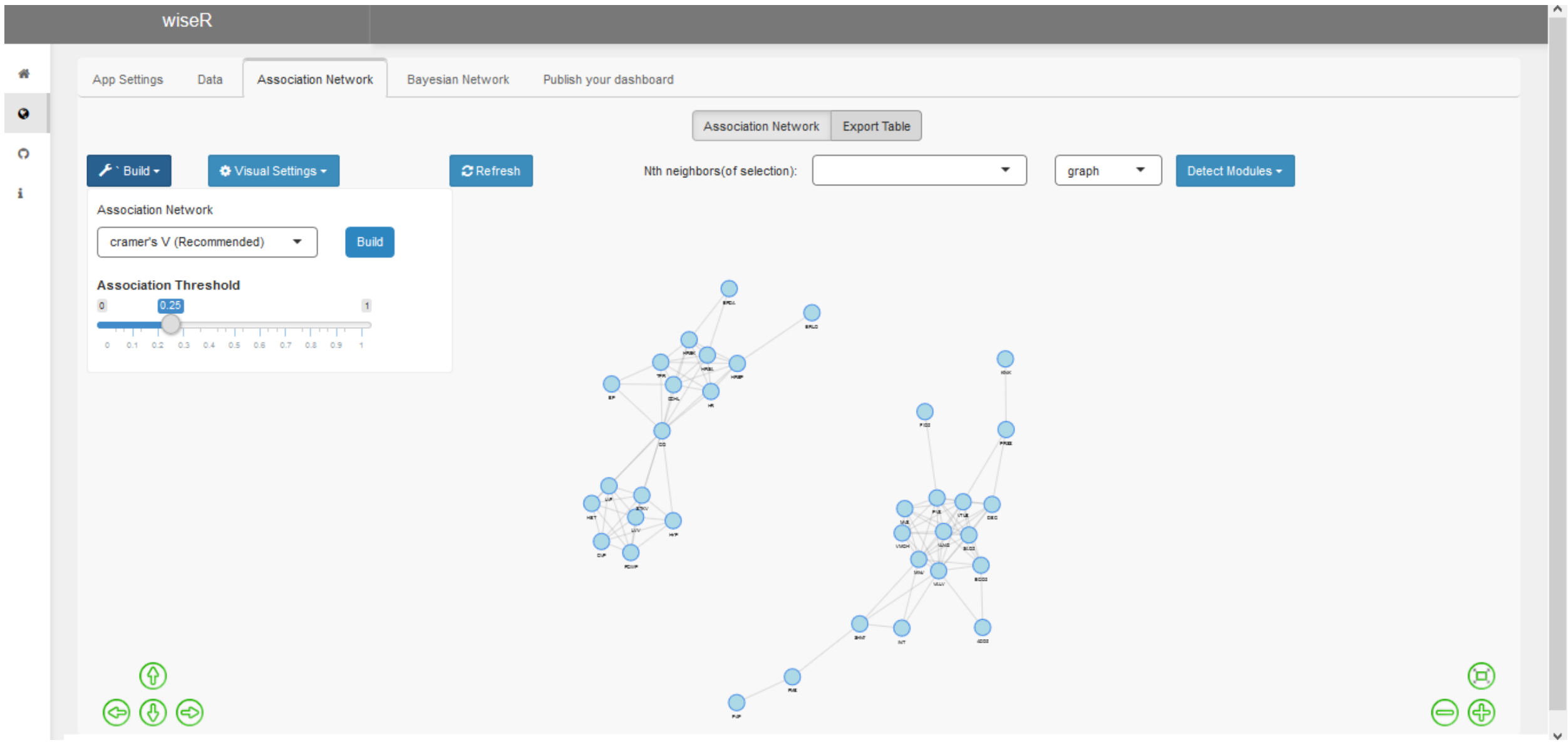

Figure S10. Options to build association networks

### 3.6.2 Detect Modules

Most real world networks have communities. These may form due to preferential attachment to hub nodes, often leading to a scale free network structure. To leverage this phenomenon, the app has a module/community detection pipeline revealing sub-graphs in the original graph. These graphs can be individually explored and analysed from the drop-down menu.

There are different clustering algorithms available in the app for module/community detection which is powered by linkcomm package, like ward.D (Recommended), ward.D2, single, complete, average, mcquitty, median, centroid for users to choose from.

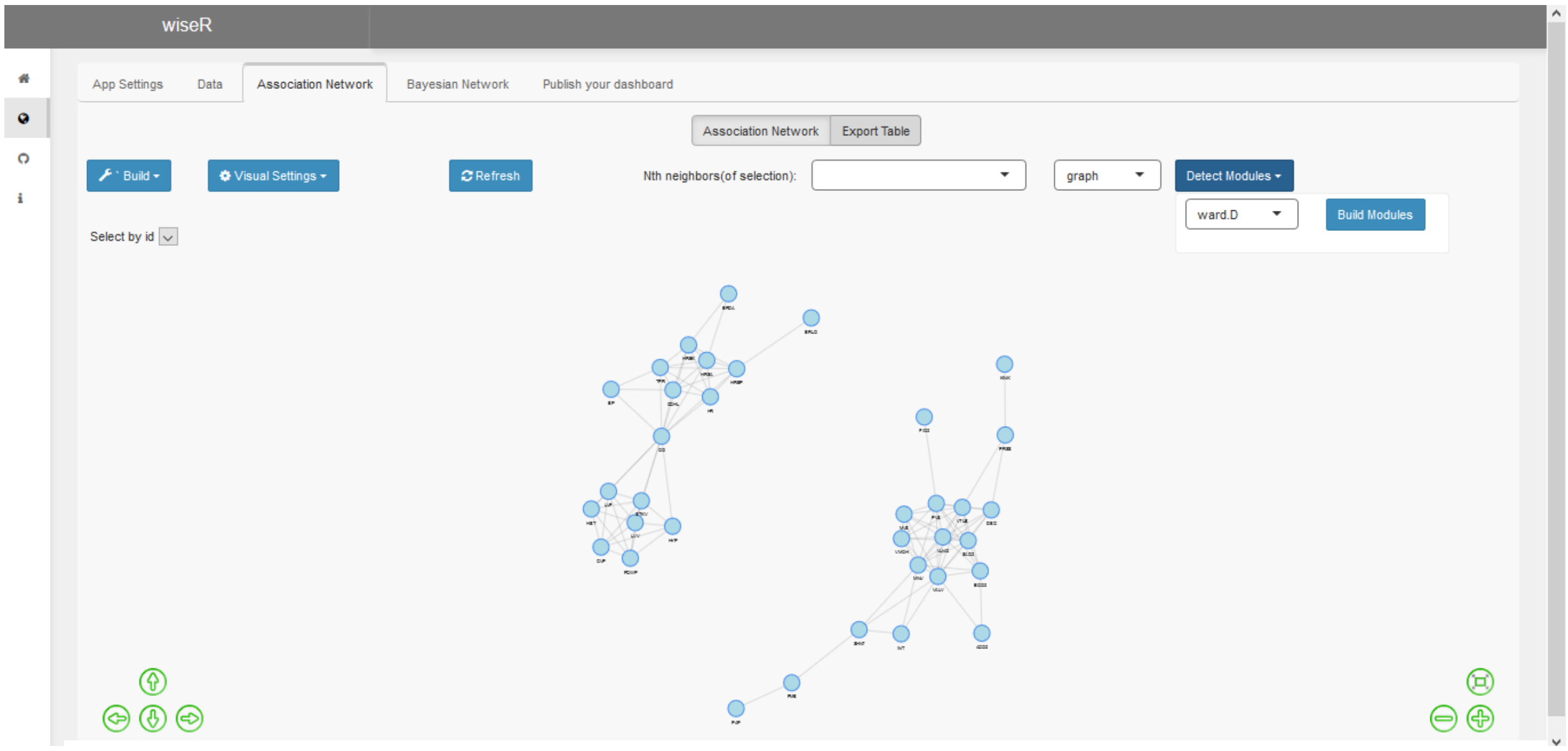

Figure S11. Detection of communities within the association network.

### 3.6.3 Visual settings

Visualization is often key in deriving useful inferences from a complex dataset and these functions are provided through the “Visual Settings” menu.

- **Color/Shape attributes:** The user can either select variable names or provide a vector of variable indices that need to be grouped for node shape and color. These tools are useful in analyzing bipartite or multi-partite networks.

- **n-Nearest Neighbor Visualization:**To ease the exploration of large graphs a threshold for visible neighbor chain is provided, such that clicking on a node only displays its neighbors upto that degree. Alternatively, a user can see the list of n-th degree neighbors by clicking on a variable return. This returns a list of nth degree neighbors. n can be interactively set.

- **Graph layout:** Further, the app allows various graph layout options, some of which are more suitable for particular kinds of data. Useful layouts include tree (for directed), Sugiyama (for directed), star and circle (both for undirected graphs).

- **Font Size:** User can adjust the font size of the node labels to improve visibility

- **Download graph:**Users also have the option to download the network graph in html format from the current tab or using the save network option present in the bottom right corner of the graph.

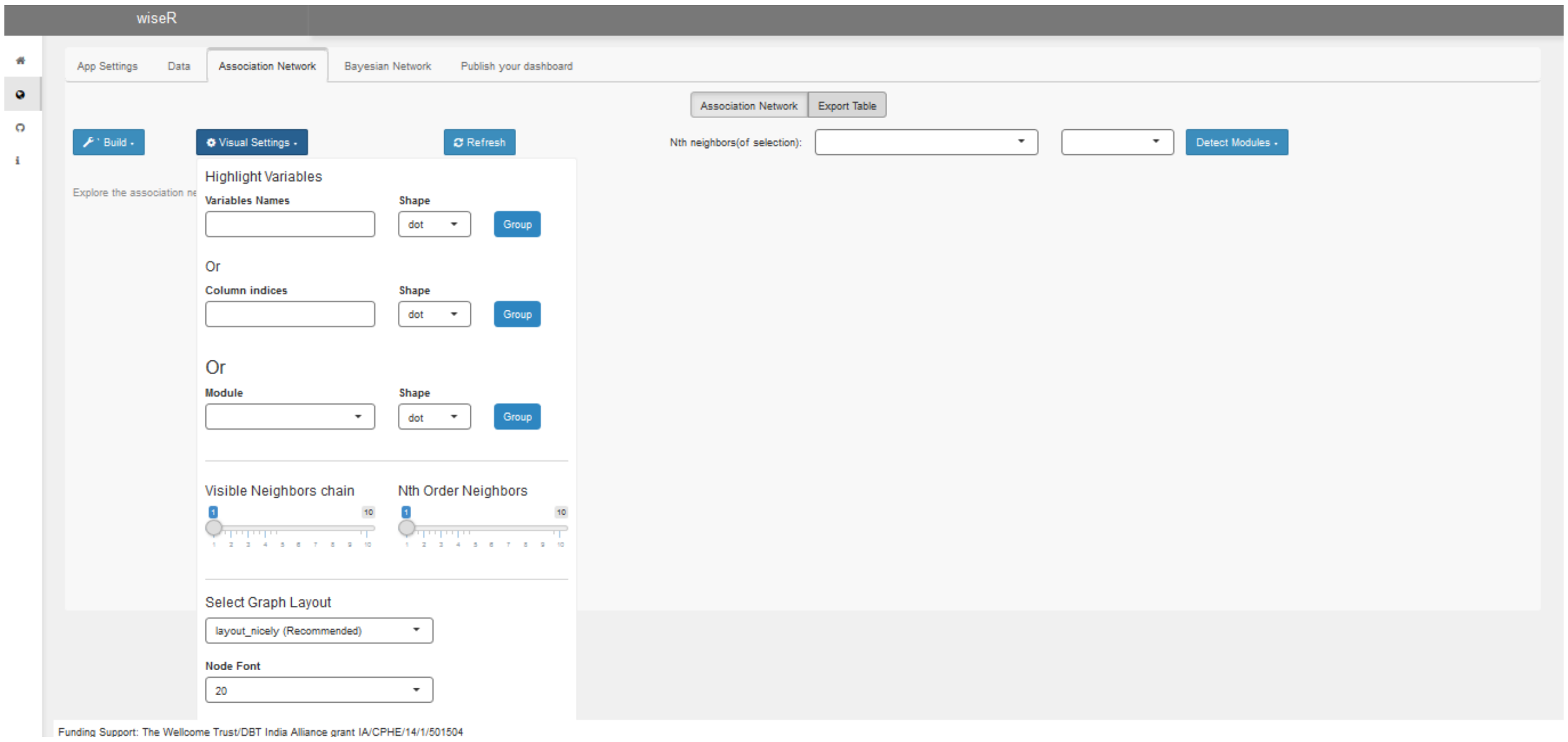

Figure S12. Visual settings to explore association graphs.

### 3.6.4 Export tables

Association networks are essentially a list of edges with corresponding weights. This data obtained from association networks analysis can be exported as an edgelist in tabular form and can also be downloaded as a .CSV file to allow users to switch between different networks analysis tools that can utilize edgelists.

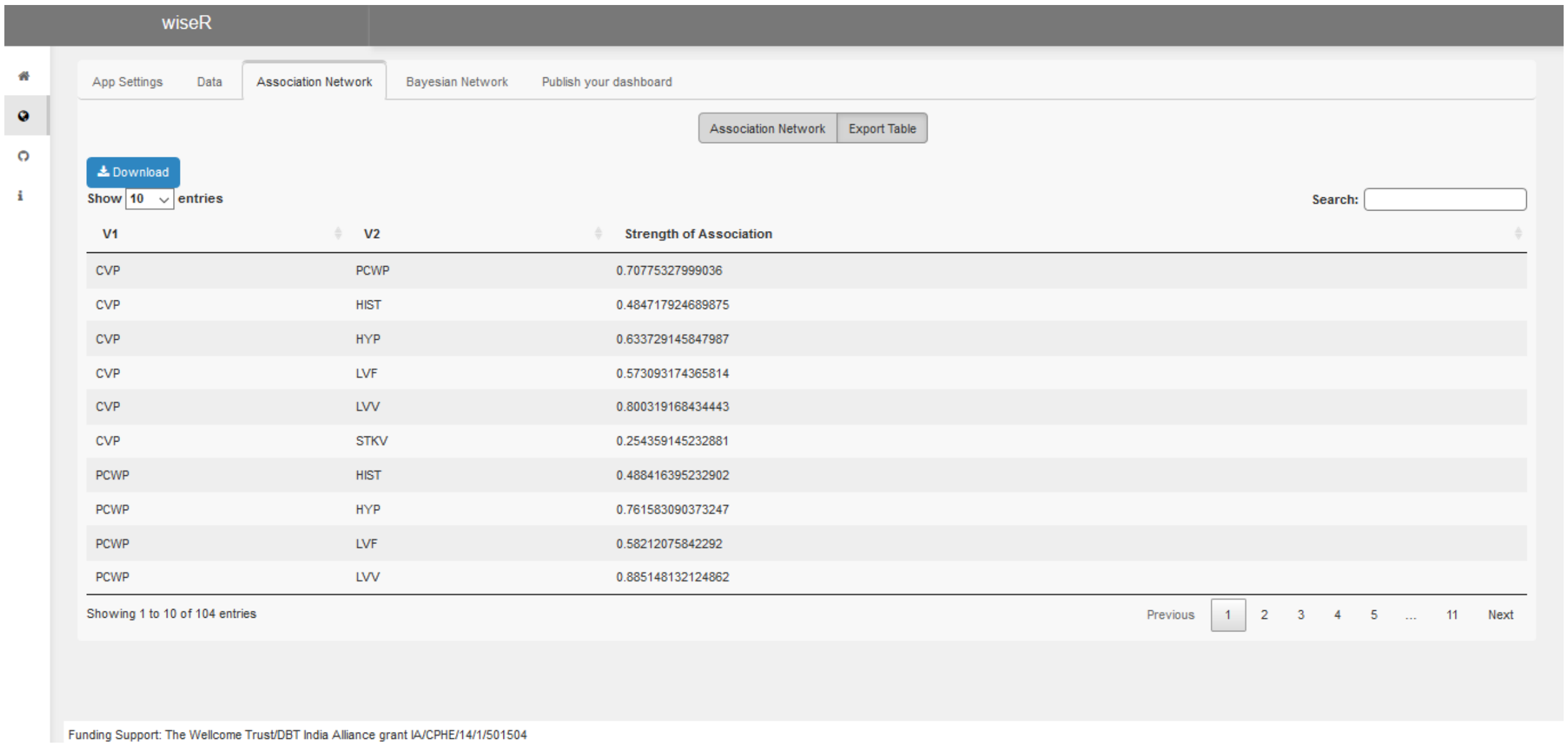

Figure S13. Export the association graph

.

### 3.7 Bayesian Network Analysis

This is the core functionality of the wiseR app. Multiple features have been added to state-of-the-art practices including bootstrapping, ensemble averaging, error bars of approximate inference etc. These ensure robustness of reasoning and decisions derived from the BN.

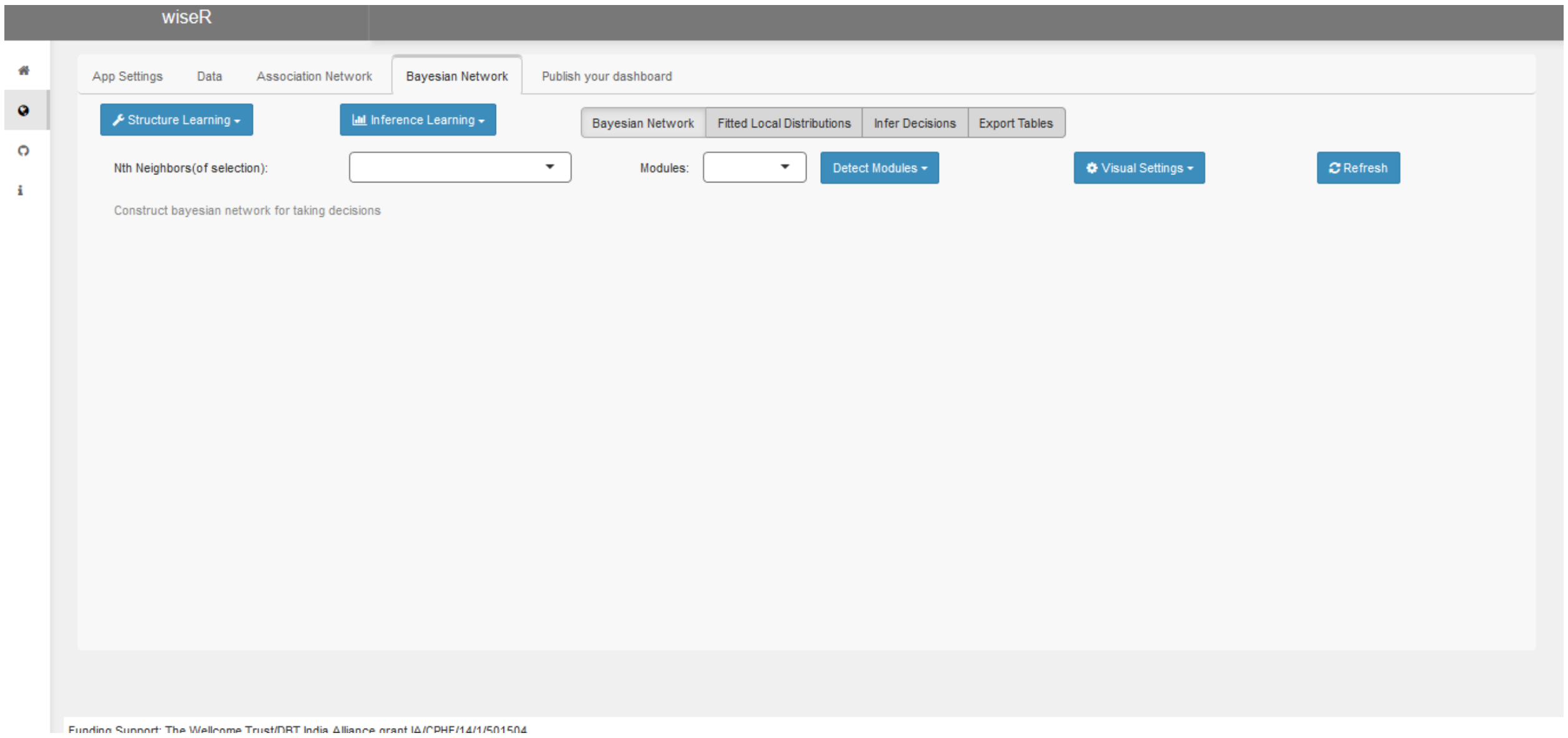

Figure S14. Build and Explore Bayesian Networks

#### 3.7.1 Structure learning

The structure of a graphical model reveals the generative process of the data at hand. For this reason, BNs are interpretable and intuitive. Learning the structure from data is the most challenging yet insightful step in a BN analysis. However, if the structure is already known, or can be guessed, this can be pre-specified by an expert. In this way, BNs represent a perfect hybrid of expert and data-driven learning.

##### 3.7.1.1 Initialize structure

This section relies on expert/prior knowledge and enables the user to upload a network graph as a .CSV file to initialize the bayesian learning graph. The user can also add, remove or reverse the direction of any arc, once the graph has been uploaded. (Only used in score-based learning)

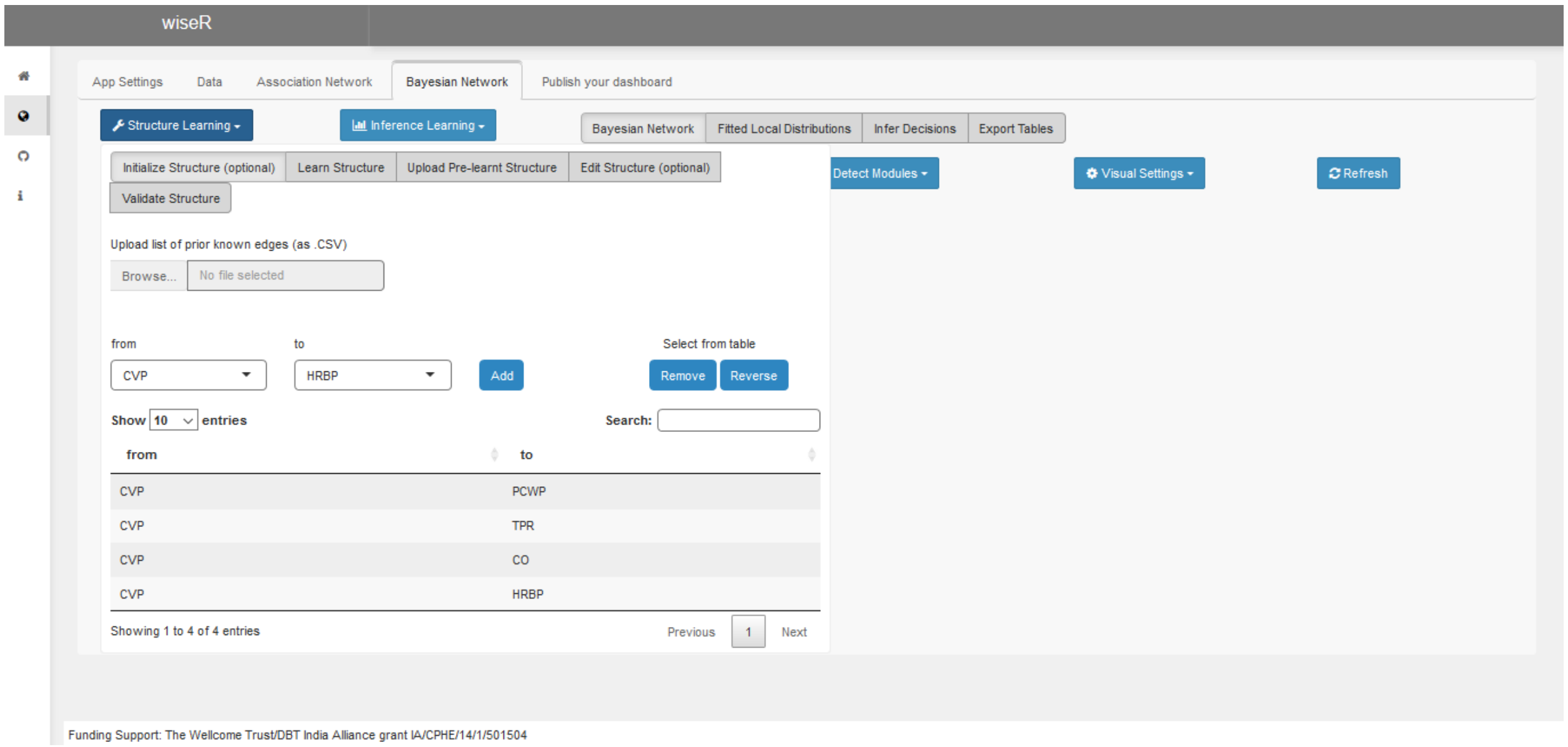

Figure S 15. Graph initialization settings

### 3.7.1.2 Learn Structure

In the absence of expert knowledge, independencies can be directly learnt from the data itself. The computational learning of structure can be done using a number of bayesian learning algorithms such as

- Score-based (Recommended):- Hill Climbing, Tabu
- Constraint-based:- Grow-Shrink, Incremental Association, Fast IAMB, Inter IAMB,PC
- Hybrid Learning:- Max-Min Hill Climbing, 2-phase Restricted Maximization
- Local Discovery:- Max-Min Parents and Children,Semi-Interleaved HITON-PC, ARACNE, Chow-Liu

Options used in structure learning include:-

- Parameter fitting algorithm:- Bayesian parameter estimation (recommended), maximum likelihood parameter estimation
- Network score:- Bayesian Information Criterion, Bayesian Dirichlet Equivalent, modified Bayesian Dirichlet Equivalent, log-likelihood,Akaike Information Criterion,Bayesian Dirichlet Sparse,Locally Averaged Bayesian Dirichlet (Only used in score-based learning)
- Imaginary sample size (Only used in score-based learning)
- Incorporate expert knowledge via blacklisting (explicitly restrict edges in structure learning) and whitelisting (explicitly enforce edges in structure learning) edges by uploading a .CSV file for the same
- Disabling data resampling in bootstrap learning. This is particularly useful for data with low sample size. In this case, the data remains the same but the graph is re-initialized every time. This method is only possible for score based algorithms as it requires random graph initialization.
- Bootstrap iterations:- Bootstrapping means sampling the data with replacement. This parameter specifies the number of bootstrap iterations to run for bootstrapped structure learning.
- Bootstrap sample size:- Proportion of data to be used as sample for bootstrap learning.
- Edge strength:- Set a threshold value between 0 and 1 to remove edges below specified strength from the final structure. Edge strength can be interpreted as the probability of occurrence of an edge in bootstrap learning.

- Edge direction:- Set a threshold value between 0 and 1 to remove edges below specified direction confidence from the final structure. Edge direction can be interpreted as the confidence of direction of a learned arc in bootstrap learning.
- Although not recommended, bootstrap based structure learning may be by-passed via the “Direct Learning” option for learning the structure without bootstraps. Please note that this is suitable only for quick exploration and may give less robust structure.

**note*** The user does not have to do the bootstrap learning every time if they wish to explore different thresholds for edge and direction strength. This can be achieved by using the parameter tuning option.
The final learnt structure can be saved as a bnlearn object in .RData format through the save option

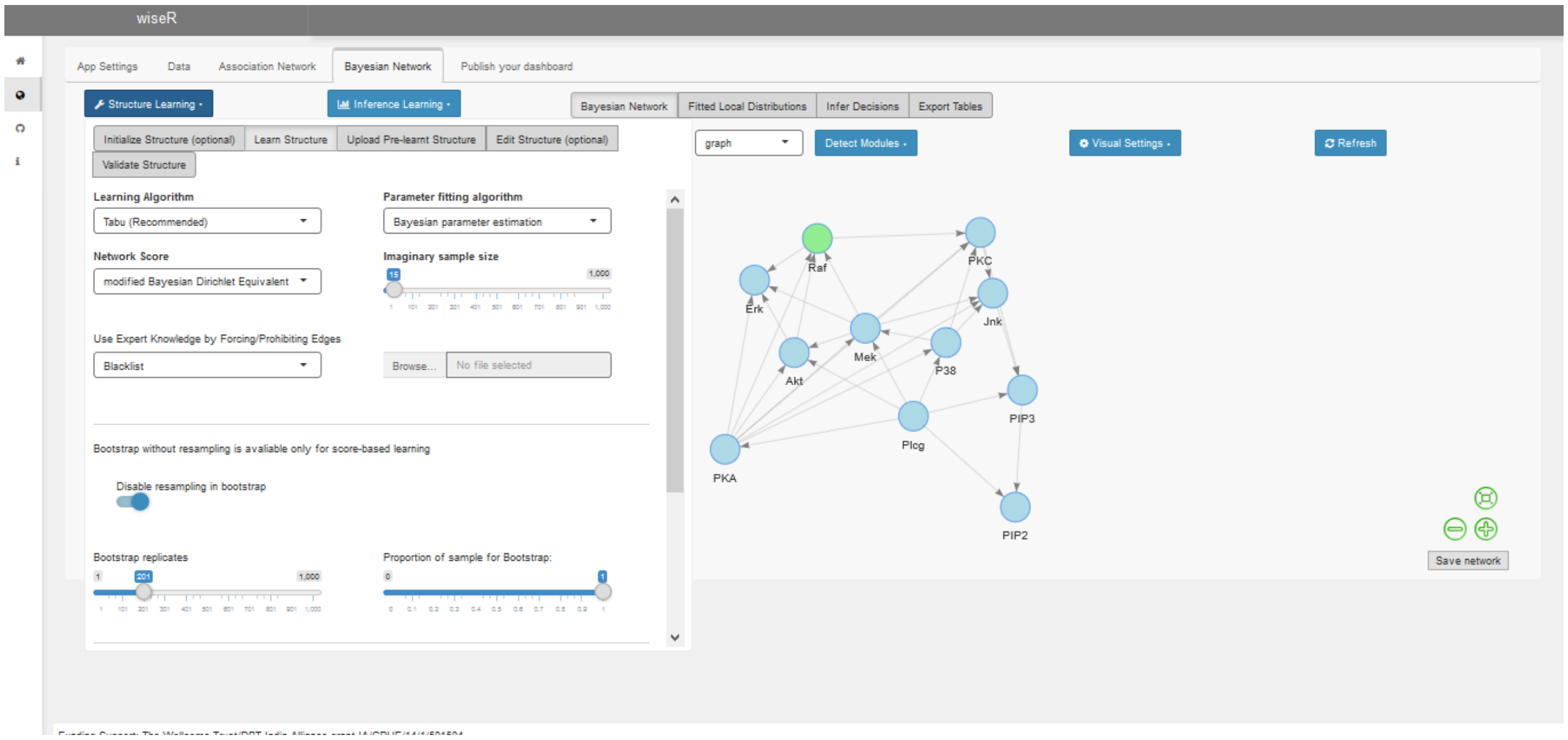

Figure S16. Structure learning settings

### 3.7.1.3 Upload pre-learnt structure

A user returning to analysis should not need to repeat the above steps. An already learnt and saved bayesian network structure can be uploaded as an .RData file.Options to upload both bootstrapped and direct structure and adjust their parameters are available in the app.

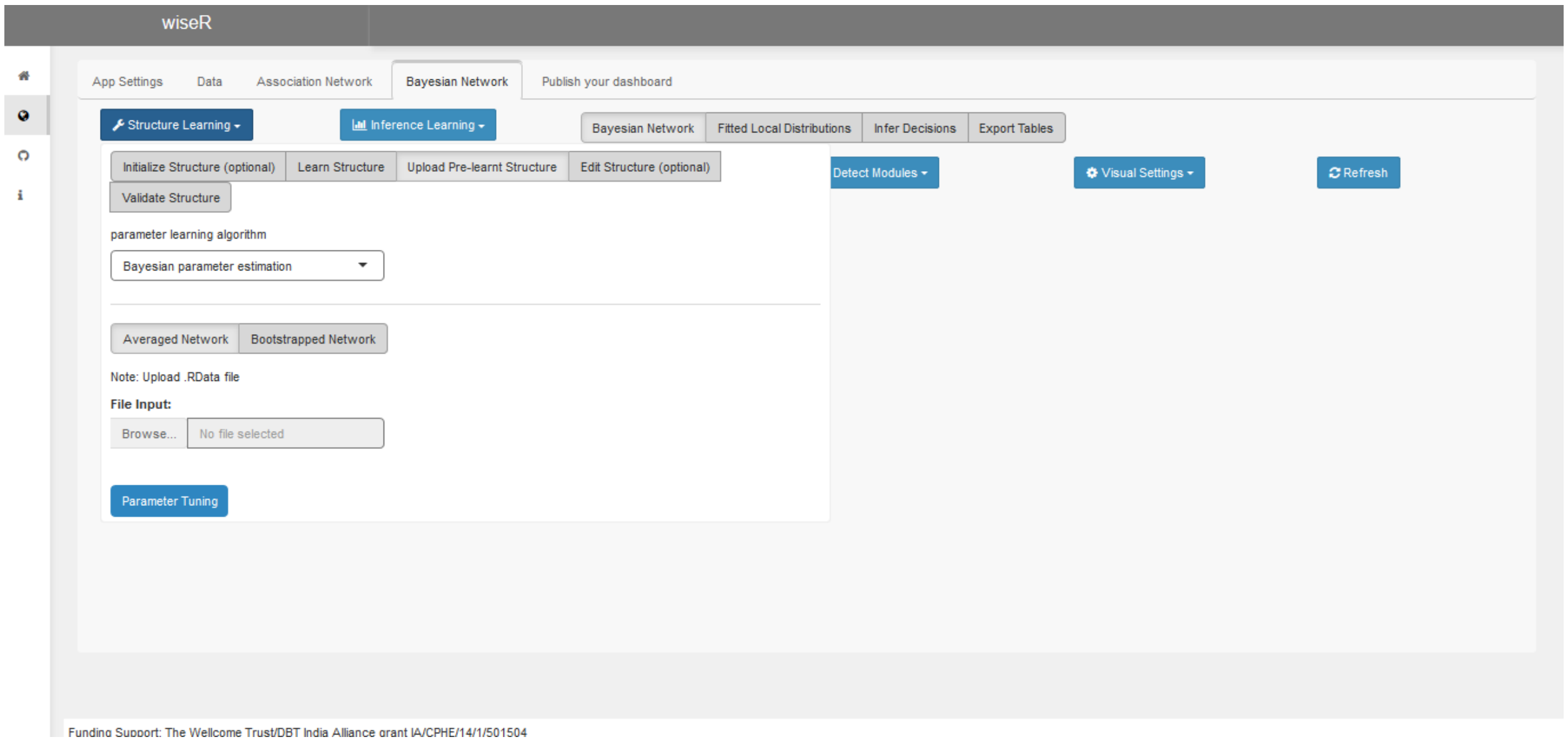

Figure S17. Upload pre-learnt structure

### 3.7.1.4 Edit structure

wiseR allows expert checks at every step and an expert can identify edges and directions that are not plausible in the real world (e.g. smoking cannot influence natural gender, but the reverse can happen). An expert can add/remove arcs, or reverse the direction of the existing ones on the basis of plausibility.

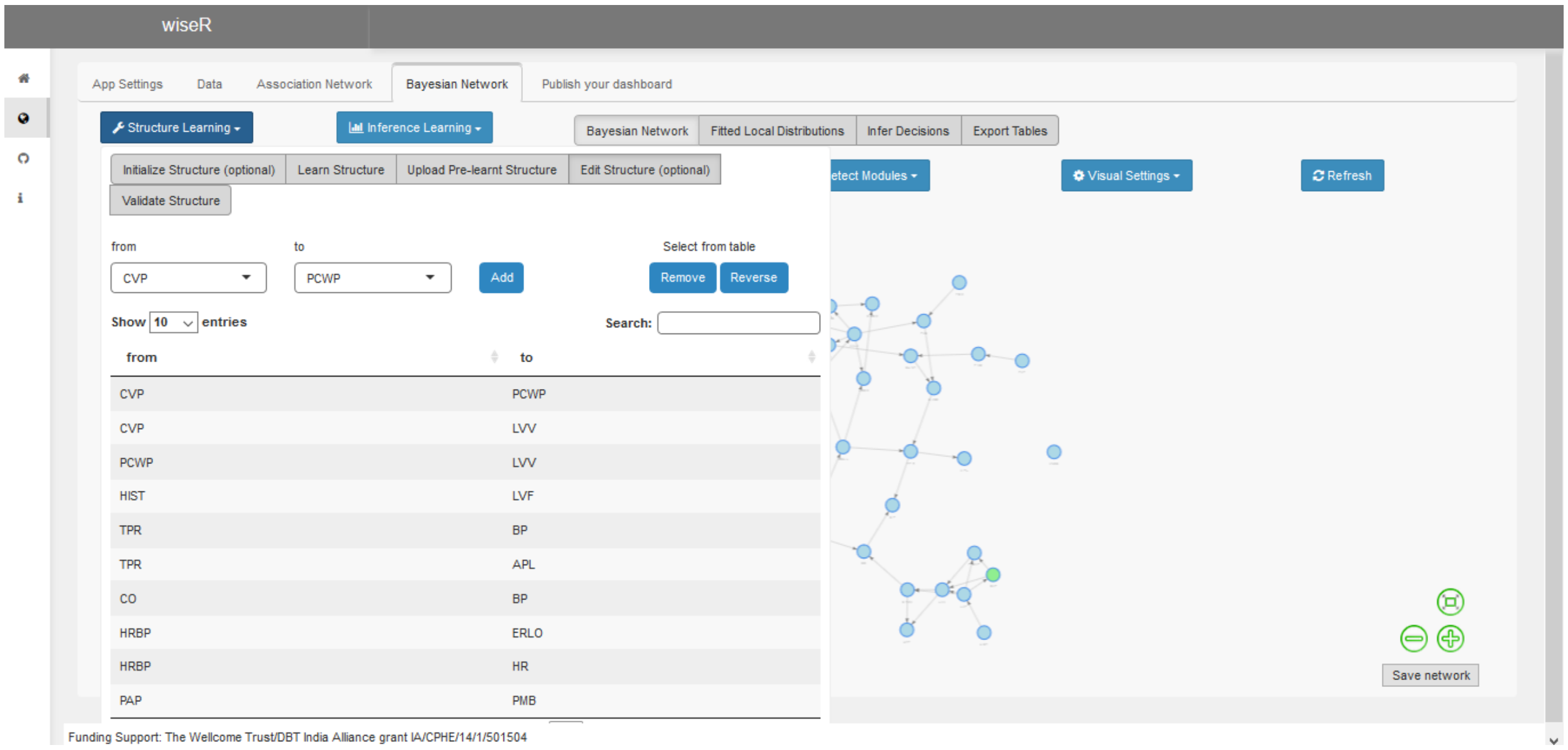

Figure S18. Edit learned structure via adding, removing or reversing directions based upon expert knowledge and plausibility.

### 3.7.1.5 External Graph

wiseR allows users to work with an external graph through the app. It is possible users may want to test out newer graph learning algorithms which are not available through the app. This ensures the user access to app features provided they have a edgelist of the external graph in “from” and “to” structure saved as a “.csv”

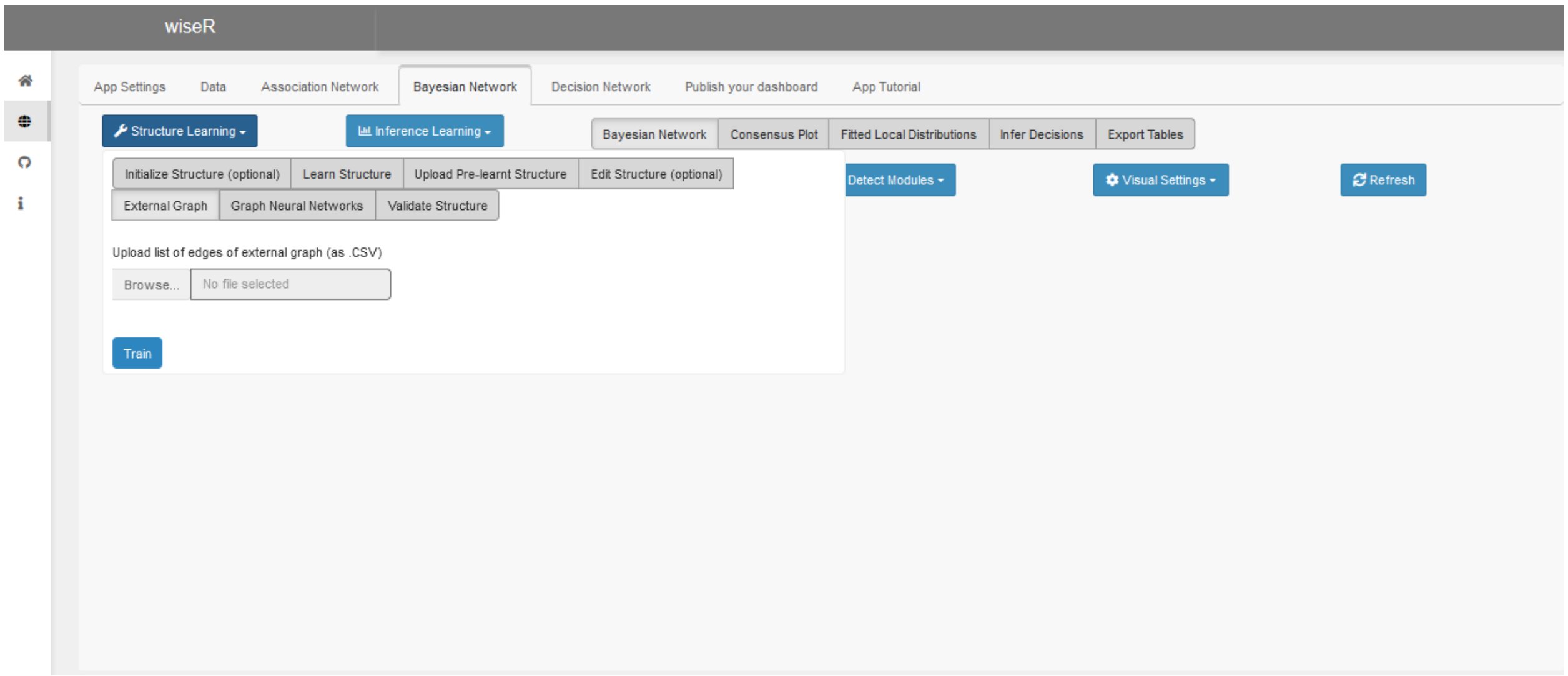

Figure S19. Provision for uploading externally learned graphs to work through the App.

### 3.7.1.6 Graph Neural Networks

With the current advances in Graph based machine learning, wiser provides the standard GNN based structure learning method through the app. The user is asked to provide the Python and Conda environment path with all requisite libraries such as:

- pytorch
- matplotlib
- networkx
- pandas
- scipy
- scikit-learn
- argparse
- Causalnex
- Ipython

The user can provide the no. of bootstraps for averaged structure learning.User can choose between DAG-GNN (recommended) and NoTears algorithm for structure learning.

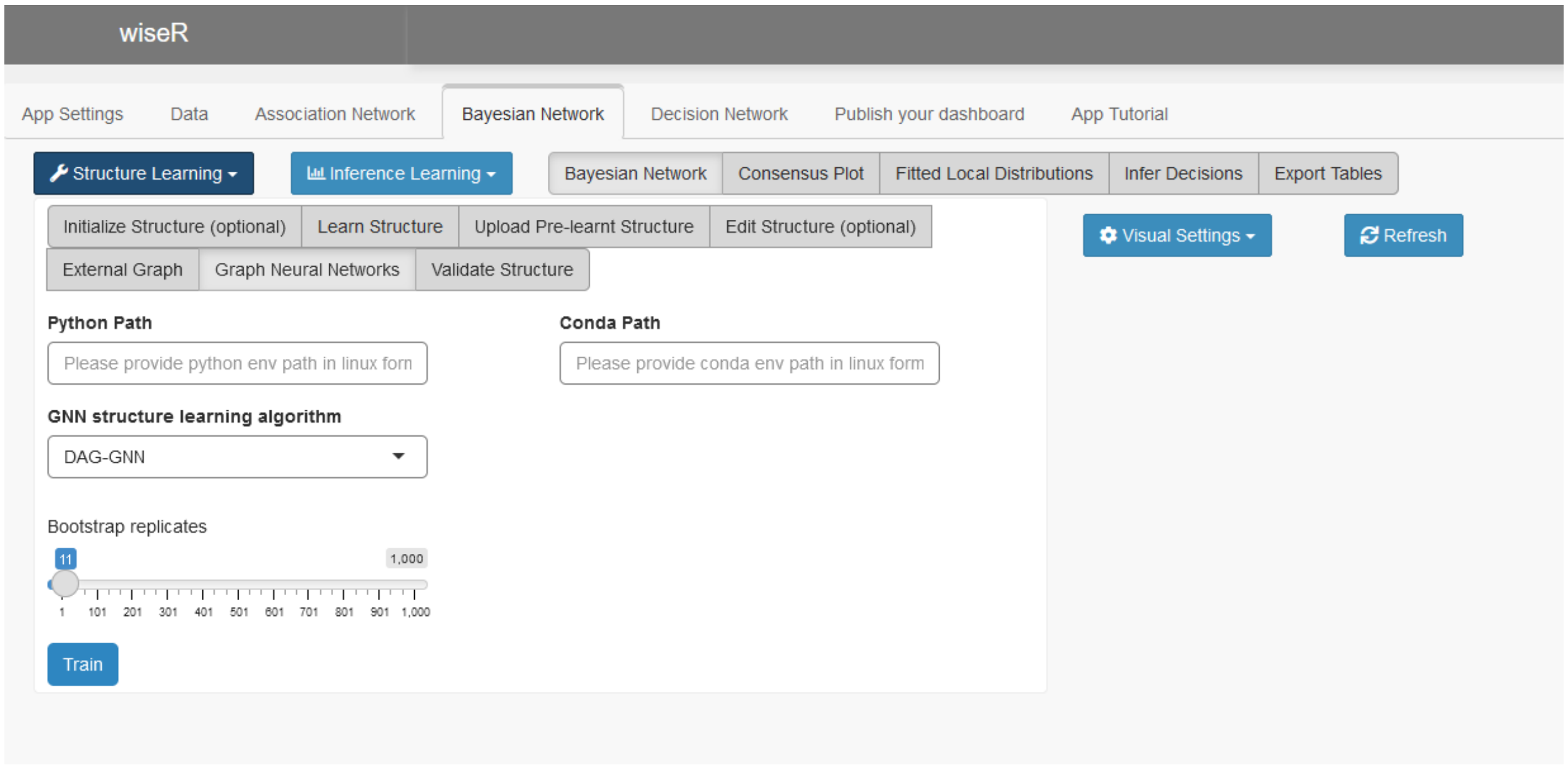

Figure S20. GNN based structure learning through the App.

### 3.7.1.7 Validate structure

This section enables the analyst to validate the learnt structure using 10-fold cross-validation or hold-out testing. The log-likelihood loss and network score outputs produced can then be used to judge the usability of the learned structure.

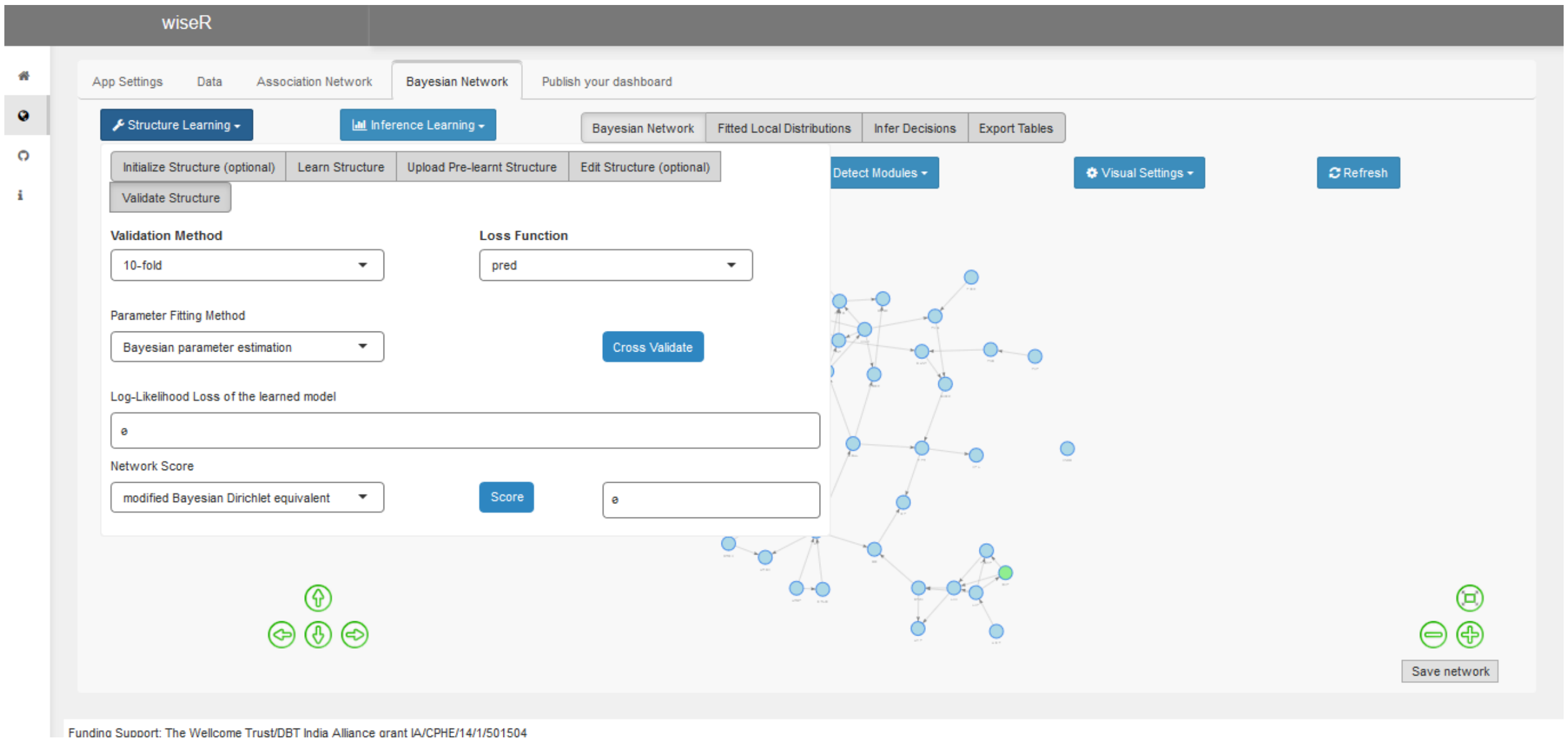

Figure S21. Validation of the learnt structure through cross-validation and hold-out testing.

### 3.7.2 Inference

After the parametrization of BN is complete, it can be queried to derive inferences and reason about the data. The inference learning menu enables the user to learn and explore conditional probability plots on the learned structure. The user can set an event and multiple pieces of evidence can be inserted as per the query. This crucial feature enables the user to explore probability plots on event nodes in the network conditional on chain of evidence, which is the essence of decision making through Bayesian networks.

Inferences can be learnt through two methods using wiseR

Approximate Inference (Fast, Monte Carlo sampling of local structure). This option is useful for large datasets where exact inference is intractable. However, the sampling is implemented every time, hence adding incrementally to computational overhead over time. Also it yields slightly different results on each iteration. The app has the feature to re-compute approximate inferences for a specified number of times and to plot error bars in the inference.
Exact inference (Fast for small networks, slow for large networks, one-time computation)
Assuming that the User is working with large datasets, approximate inference is the default setting. This can be switched in the app. It is observed that the inference plots produced using approximate inference with error bars, is almost as accurate as exact inference, hence can enable robust reasoning and decisions on large datasets where exact inference is not possible.

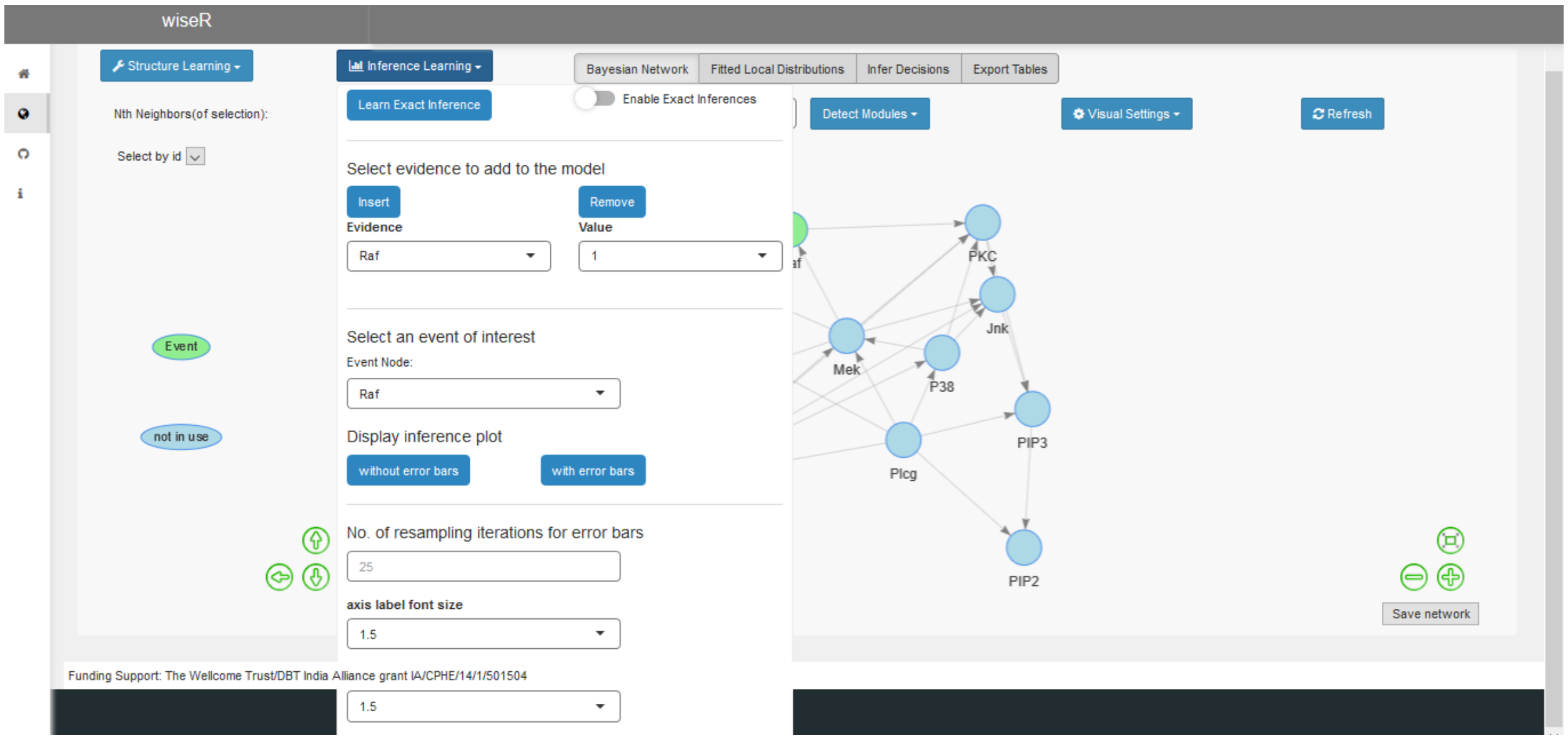

Figure S22. BN Inference. Approximate, Exact and Robustness.

Apart from structure and inference learning which are the workhorse of a BN analysis, the bayesian network tab allows options for visualization and exploration

- Bayesian network panel is used to explore the learned structure and is equipped with all the graph exploration features as available in association graph

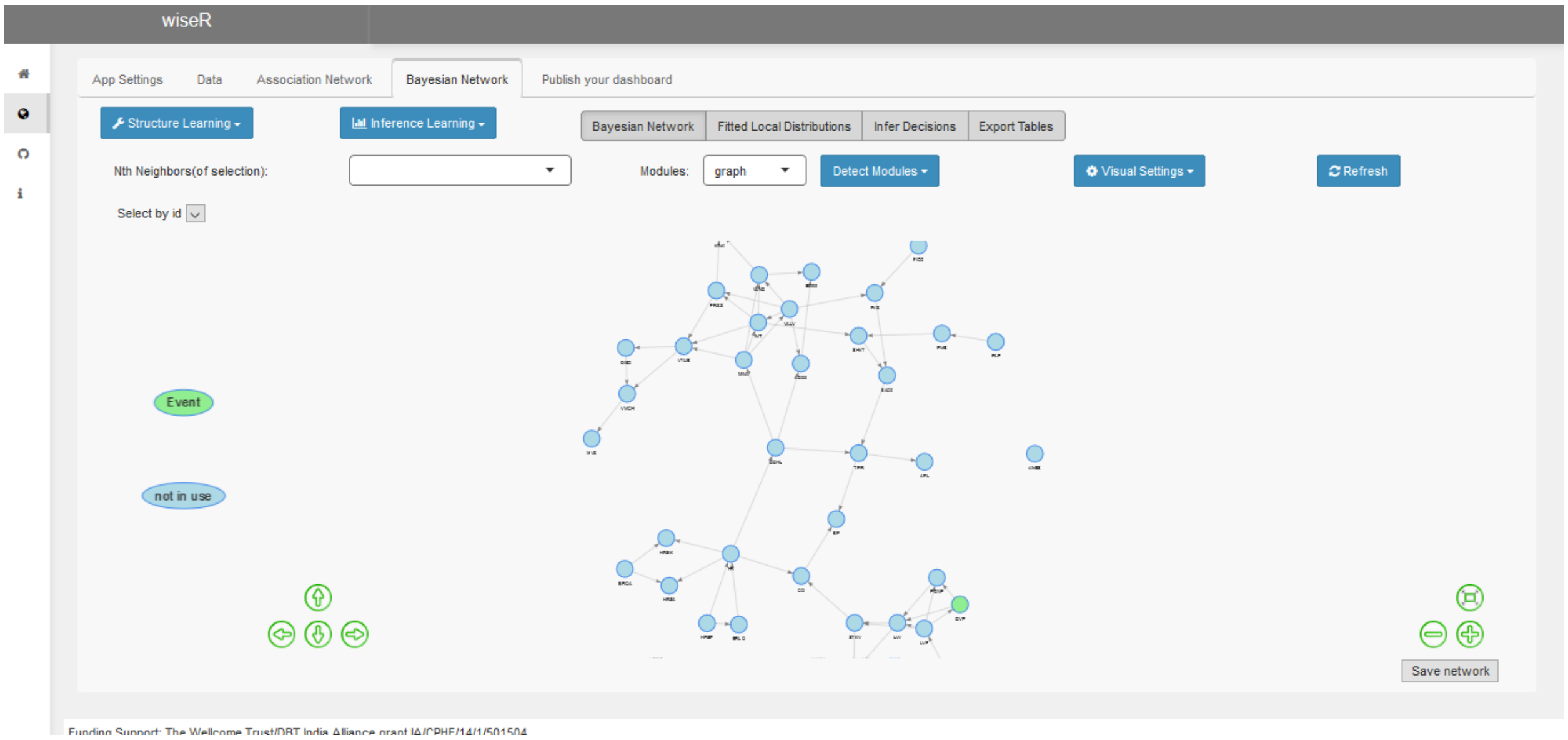

Figure S23. Example of a learnt bayesian graph

- Fitted local distribution panel is used to explore the local probability distribution tables of variables in the learned network structure

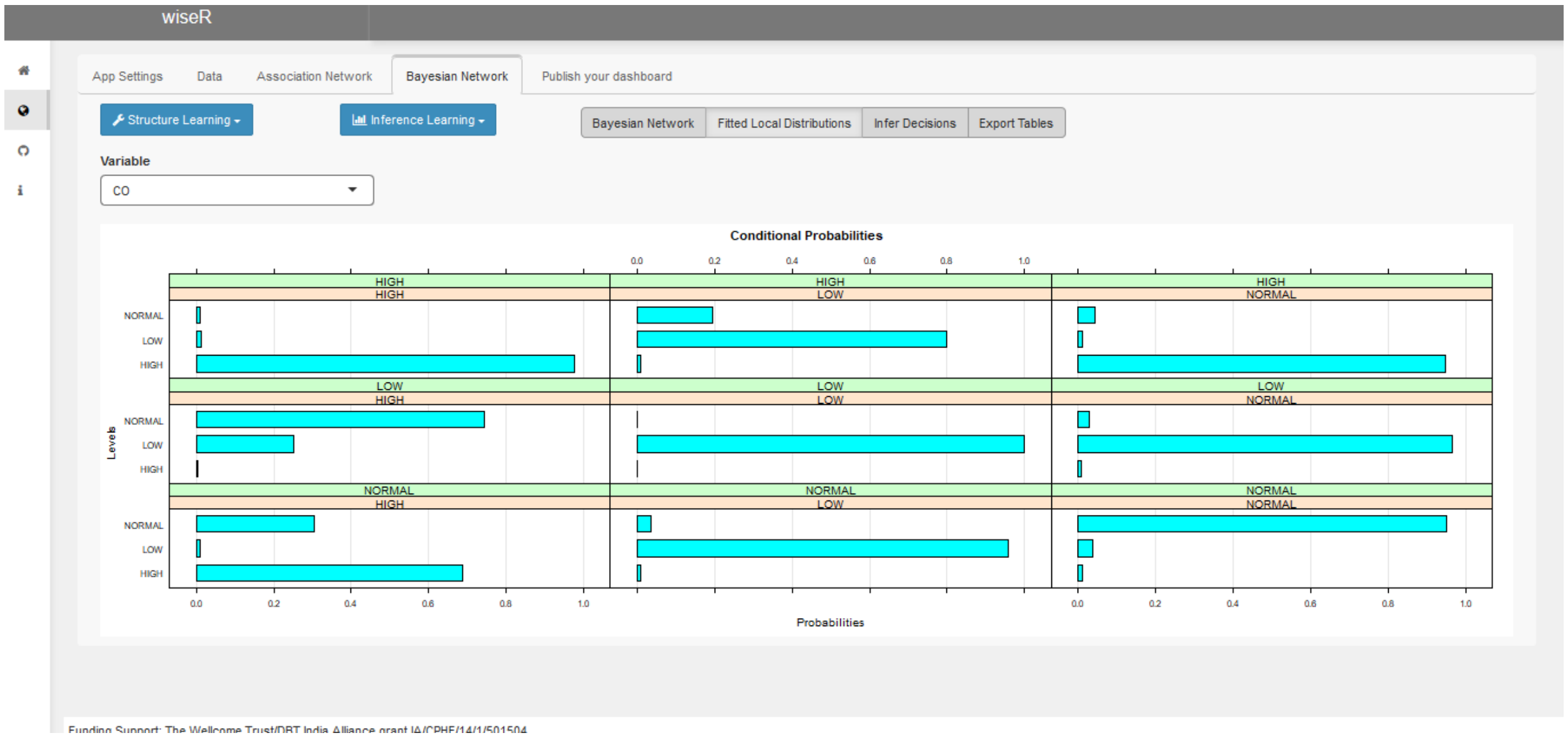

Figure S24. Local probability distribution tables

- Infer decisions are used to visualize the inference plots once the user has chosen the event and evidence nodes. It also has option to sort the plot axis and prune the no. of plot bars

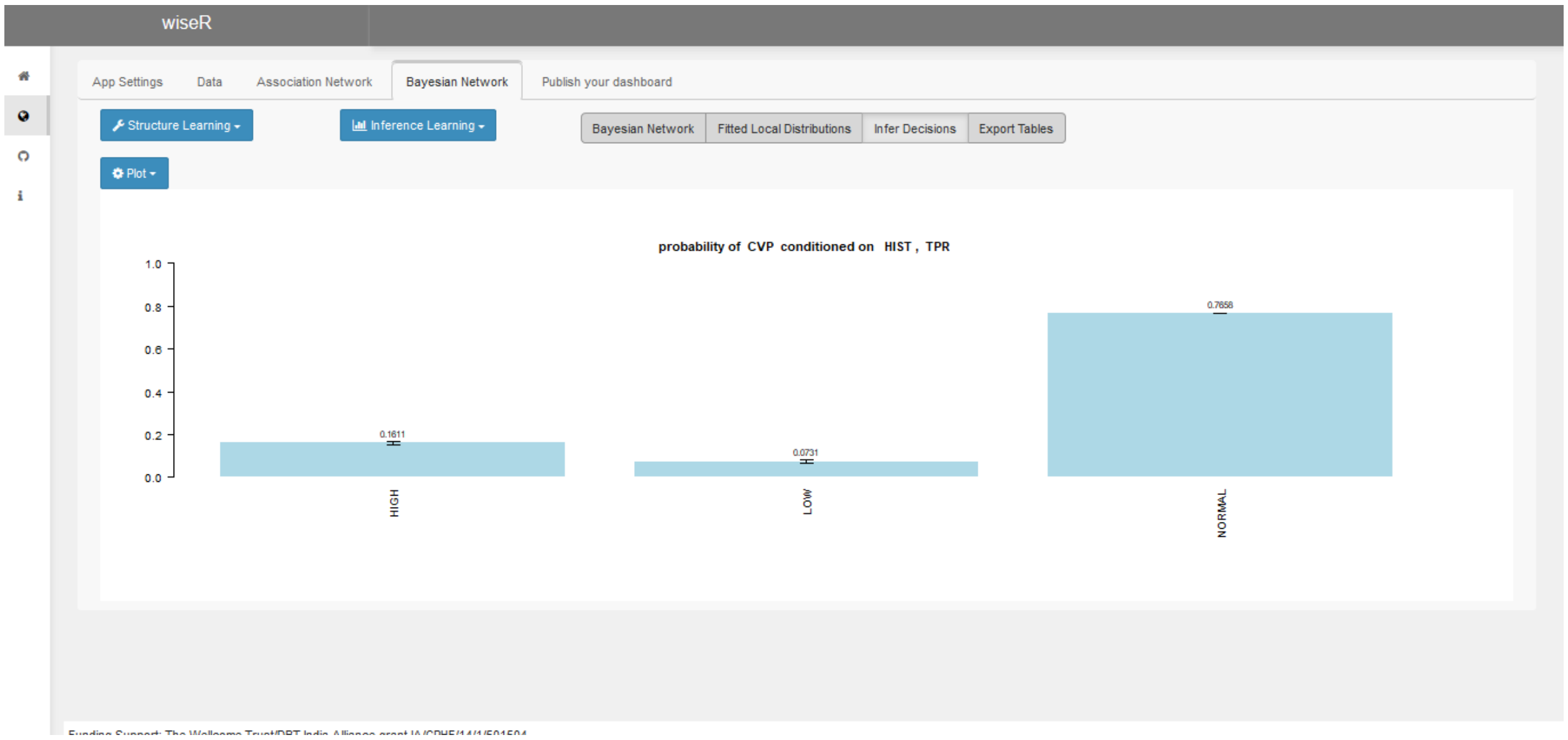

Figure S25. Inference plot

- Export tables section is used to visualize the network graph, blacklist-whitelist edges and variable wise validation results in tabular format. User can also download these tables as .CSV file

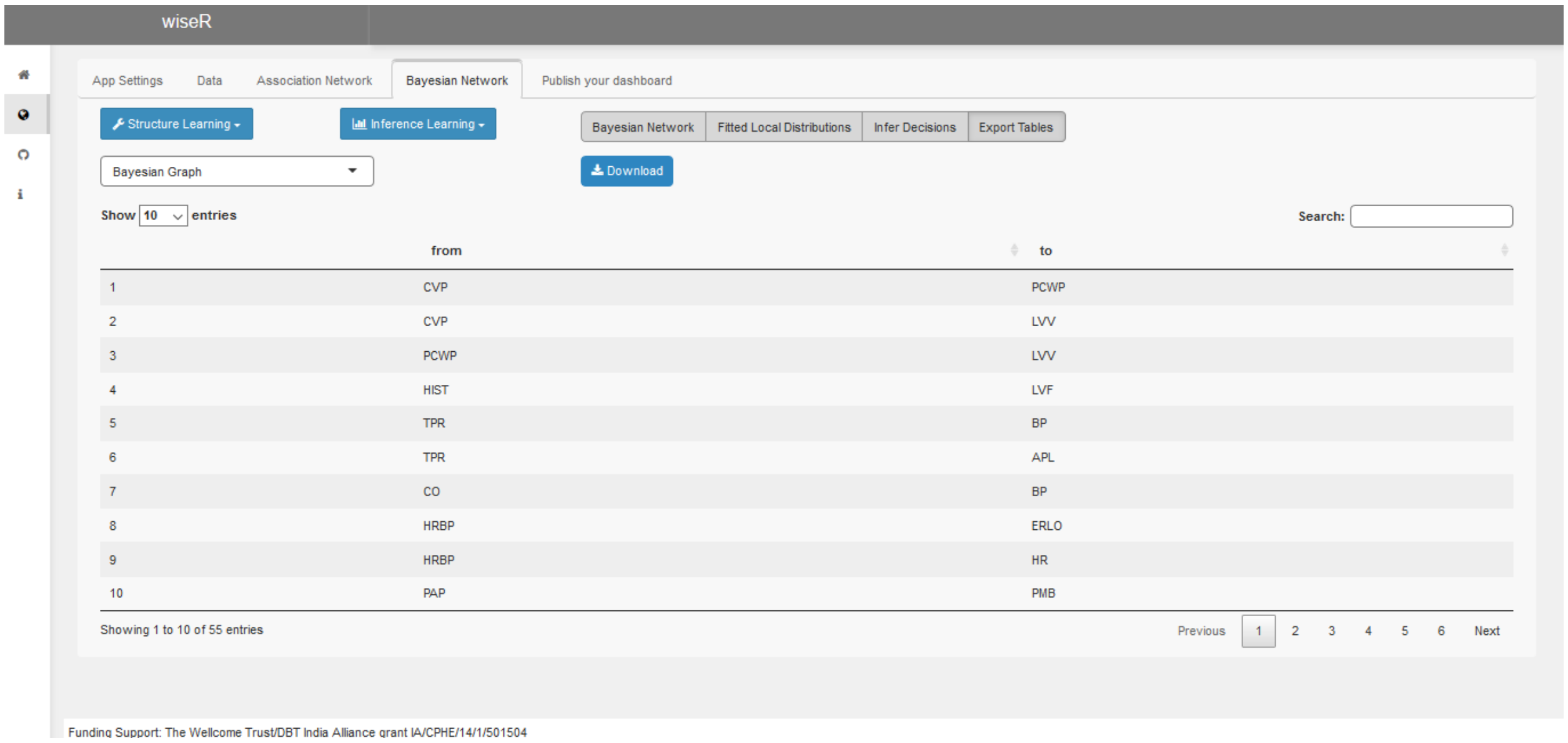

Figure S26. Visualization and tables available to be exported from the application.

### 3.8 Taking Decisions with the learnt network

Heads-up: This section relies upon JAGS installation on their system. JAGS can be downloaded from its official page.

As per our knowledge, there is no available open-source tool or a package in R that integrates structure learning with decision making. Influence diagrams have typically been defined by hand by experts. wiseR integrates these two capabilities to help the analyst arrive at the best (or next-best) policy for optimizing a payoff. For this purpose, the use can create a payoff node based on the variable that they wish to optimize (known as the utility node, e.g. Blood

Pressure optimization in the Intensive Care Unit). This brings up the table to enter the utility of each state in the range of -1 to 1. For example, a user may assign a negative utility to an abnormally low or high blood pressure event (-1) and a positive (+1) utility to normal blood pressure. Alternatively, if the user does not care as much about the high blood pressure as they would care about low blood pressure (e.g. shock in the ICU), they can assign a zero utility to the high blood pressure event. After this the user sets the decision nodes that are direct parents of the node to their utility node. If the policy from parents is expected to be obvious, they can go to grandparents or orders higher, however the effects may diminish. Finally the user runs the Monte-carlo simulation that displays the best policy along with the payoffs in a sorted table. This leap from making subjective inferences to policy advice makes wiseR a white-box approach to actionable, transparent and interpretable machine learning. A complete example is available towards the end of the tutorial for better understanding.

### 3.9 Publish your dashboard

Finally, the user can abstract the learning and decision making process into their own dashboard deployments. We consider this section to be a key highlight of wiseR. Publishing the analysis as an interactive dashboard helps reproducibility and effective communication. wiseR enables the user to produce a custom dashboard of their results as an R package.

The user has options to select the name and theme for the dashboard, and select the build option. wiseR builds the dashboard as a downloadable R/Shiny package which can be served as a web-application or launched as a local Shiny app. The custom dashboard is equipped with the most useful features of wiseR such as the interactive graph exploration and inference learning and visualization.

This feature saves authors the hassle of creating a dashboard for their findings, and helps propagate research output to the community in an efficient manner, thus helping bridging the gap between the research and actual use.

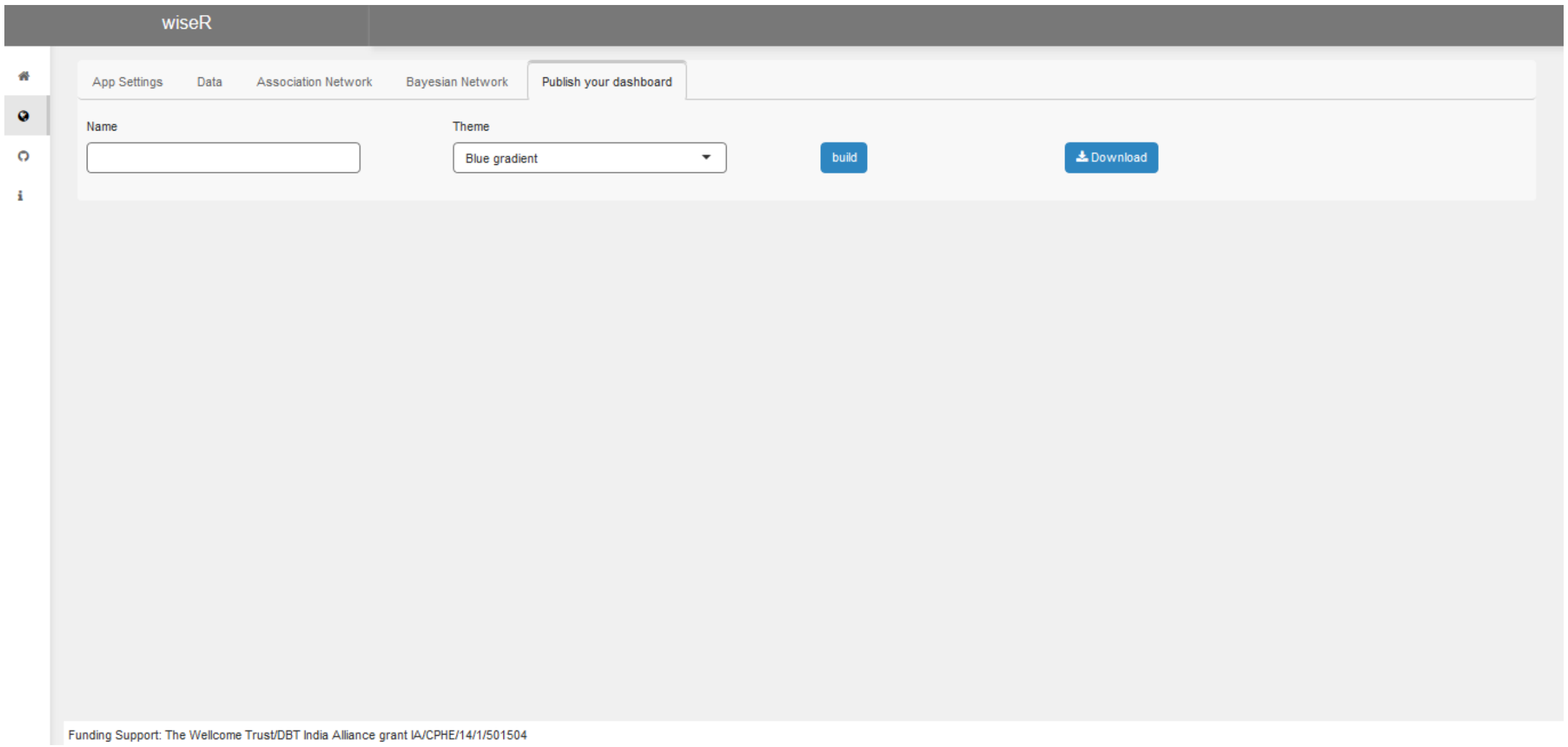

Figure S27. Publish your dashboard as a web-application or a local package

## 4 Comparison

- Open source (OS): Is the source code available ? If yes, what programming language?
- Target Audience (TA) :What is the target audience? Based on the source code availability and programming language used, the app can be targeted to end users, data scientists or both.
- Enabled Statistical/ML extensions (MLE) : Does the backend language support AI/ML libraries for customizations in those directions?
- Structure Learning (SL): Is structure learning possible?
- Bootstrapped Learning (BL): Is robust bootstrap structure learning possible?
- Interventional Data(INT): Can it handle interventional data and incorporate it in structure learning?
- Cross Validation(CV): Does the app provide means for cross validation and evaluation of learned models?
- Parameter learning(PL): is parameter learning possible using the app?
- Informed Layout(IL): is network visualization using a wide variety of efficient and targeted graph layouts for exploration ?
- Community Detection(CD) : are subgraphs based on communities/modules detected in the graph available for visualization and exploration?
- Inference with Confidence (IC): Can the app perform inference learning with confidence intervals ?
- Chained Inferences(CI): can the app handle multi-evidence inference learning?
- Exact Inference(EI): can the app perform exact inference learning?
- Approximate Inference with error bars(AIE): can the app perform much faster approximate inferences, verified over multiple iterations and produce final results with error bars(Much faster and equally accurate than exact inference on large datasets)?
- Decision Networks (DN): Does the app learn decision networks for policy optimization on specific utility nodes ?
- Graph Neural Network Structure (GNN'S): Does the graph have provision for GNN based structure Learning?
- Parallel processing(PP) : Does the app enable the user to run the process in parallel for faster runtime?
- Free: Is the app free for public use?

A comparison of wiseR and other structure learning and bayesian network applications

| Name | OS | TA | MLE | SL | BL | INT | CV | PL | IL | CD | IC | CI | EI | AIE | DN | GNNL | PP | Free |
|---|---|---|---|---|---|---|---|---|---|---|---|---|---|---|---|---|---|---|
| wiseR | Yes (R) | Data Scientist,End user | Y | Y | Y | Y | Y | Y | Y | Y | Y | Y | Y | Y | Y | Y | Y | Y |
| Bayesian Networks | Yes (R) | Data Scientist, End user | Y | Y | N | N | N | Y | N | N | N | N | N | N | N | N | N | Y |
| BayesiaLab | No | End User | N | Y | N | N | N | Y | N | N | N | Y | Y | N | N | N | N | N |
| Bayes Server | No | End User | N | Y | N | N | N | Y | N | N | N | Y | Y | N | N | N | N | N |
| BNJ | Yes (Java) | End User | N | N | N | N | N | Y | N | N | N | N | Y | N | N | N | N | Y |
| CIspace | Yes | End User | N | N | N | N | N | N | N | N | N | N | Y | N | N | N | N | Y |
| | (Java) | | | | | | | | | | | | | | | | | |
| OpenMarkov | Yes (Java) | End User | N | Y | N | N | N | Y | N | N | N | Y | Y | N | N | N | N | Y |
| Sam Iam | No | End User | N | Y | N | N | N | N | N | N | N | N | Y | N | N | N | N | Y |
| UnBBayes | Yes (Java) | End User | N | Y | N | N | N | Y | N | N | N | Y | Y | N | N | N | N | Y |

## 5 Example-1: Bayesian Inference

This section provides a hands-on example to conduct a step by step analysis to replicate one of the landmark results obtained using data-driven Bayesian Network learning and Inference, the Sachs dataset on yeast pathway modeling presented in Sachs et al. (2005), which utilizes bayesian learning on complex interventional data.

### 5.1 Steps

We upload the sachs interventional data which is in a space separated text file into the app. Variables are read as factors as it is a numerically encoded discrete data.

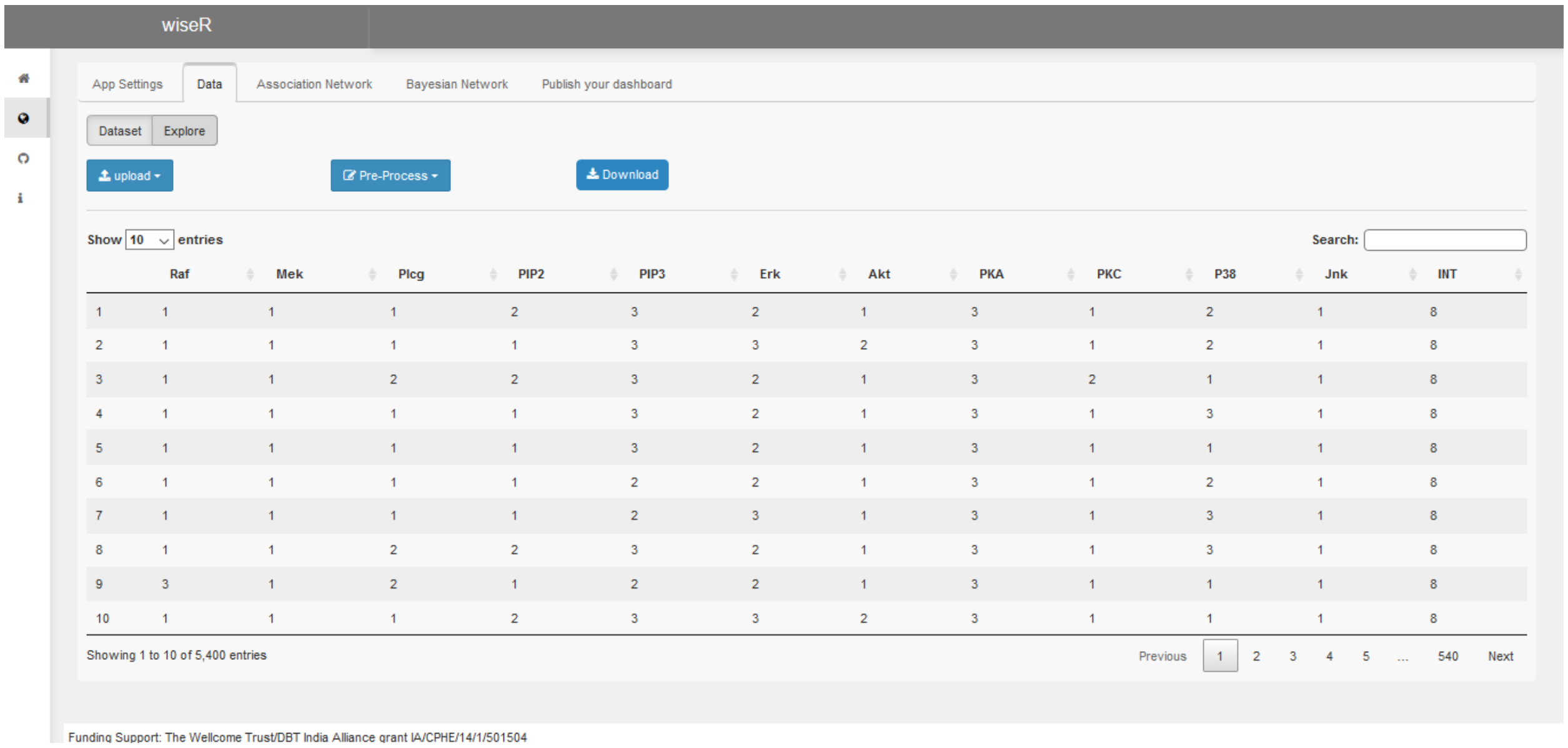

Figure S28. Sachs data uploaded

The variable INT holds interventional information, adjustment is made for the same in the preprocess menu. This will now be used as a part of Bayesian structure learning.

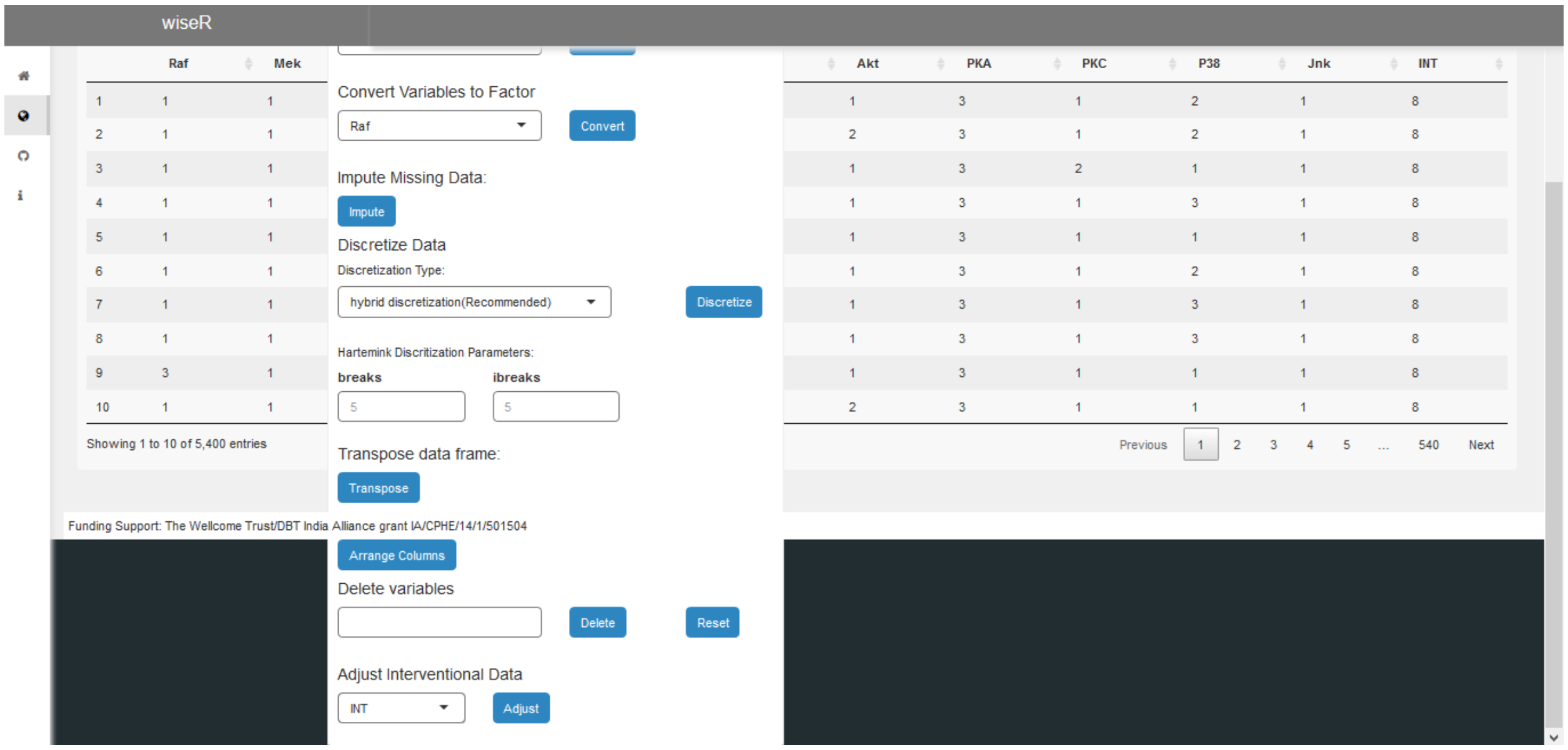

Figure S29. Handling interventional information

Next we learn a simple bayesian network on the data using Hill climbing algorithm and mbde network score. This graph will now be used to initialize structure learning using the tabu algorithm. This mechanism of graph initialization in structure learning is especially useful in score-based learning like Tabu, which prevents it from getting stuck in local maxima.

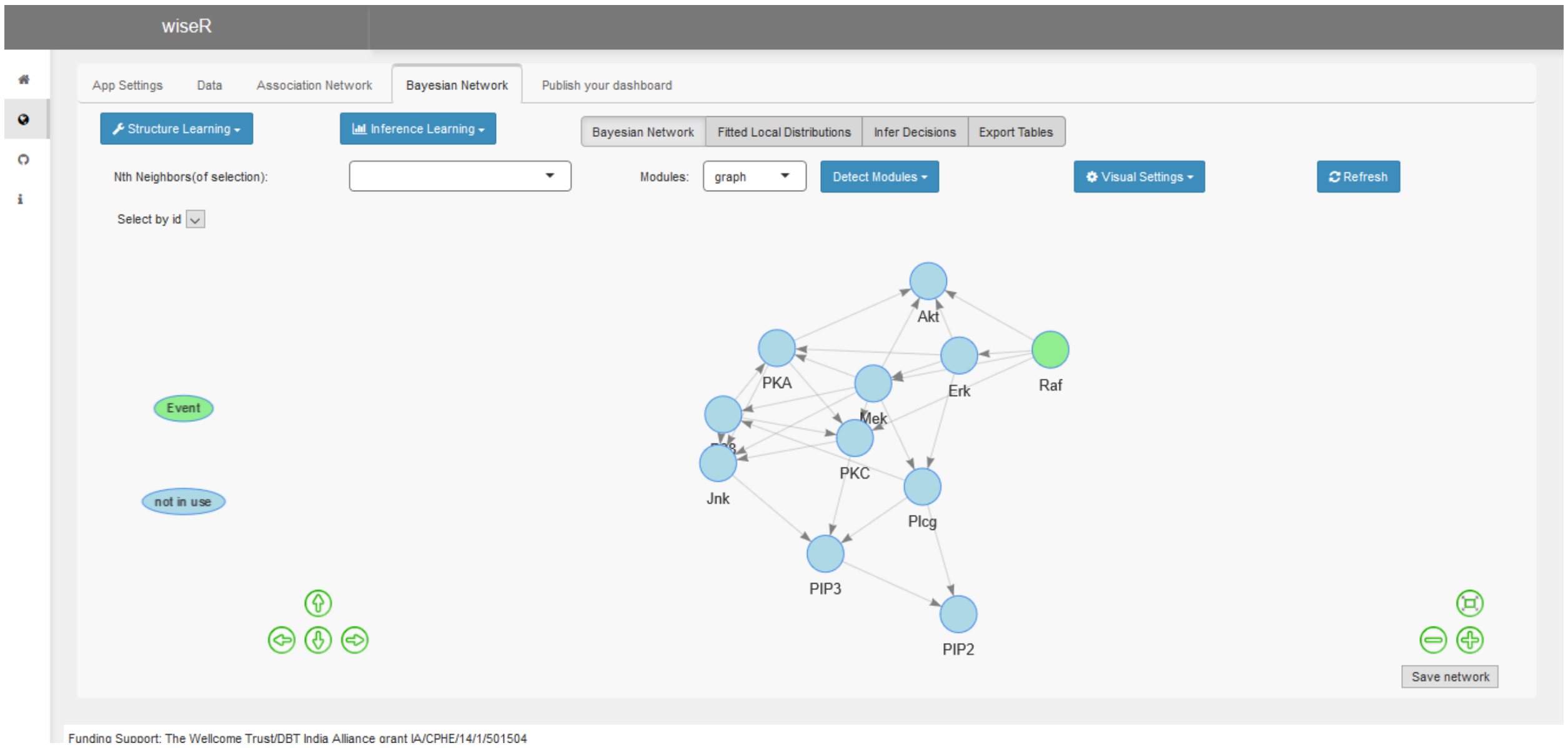

Figure S30. Learn structure using hill climbing

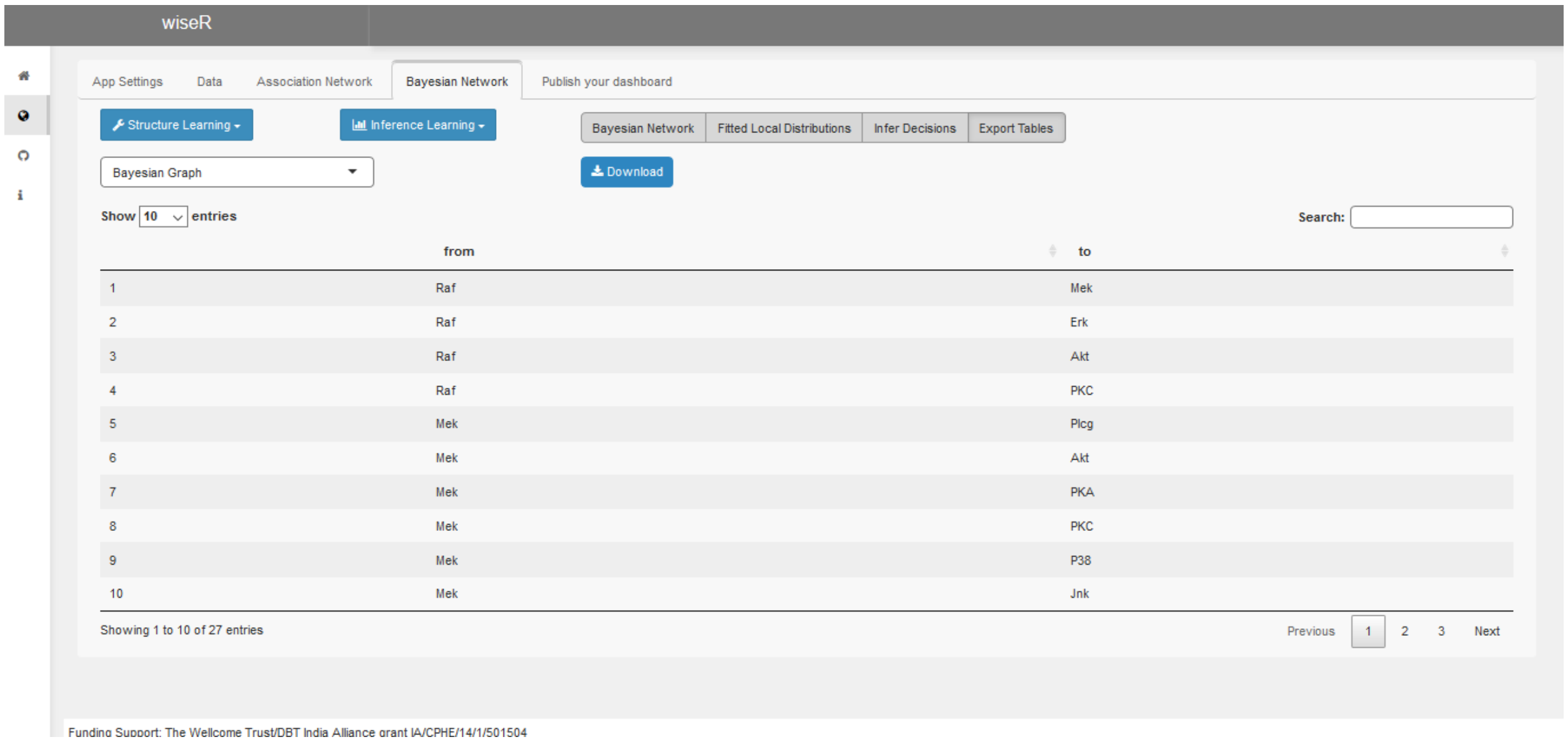

Figure S31. Download network graph as CSV file.

We now upload the graph as initialization for structure learning.

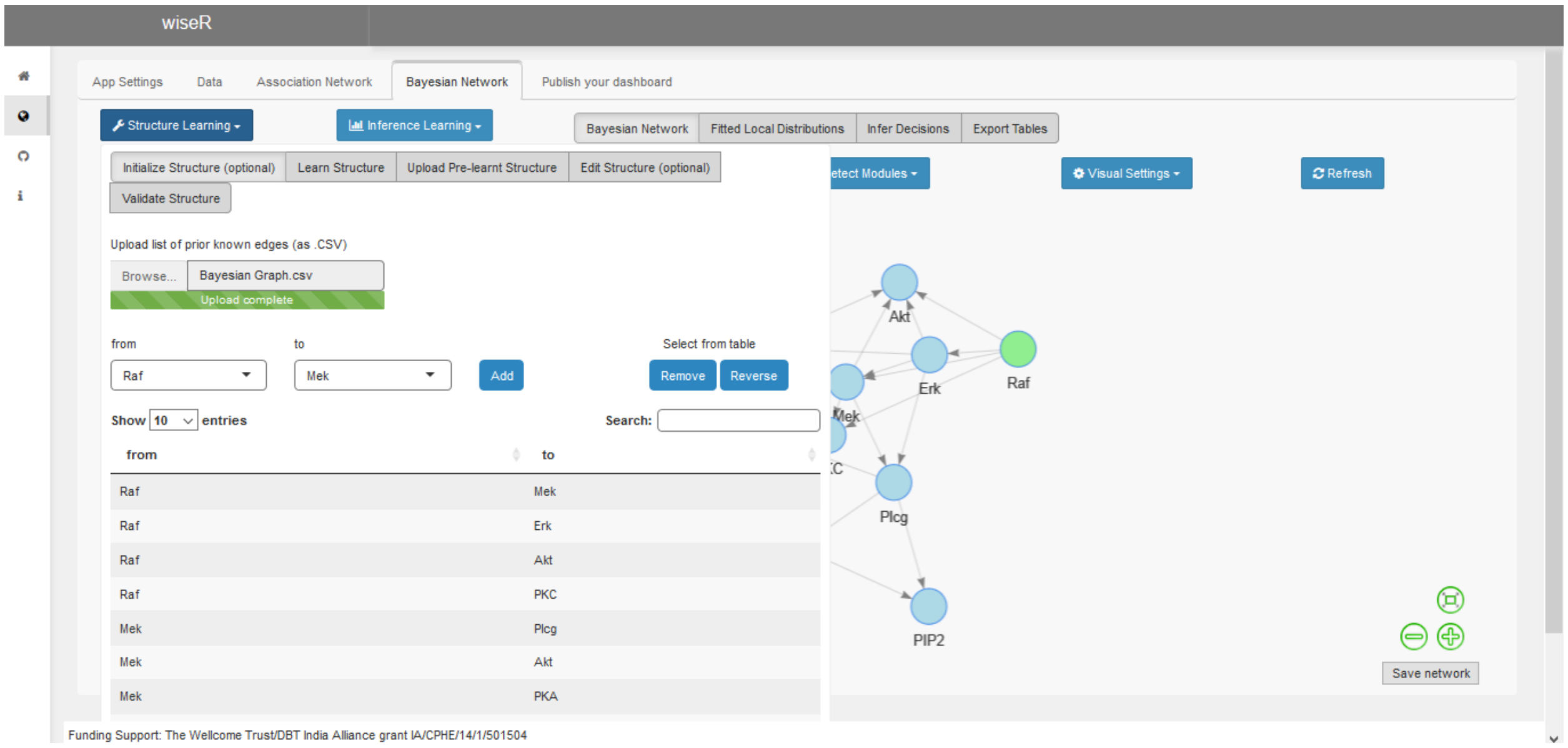

Figure S32. Initialize a new structure using the previously learnt graph

Next, we set the appropriate parameters for structure learning:

- Algorithm: tabu
- Network score: modified bayesian dirichlet equivalent
- ISS(imaginary sample size): 15
- Parameter fitting algorithm: bayesian parameter estimation

This performed a direct structure learning without bootstraps using tabu, to replicate the results as it was published in the study. However, we recommend that the user should select the bootstrap parameters and do a bootstrap learning, to produce more robust structures.

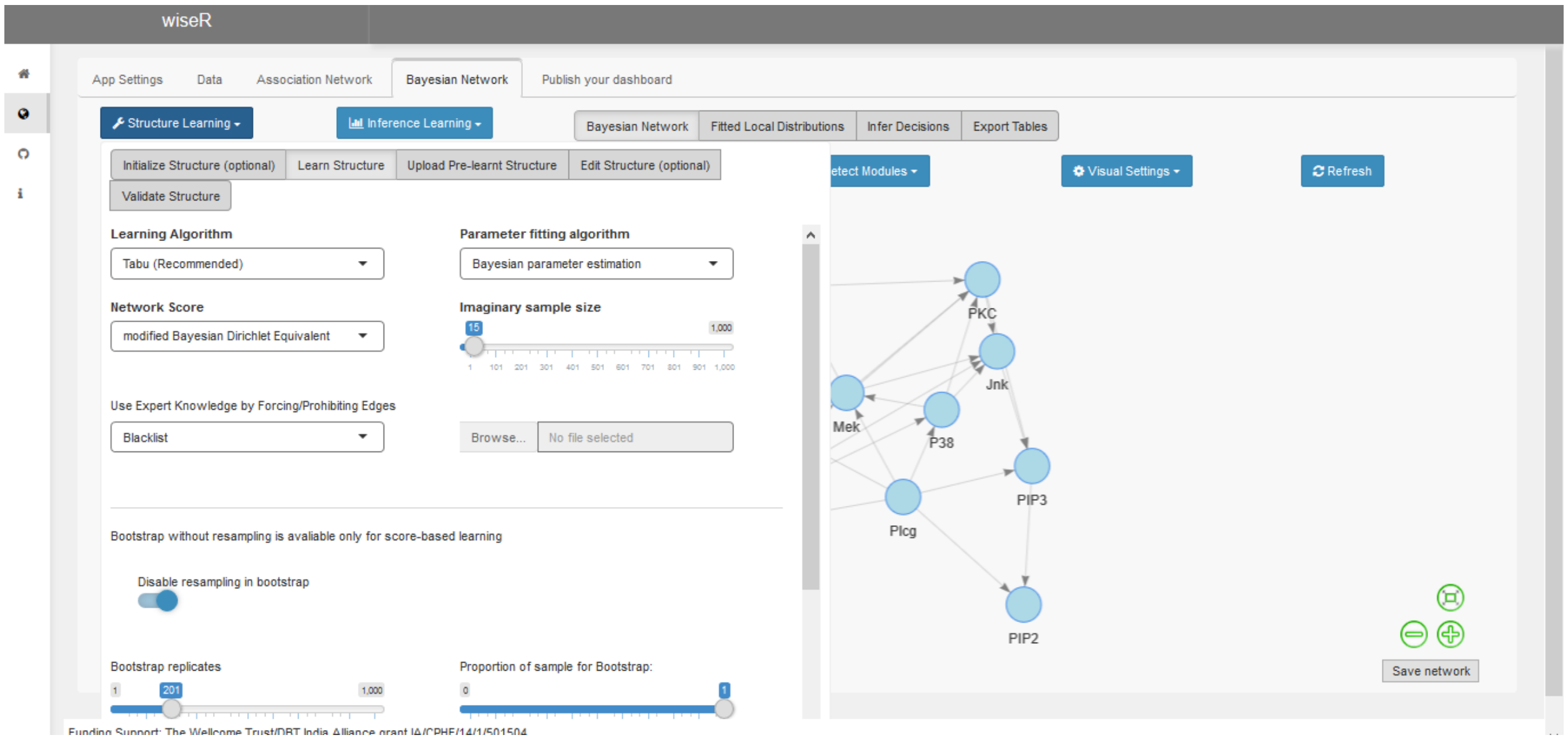

Figure S33. Structure learning using tabu algorithm on Sachs data

The final learnt structure replicates the results of the original paper and has all the validated arcs presented therein. It also discovered additional arcs which were not present in the original paper and may be useful for future research and exploration

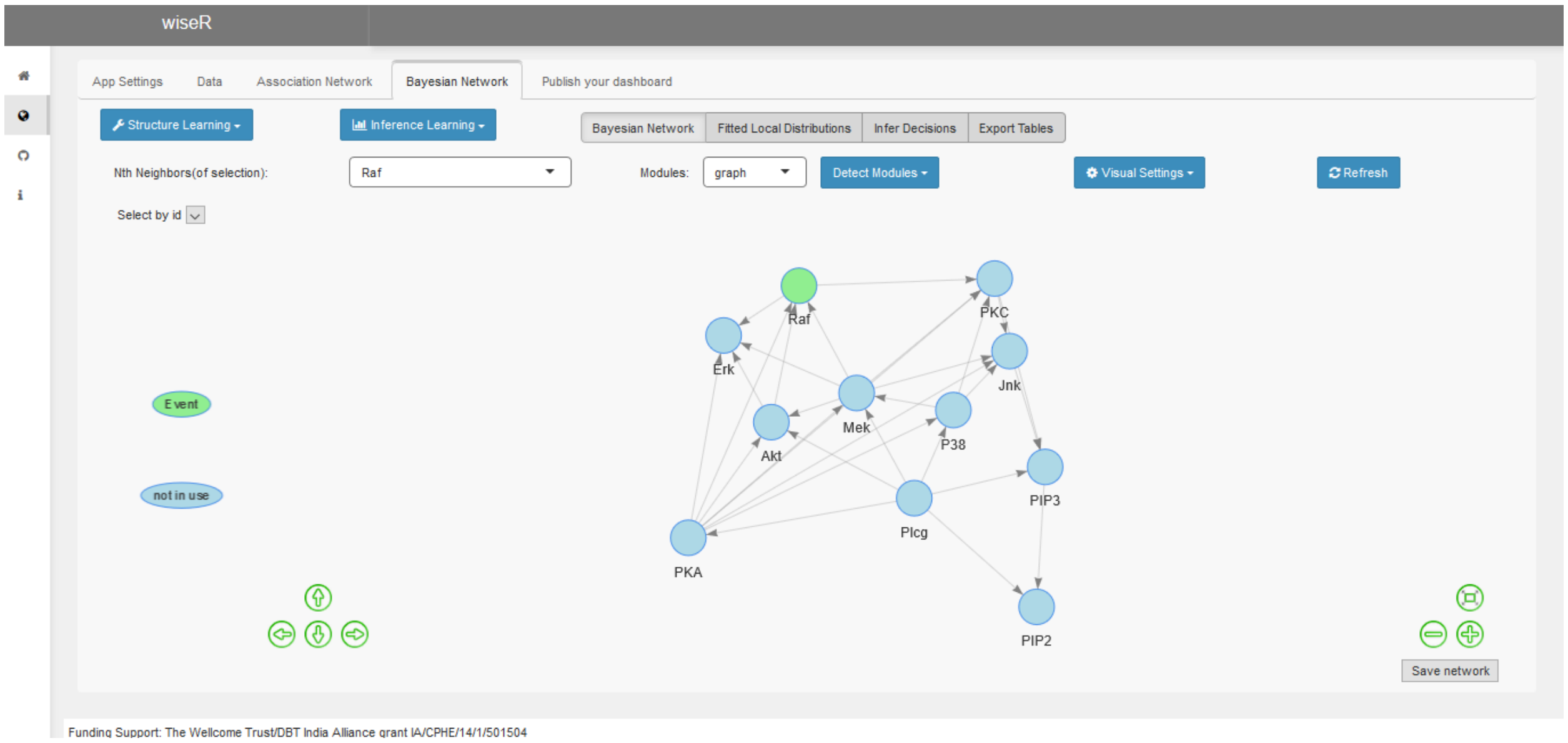

Figure S34. Final learned structure

We now confirm exact inference results with approximate inference with error bars to show that the unique approximate inference mechanism provided in the app produces as accurate results as exact inference without computational constraints in case of large structures. For this purpose we set the node 'Akt' as an event conditional on ‘Erk = 1’. In the probability plots produced we can clearly see that approximate inference with error bars produces as accurate results as exact inference within 0.05 range of error, which is negligible

### 5.2 Exact Inference

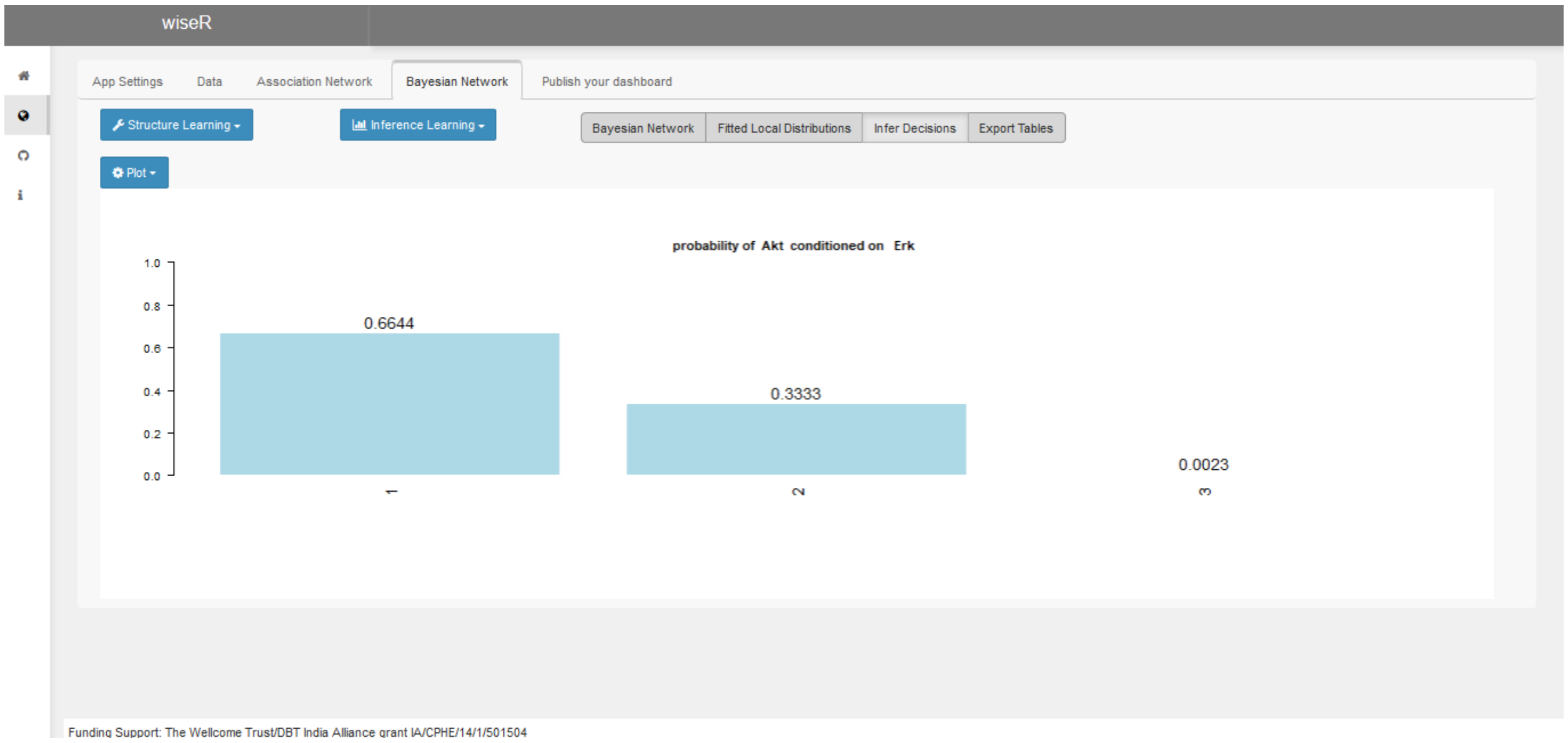

Figure S35. Exact inference plot

### 5.3 Approximate Inference with error bars

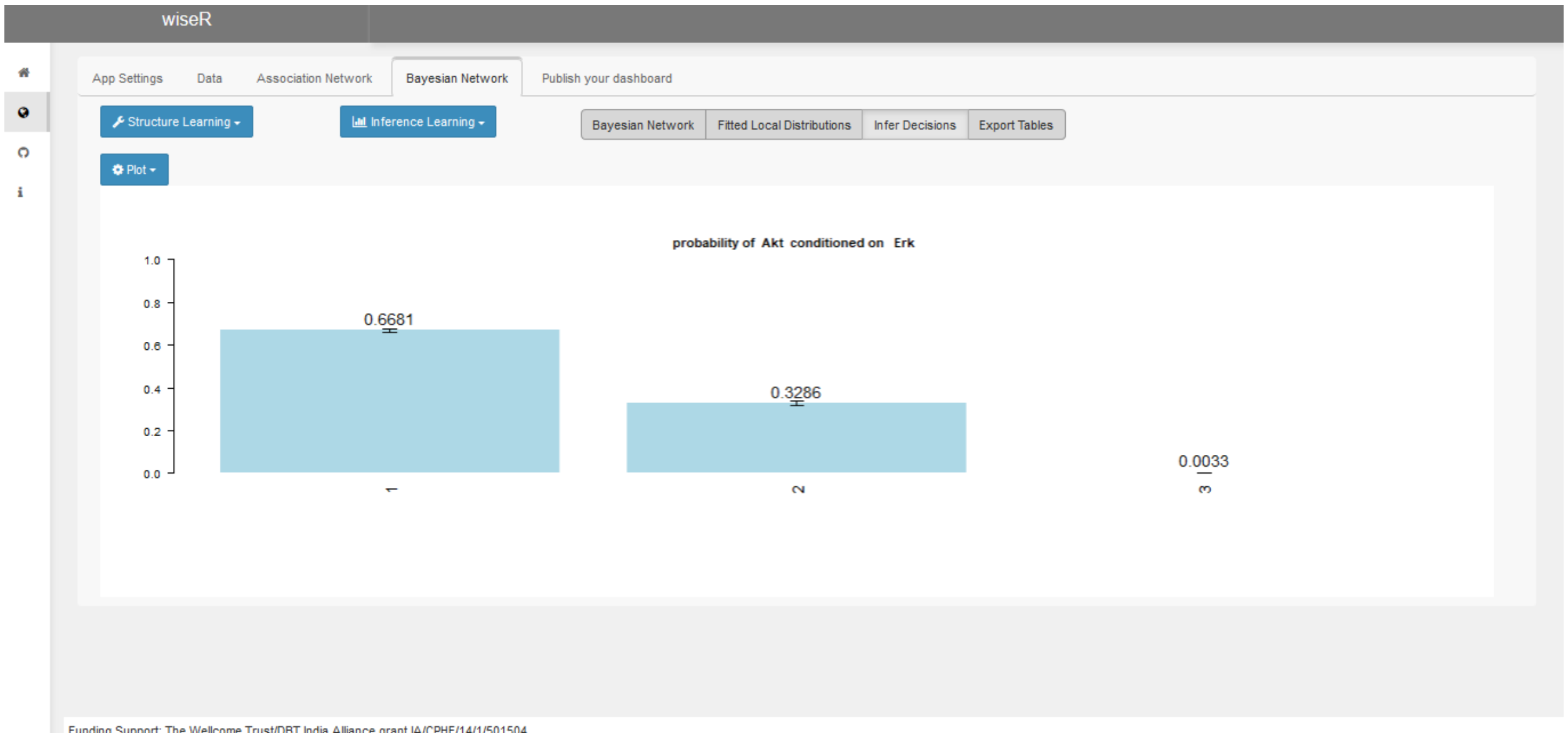

Figure S36. Approximate inference plot

This concludes the walk-through example for replicating a BN analysis using wiseR

## 6 Example-2: Decision Network

This section provides a hands-on example to learn decision networks and policy optimization through wiser. We use the Alarm dataset available in the pre-loaded datasets of the app.

We upload the already discretized and complete Alarm dataset from the app

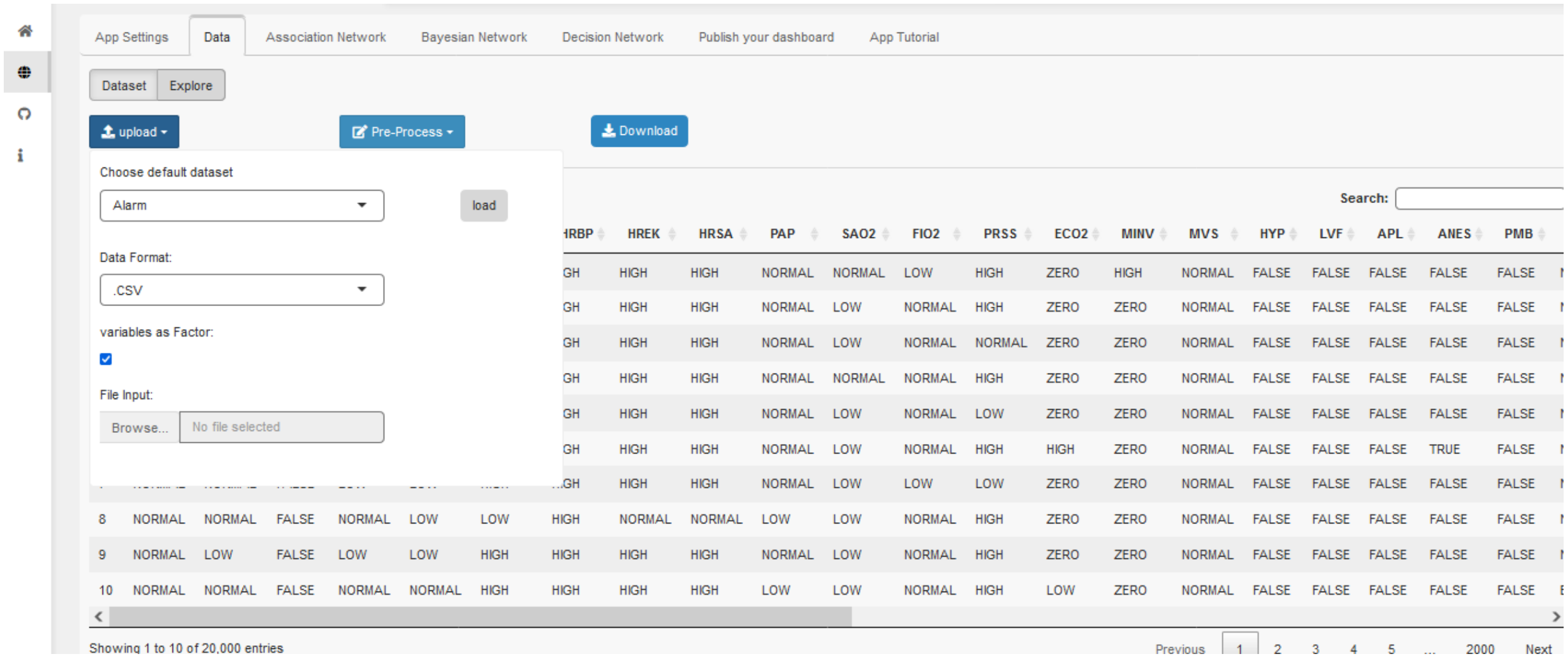

Figure S37. Alarm Data uploaded

We learn the Network structure using the following parameters:

- Algorithm: Hill Climbing
- Network Score: Akaike Information Criterion
- Parameter Fitting:Bayesian Parameter Estimation
- Blacklisting: No
- Bootstrap Model: Yes
- No. of Bootstrap:101
- Edge Strength: >0.51
- Direction Strength: >0.51
- Sample Portion: 1

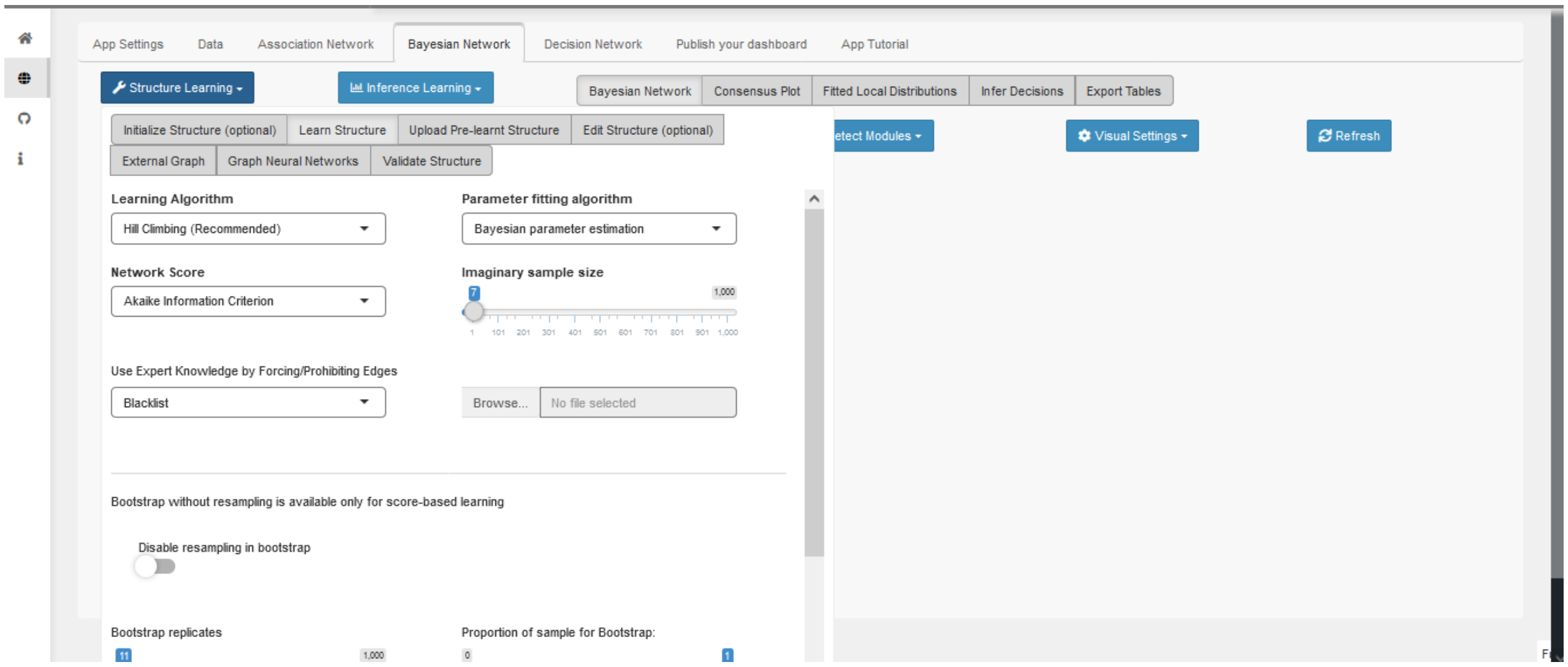

Figure S38. Structure Learning Parameters

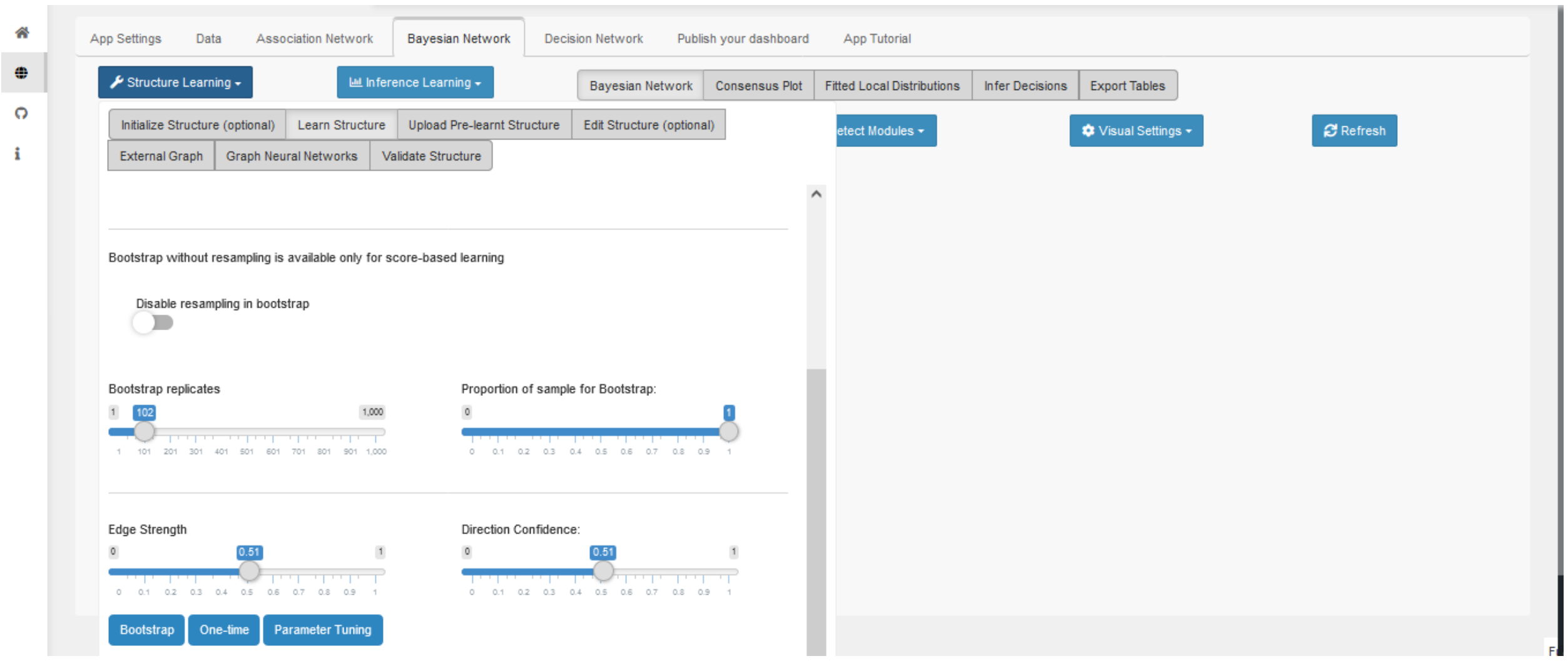

Figure S39. Bootstrap Parameters

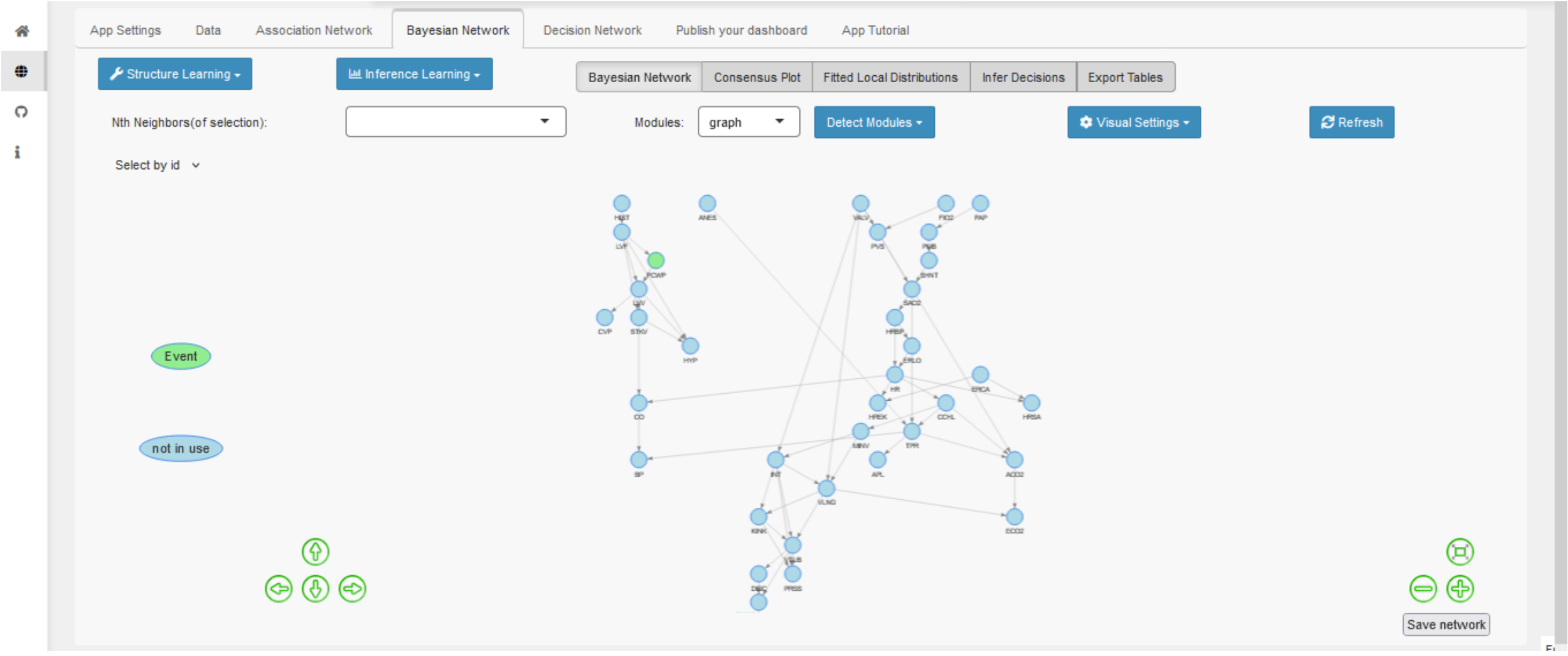

Figure S40. Learnt Bayesian Structure

As shown in the network the BP (blood pressure) has 2 causal parents the CO(cardiac output) and TPR (Total Peripheral Resistance)

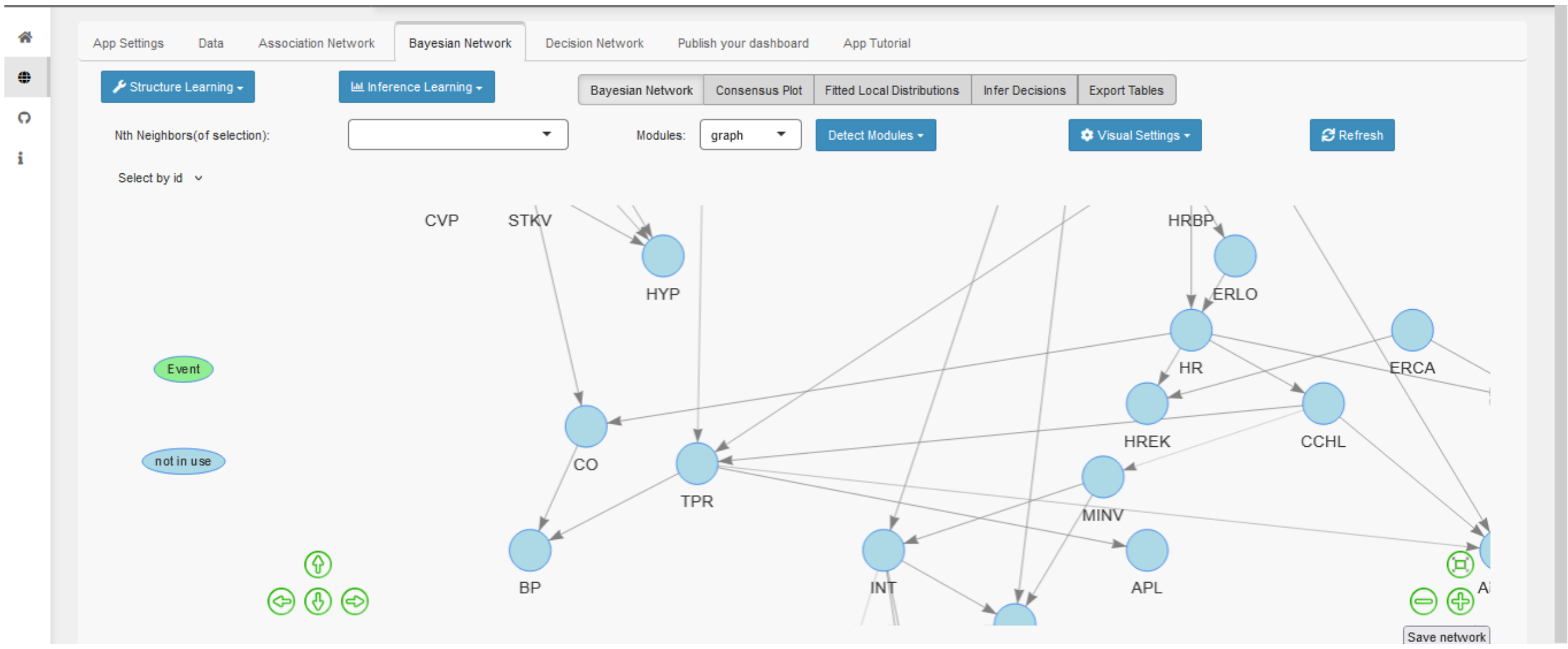

Figure S41. Nodes of interest in our structure

The aim of our decision network now is to find optimal values of CO and TPR to get a NORMAL value of BP. Thus we set our BP as utility node, payoffs of 1 for NORMAL, -0.5 for HIGH and -1 for LOW BP.

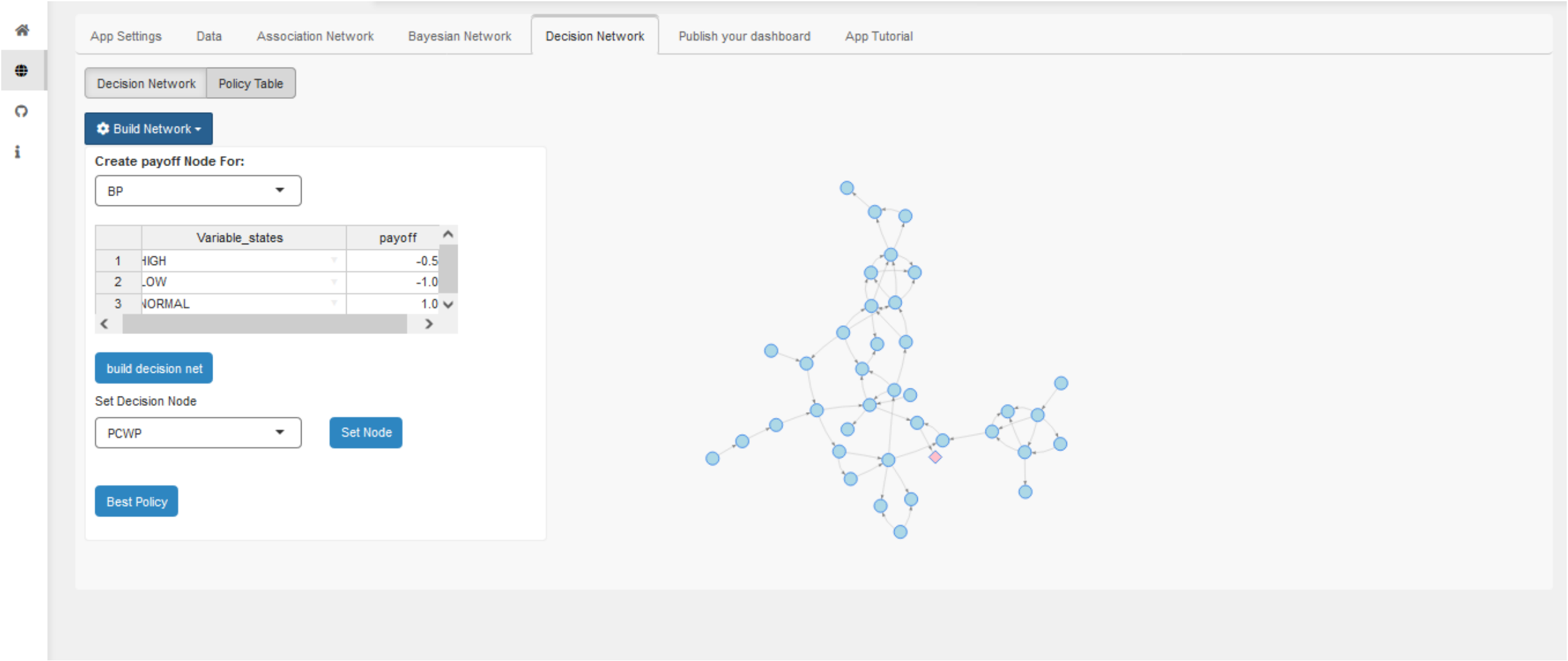

Figure S42. setting up utility node, corresponding payoffs and building the decision network

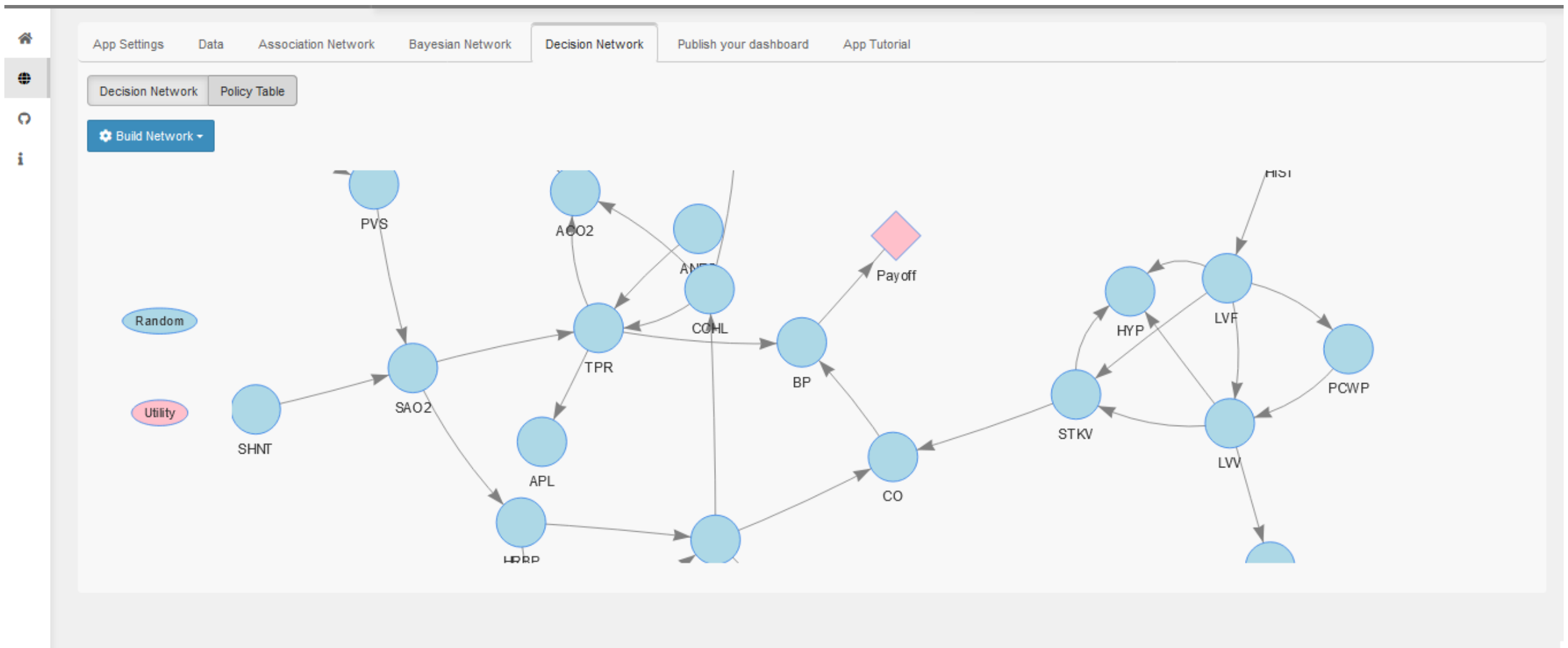

Figure S43. Payoff node in the Decision Network

Now we set CO and TPR as the decision nodes in the network. We aim to learn an optimal policy for both to maximize payoff on utility nodes.

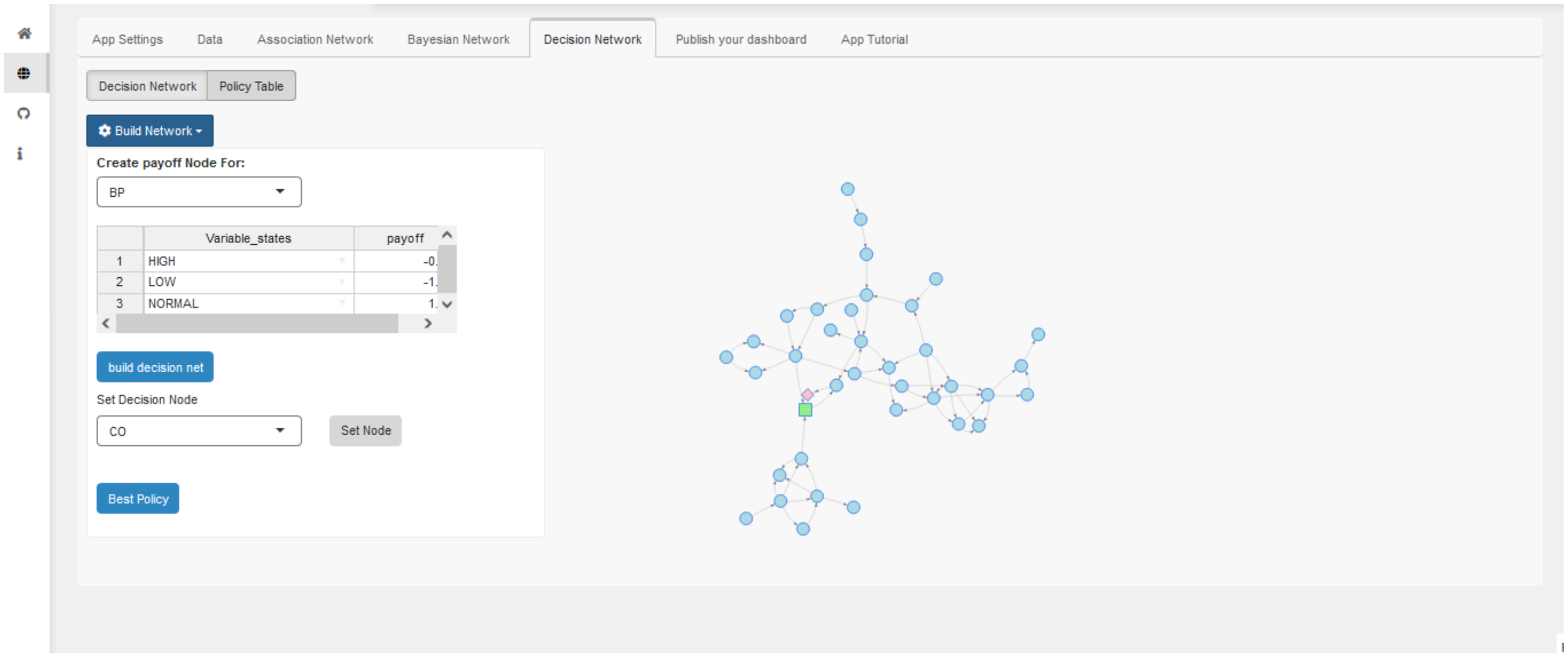

Figure S44. Setting Decision node for policy learning

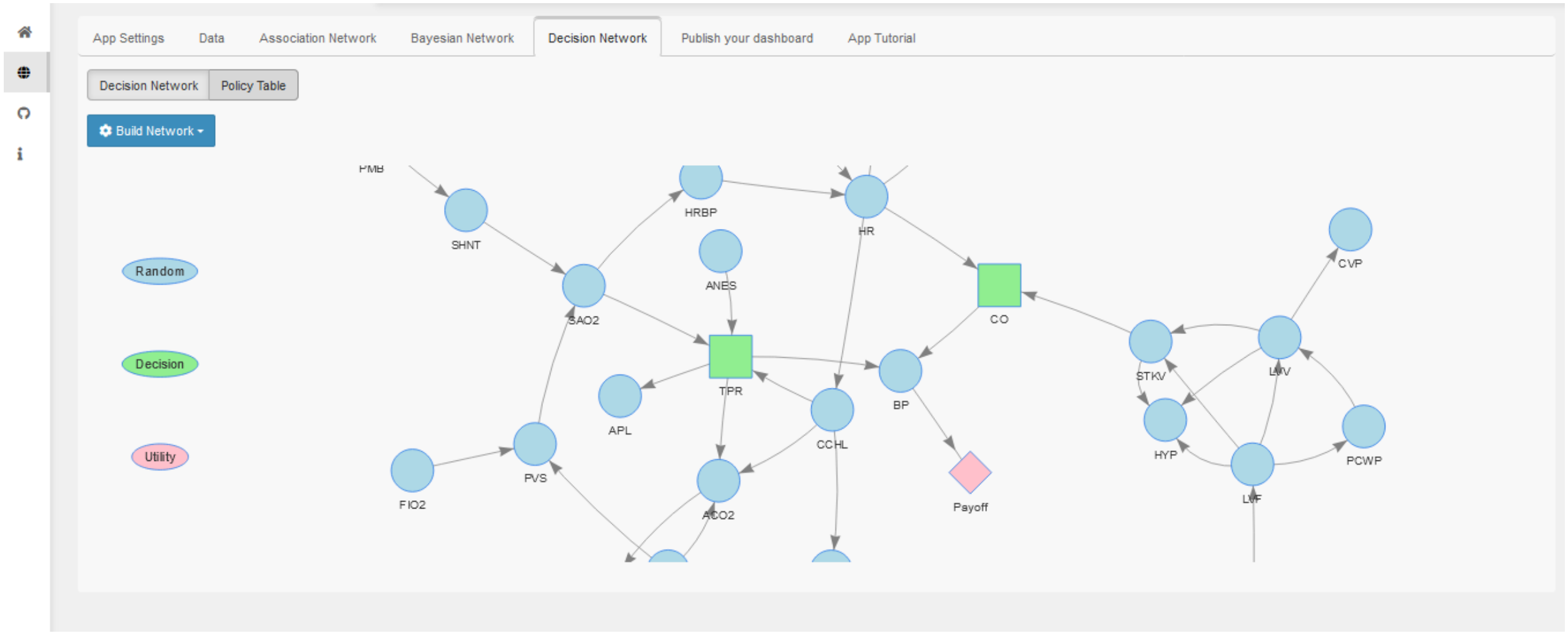

Figure S45. Final Layout of decision network depicting the utility and decision nodes

Finally from policy learning we can see that NORMAL values for both CO and TPR ensure maximum payoff and thus maximum probability of NORMAL BP. It is also interesting to note that a HIGH TPR more strongly affects high probability of NORMAL BP as compared to CO. Thus verifying the knowledge already available in the medical domain and validating the utility of wiser as a policy learning platform.

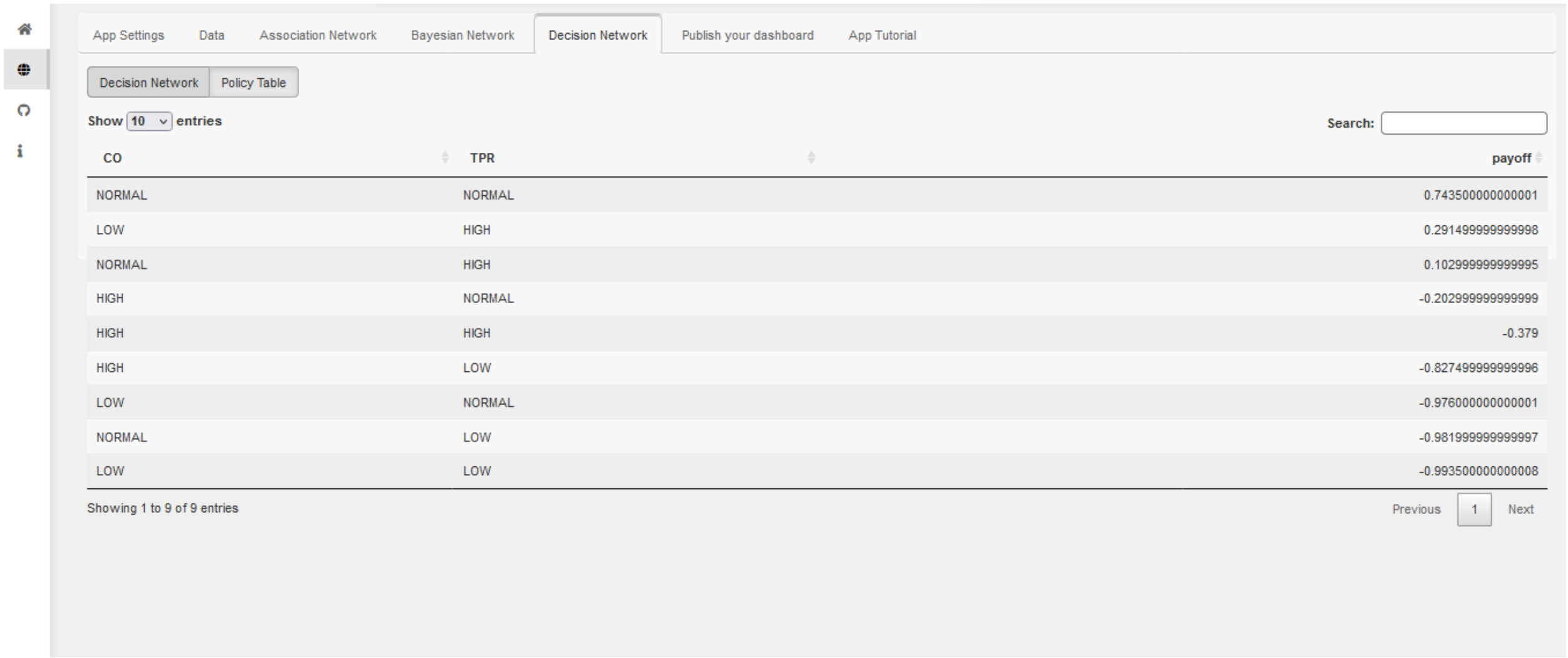

| CO | TPR | payoff |
|---|---|---|
| NORMAL | NORMAL | 0.743500000000001 |
| LOW | HIGH | 0.291499999999998 |
| NORMAL | HIGH | 0.102999999999995 |
| HIGH | NORMAL | -0.202999999999999 |
| HIGH | HIGH | -0.379 |
| HIGH | LOW | -0.827499999999996 |
| LOW | NORMAL | -0.976000000000001 |
| NORMAL | LOW | -0.981999999999997 |
| LOW | LOW | -0.993500000000008 |

Figure S46. Learned policy table with corresponding payoffs

This concludes the walk-through example for replicating a Policy learning framework using Decision Networks using wiseR

## 7 wiseR Libraries

- RBGL (Vince Carey, Li Long and R. Gentleman (2017). RBGL: An interface to the BOOST graph library. R package version 1.52.0. http://www.bioconductor.org )
- graph (R. Gentleman, Elizabeth Whalen, W. Huber and S. Falcon (2017). graph: graph: A package to handle graph data structures. R package version 1.54.0.)
- bnlearn (Marco Scutari (2010). Learning Bayesian Networks with the bnlearn R Package. Journal of Statistical Software, 35(3), 1-22. URL http://www.jstatsoft.org/v35/i03/ .)
- rhandsontable (Jonathan Owen (2018). rhandsontable: Interface to the 'Handsontable.js' Library. R package version 0.3.6. https://CRAN.R-project.org/package=rhandsontable )
- shiny (Winston Chang, Joe Cheng, JJ Allaire, Yihui Xie and Jonathan McPherson (2017). shiny: Web Application Framework for R. R package version 1.0.5. https://CRAN.R-project.org/package=shiny )
- shinydashboard (Winston Chang and Barbara Borges Ribeiro (2017). shinydashboard: Create Dashboards with 'Shiny'. R package version 0.6.1. https://CRAN.R-project.org/package=shinydashboard )
- dplyr (Hadley Wickham, Romain Francois, Lionel Henry and Kirill Müller (2017). dplyr: A Grammar of Data Manipulation. R package version 0.7.4. https://CRAN.R-project.org/package=dplyr )
- visNetwork (Almende B.V., Benoit Thieurmel and Titouan Robert (2018). visNetwork: Network Visualization using 'vis.js' Library. R package version 2.0.3. https://CRAN.R-project.org/package=visNetwork )
- shinyWidgets (Victor Perrier and Fanny Meyer (2018). shinyWidgets: Custom Inputs Widgets for Shiny. R package version 0.4.1. https://CRAN.R-project.org/package=shinyWidgets )
- missRanger (Michael Mayer (2018). missRanger: Fast Imputation of Missing Values. R package version 1.0.2. https://CRAN.R-project.org/package=missRanger )
- tools (R Core Team (2017). R: A language and environment for statistical computing. R Foundation for Statistical Computing, Vienna, Austria. URL https://www.R-project.org/ .)
- shinyalert (Dean Attali and Tristan Edwards (2018). shinyalert: Easily Create Pretty Popup Messages (Modals) in 'Shiny'. R package version 1.0. https://CRAN.R-project.org/package=shinyalert )
- shinycssloaders (Andras Sali (2017). shinycssloaders: Add CSS Loading Animations to 'shiny' Outputs. R package version 0.2.0. https://CRAN.R-project.org/package=shinycssloaders )

- rintrojs (Carl Ganz (2016). rintrojs: A Wrapper for the Intro.js Library. Journal of Open Source Software, 1(6), October 2016. URL http://dx.doi.org/10.21105/joss.00063 )
- arules (Michael Hahsler, Christian Buchta, Bettina Gruen and Kurt Hornik (2018). arules: Mining Association Rules and Frequent Itemsets. R package version 1.6-1. https://CRAN.R-project.org/package=arules )
- psych (Revelle, W. (2018) psych: Procedures for Personality and Psychological Research, Northwestern University, Evanston, Illinois, USA, https://CRAN.R-project.org/package=psych Version = 1.8.4.)
- DescTools (Andri Signorell et mult. al. (2018). DescTools: Tools for descriptive statistics. R package version 0.99.24.)
- DT (Yihui Xie (NA). DT: A Wrapper of the JavaScript Library 'DataTables'. R package version 0.4.11. https://rstudio.github.io/DT )
- linkcomm (Kalinka, A.T. and Tomancak, P. (2011). linkcomm: an R package for the generation, visualization, and analysis of link communities in networks of arbitrary size and type. Bioinformatics 27 (14), 2011-2012.)
- igraph (Csardi G, Nepusz T: The igraph software package for complex network research, InterJournal, Complex Systems 1695. 2006. http://igraph.org )
- parallel (R Core Team (2017). R: A language and environment for statistical computing. R Foundation for Statistical Computing, Vienna, Austria. URL https://www.R-project.org/. )
- snow (Luke Tierney, A. J. Rossini, Na Li and H. Sevcikova (2016). snow: Simple Network of Workstations. R package version 0.4-2. https://CRAN.R-project.org/package=snow )
- shinyBS (Eric Bailey (2015). shinyBS: Twitter Bootstrap Components for Shiny. R package version 0.61. https://CRAN.R-project.org/package=shinyBS )
- gRbase (Claus Dethlefsen, Søren Højsgaard (2005). A Common Platform for Graphical Models in R: The gRbase Package. Journal of Statistical Software, 14(17), 1-12. URL http://www.jstatsoft.org/v14/i17/. )
- gRain (Søren Højsgaard (2012). Graphical Independence Networks with the gRain Package for R. Journal of Statistical Software, 46(10), 1-26. URL http://www.jstatsoft.org/v46/i10/. )
- RCy3 (Ono K, Muetze T, Kolishovski G, Shannon P, Demchak, B. CyREST: Turbocharging Cytoscape Access for External Tools via a RESTful API [version 1; referees: 2 approved]. F1000Research 2015, 4:478.)
- BiocManager ( Martin Morgan (2018). BiocManager: Access the Bioconductor Project Package Repository. R package version 1.30.1. https://CRAN.R-project.org/package=BiocManager )
- HydeNet (Jarrod E. Dalton and Benjamin Nutter (2018). HydeNet: Hybrid Bayesian Networks Using R and JAGS. R package version 0.10.7. https://CRAN.R-project.org/package=HydeNet )